\documentclass{article}

\usepackage{microtype}
\usepackage{graphicx}
\usepackage{subcaption}
\usepackage{booktabs} % for professional tables

\usepackage{amsthm}

\usepackage{xurl}



\usepackage[accepted]{icml2026}

\usepackage{amsmath}
\usepackage{amssymb}
\usepackage{amsfonts}
\usepackage{mathtools}
\usepackage{amsthm}
\usepackage{dsfont}
\usepackage{algorithm}
\usepackage{algorithmic}
\usepackage{comment}
\usepackage{multirow}
\usepackage{booktabs}
\usepackage{enumitem}
\usepackage{subcaption} 
\usepackage{float}
\usepackage{wrapfig}
\usepackage{tcolorbox}
\tcbuselibrary{listings,breakable}
\newtcblisting{promptbox}{
  breakable,
  colback=gray!5,
  colframe=gray!80,
  boxrule=0.5pt,
  arc=4pt,
  outer arc=4pt,
  left=6pt,right=6pt,top=6pt,bottom=6pt,
  listing only,
  listing options={style=promptstyle},
}
\usepackage{listings}
\lstdefinestyle{promptstyle}{
  basicstyle=\ttfamily\scriptsize,
  columns=fullflexible,
  breaklines=true,
  breakatwhitespace=true,   
  breakautoindent=false,    
  breakindent=0pt,         
  showstringspaces=false,
  keepspaces=true,
  upquote=true,
  mathescape=false,
  literate={_}{{\_}}1
}

\usepackage[toc,page,header]{appendix}
\usepackage{minitoc}

\definecolor{mydarkblue}{rgb}{0,0.08,0.45}
\usepackage[pagebackref,breaklinks,colorlinks,linkcolor=mydarkblue,citecolor=mydarkblue,filecolor=mydarkblue,urlcolor=mydarkblue]{hyperref}
\usepackage[capitalize,noabbrev]{cleveref}
\crefname{appsec}{Appendix}{Appendices}
\Crefname{appsec}{Appendix}{Appendices}

\theoremstyle{plain}
\newtheorem{theorem}{Theorem}[section]

\theoremstyle{definition}

\theoremstyle{remark}

\newcommand{\oAcc}{\operatorname{Acc}}
\defcitealias{occ_fair_lending}{OCC}

\usepackage[textsize=tiny]{todonotes}

\icmltitlerunning{Beyond Procedure: Substantive Fairness in Conformal Prediction}

\begin{document}

\doparttoc 
\faketableofcontents 

\twocolumn[
  \icmltitle{Beyond Procedure: Substantive Fairness in Conformal Prediction}



  \icmlsetsymbol{equal}{*}

  \begin{icmlauthorlist}
    \icmlauthor{Pengqi Liu}{equal,sch1}
    \icmlauthor{Zijun Yu}{equal,sch1}
    \icmlauthor{Mouloud Belbahri}{td}
    \icmlauthor{Arthur Charpentier}{sch2}
    \icmlauthor{Masoud Asgharian}{sch1}
    \icmlauthor{Jesse C. Cresswell}{L6}
  \end{icmlauthorlist}

  \icmlaffiliation{L6}{Layer 6 AI, Toronto, Canada}
  \icmlaffiliation{td}{TD Insurance, Montreal, Canada}
  \icmlaffiliation{sch1}{McGill University, Montreal, Canada}
  \icmlaffiliation{sch2}{UQAM, Montreal, Canada}

  \icmlcorrespondingauthor{Pengqi Liu}{pengqi.liu@mail.mcgill.ca}

  \icmlkeywords{Conformal Prediction, Fairness}

  \vskip 0.3in
]



\printAffiliationsAndNotice{\icmlEqualContribution}

\begin{abstract}
  Conformal prediction (CP) offers distribution-free uncertainty quantification for machine learning models, yet its interplay with fairness in downstream decision-making remains underexplored. Moving beyond CP as a standalone operation (\emph{procedural} fairness), we analyze the holistic decision-making pipeline to evaluate \emph{substantive} fairness—the equity of downstream outcomes. Theoretically, we derive an upper bound that decomposes prediction-set size disparity into interpretable components, clarifying how label-clustered CP helps control method-driven contributions to unfairness. To facilitate scalable empirical analysis, we introduce an LLM-in-the-loop evaluator that approximates human assessment of substantive fairness across diverse modalities. Our experiments show that label-clustered CP often provides a favorable balance between utility and substantive fairness, while reducing set-size disparities in line with our theory. Finally, we empirically show that equalized set sizes, rather than coverage, strongly correlate with improved substantive fairness, enabling practitioners to design more fair CP systems. Our code is available at \url{https://github.com/layer6ai-labs/llm-in-the-loop-conformal-fairness}.

\end{abstract}

\section{Introduction}\label{sec:intro}
Conformal prediction (CP)~\citep{vovk2005algorithmic,shafer2008tutorial} provides finite-sample, distribution-free statistical guarantees through a well-defined procedure; yet, whether these \emph{procedural} guarantees translate into \emph{equitable outcomes} in downstream decision-making remains unclear. In high-stakes domains, reliable uncertainty quantification is essential for building trustworthy models. Unlike other methods that rely on strong assumptions about the data distribution \citep{gal2016dropout, lakshminarayanan2017simple} or require architectural modifications \citep{neal2012bayesian}, CP is distribution-free, model-agnostic, and applies directly to any black-box predictor~\citep{angelopoulos2023conformalintro}. However, the rigorous procedural nature of CP does not automatically ensure equitable outcomes, necessitating a deeper investigation into how these statistical bounds influence fairness in practice.

Fairness in machine learning~\cite{barocas2023fairness}, particularly in regulated fields such as healthcare and finance, is commonly understood through two complementary perspectives: \emph{procedural fairness}, which concerns the integrity of the decision process (e.g., fairness through unawareness \citep{zemel2013learning,kusner2017counterfactual}); and \emph{substantive fairness}, which focuses on equitable outcomes across groups (e.g., Equalized Odds \citep{hardt2016equality}). In the CP setting, procedural fairness refers to properties of the prediction sets themselves, such as equalized coverage or equalized set size across groups. By contrast, substantive fairness asks whether these sets provide comparable downstream benefits to different groups; in our setting, this means whether CP-assisted decisions improve accuracy by similar amounts across protected groups relative to a control condition without prediction sets. For example, a CP method may achieve 90\% coverage for every demographic group, yet produce compact, informative prediction sets for one group and much larger, less actionable sets for another. Such a method is procedurally fair with respect to coverage, but may still be substantively unfair if the downstream accuracy improvement is much larger for the first group than for the second. Prior research in CP has mainly focused on procedural fairness, treating CP as a standalone process \cite{romano2020with}. In practice, CP constitutes one step in a larger pipeline that includes downstream decisions, a perspective also studied from a decision-theoretic viewpoint by \citet{kiyani2025decision}. The interactions of CP with procedural and substantive notions of fairness in this broader context remain less well understood \cite{cresswell2025trustworthy}.

In this work, we move beyond viewing CP as a standalone operation to analyze the \emph{holistic} decision-making pipeline. While ultimate fairness is defined by substantive outcomes, procedural choices within CP play a critical role in shaping these results. We aim to uncover the specific connections between procedural properties and substantive fairness, enabling the design of procedures that positively influence downstream equity. By evaluating fairness as an emergent property of the entire pipeline, we can distinguish between procedural metrics that are merely performative and those that genuinely drive fair outcomes.

Our main contributions are threefold:\\
\textbf{Scalable LLM-in-the-loop fairness evaluation.} To overcome the resource constraints of human-subject experiments, we leverage large language models (LLMs) in an evaluation protocol that approximates human decision behavior. We validate that this evaluator produces results comparable to human-in-the-loop benchmarks, enabling us to scale our analysis of substantive fairness across a broader range of datasets and algorithms than prior work.\\
\textbf{Connecting procedural properties to substantive fairness.} We explicitly map the relationships between procedural CP metrics and substantive outcomes. Crucially, we find that \emph{Equalized Set Size} correlates strongly with improved substantive fairness, whereas the standard goal of \emph{Equalized Coverage} often has negative effects. This insight shifts the design objective from coverage parity to set size parity.\\
\textbf{Theoretical and empirical validation of Label-Clustered CP.} Guided by the connection between set sizes and substantive fairness, we analyze Label-Clustered CP. We derive a theoretical upper bound decomposing the set size disparity into interpretable components. Experimentally, we confirm that Label-Clustered CP reduces set size disparity more effectively than marginal or group-conditional approaches, and offers an effective balance between prediction-set usefulness and substantive fairness across our evaluations.

\section{Background}\label{sec:background}
\subsection{Conformal Set Predictors}
Consider inputs \(x \in \mathcal{X} \subset \mathbb{R}^d\) with ground truth labels \(y \in \mathcal{Y} = [m] := \{1,\dots,m\}\), drawn from a joint distribution \((x, y) \sim \mathbb{P}\). Let \(f : \mathcal{X} \to \Delta^{m-1} \subset \mathbb{R}^m\) be a classifier outputting predicted probabilities, where $\Delta^{m-1}$ is the $(m{-}1)$-dimensional probability simplex. CP constructs a set-valued function \(\mathcal{C} : \mathcal{X} \to \mathcal{P}(\mathcal{Y})\) where \(\mathcal{P}(\mathcal{Y})\) denotes the power set of \(\mathcal{Y}\), such that the following marginal \emph{coverage guarantee} holds,
\begin{equation}\label{eq:marginal-coverage}
    \mathbb{P}[y \in \mathcal{C}(x)] \geq 1 - \alpha,
\end{equation}
where $\alpha \in [0,1]$ is user-specified \citep{vovk1999machine, vovk2005algorithmic}.

CP achieves coverage by varying set size \( |\mathcal{C}(x)| \) based on a calibrated notion of model confidence. Calibration relies on a held-out dataset \(\mathcal{D}_{\mathrm{cal}} = \{(x_i,y_i)\}_{i=0}^{n_{\mathrm{cal}}}\) consisting of $n_{\mathrm{cal}}$ datapoints drawn from $\mathbb{P}$. A \emph{conformal score function} \(s : \mathcal{X} \times \mathcal{Y} \to \mathbb{R}\) measures non-conformity between a candidate label and an input datapoint $x$, with higher scores indicating poorer agreement. The score function is often defined to make use of information from the classifier $f$ in judging the level of agreement.

Let \(S_i := s(x_i,y_i)\) for \(i \in [n_{\mathrm{cal}}]\), and define
\begin{equation}\label{eq:conformal-quantile}
  \tau_\alpha := \frac{\lceil (n_{\mathrm{cal}} + 1)(1-\alpha) \rceil}{n_{\mathrm{cal}}}
  \in (0,1].
\end{equation}
The empirical conformal threshold is then given by
\begin{equation}\label{eq:conformal-threshold}
  \hat{q}_\alpha
  := \operatorname{Quantile}_{\tau_\alpha}(S_1,\dots,S_{n_{\mathrm{cal}}})
  \in \mathbb{R}.
\end{equation}
For a test point \(x_{\mathrm{test}}\) drawn from the $x$-marginal of distribution $\mathbb{P}$, a \emph{conformal prediction set} is constructed as
\begin{equation}\label{eq:prediction-set}
  \mathcal{C}_{\hat{q}_\alpha}(x_{\mathrm{test}})
  :=
  \{y \in \mathcal{Y} \mid s(x_{\mathrm{test}},y) \le \hat{q}_\alpha\}.
\end{equation}
Sets constructed this way will satisfy $1-\alpha$ coverage (\Cref{eq:marginal-coverage}) for any score function \(s\), but smaller sets are more useful for downstream uncertainty quantification applications \cite{cresswell2024cphuman}. The average set size $\mathbb{E}[|\mathcal{C}|]$ is dictated by the quality of $s$, and in turn by the accuracy and calibration of the classifier \(f\). Efficient score functions like APS \citep{romano2020aps}, RAPS \citep{angelopoulos2021uncertainty}, and SAPS \citep{huang2024saps} aim to minimize $\mathbb{E}[|\mathcal{C}|]$ while maintaining coverage.

\subsection{Fairness Notions for Set Predictors}

We briefly review common fairness notions in machine learning and discuss how they apply to conformal set predictors. Let \(\mathcal A = [k_g]\) denote a finite set of sensitive group labels, and let
\(g : \mathcal X \to \mathcal A\) be a group assignment function. Each group is
defined as
\begin{equation}\label{eq:subgroup-Ga}
    G_a := \{x \in \mathcal X : g(x) = a\}, \qquad a \in \mathcal A.
\end{equation}
\textbf{Fairness via non-discrimination criteria.}
In classical supervised learning with point predictions, statistical fairness notions often require parity of prediction behavior across groups. For example, demographic parity requires
\[
  \mathbb P(\hat y = 1 \mid X \in G_a)
  =
  \mathbb P(\hat y = 1 \mid X \in G_b), \quad \forall \ a, b\in \mathcal{A},
\]
with $y = 1$ denoting some important outcome, while Equalized Odds further conditions on the true label \citep{hardt2016equality}. These criteria aim to ensure that outcomes are not systematically skewed by group membership, and hence are aligned with substantive fairness—the predominant paradigm for fairness in regulatory frameworks (\citetalias{occ_fair_lending}, \citeyear{occ_fair_lending}), and in machine learning~\cite{green2022escaping}.

In CP, non-discrimination fairness is commonly formulated in terms of \emph{group-conditional coverage}, where \(\mathcal C\) satisfies
\begin{equation}\label{eq:group-coverage}
  \mathbb P\bigl[y \in \mathcal C(x) \mid x \in G_a\bigr] \ge 1 - \alpha,
  \quad \forall \ a \in \mathcal A.
\end{equation}
Each group receives the same nominal statistical guarantee, achieving \emph{Equalized Coverage} \cite{romano2020with}. \textbf{Mondrian CP} achieves \Cref{eq:group-coverage} by using the predefined grouping function \(g\) to calibrate conformal thresholds separately within each group \citep{vovk2003mondrian}. Since this means partitioning the calibration set \(\mathcal{D}_{\mathrm{cal}}\), each group is calibrated on a smaller sample, leading to increased variance of empirical coverage \citep{zwart2025probabilistic, gibbs2025conformal}. 

However, Equalized Coverage focuses on the construction of prediction sets—an intermediate tool for uncertainty quantification. Hence it is a procedural notion, ignoring how sets are used and what their downstream impact may be. \citet{cresswell2025conformal} showed via randomized controlled trials that equalizing coverage causes disparate impact in downstream tasks where people use prediction sets as decision aids. As an alternative fairness notion for CP, \citet{cresswell2025conformal} proposed \emph{Equalized Set Size} which requires 
\begin{equation}\label{eq:equalized-set-size}
    \mathbb{E}[|\mathcal{C}(x)| \mid x \in G_a] \equiv c, \quad \forall \ a \in \mathcal A,
\end{equation}
for some constant $c$.
While still procedural in nature, this notion better correlated with reduced disparate impact.

\subsection{Advanced Conformal Prediction Variants}
\label{sec: advanced-CP-variants}

Beyond marginal and group-conditional coverage, several CP variants target alternative statistical guarantees. Exact conditional coverage at every \(x \in \mathcal{X}\) is known to be impossible without strong assumptions \citep{vovk2012conditional, lei2013distribution, foygelbarber2020limits}.
Instead, clustered conformal prediction~\citep{ding2023classconditional} seeks approximate conditional coverage by partitioning the \emph{label space} into clusters via a learned clustering function $h: \mathcal{Y}\to [K]$, and calibrating independent thresholds $\hat{q}_{k}$ for each cluster $k \in [K]$. For a test input $x_{\mathrm{test}}$, each label $y$ is included in $\mathcal{C}(x_{\mathrm{test}})$ if its score $s(x_{\mathrm{test}}, y)$ is below the threshold $\hat{q}_{h(y)}$. Clustered conditional coverage follows as
\begin{equation}\label{eq:clustered-coverage}
\mathbb{P}[y_{\mathrm{test}} \in \mathcal{C}(x_{\mathrm{test}}) \mid h(y_{\mathrm{test}}) = k] \geq 1 - \alpha,
\end{equation}
for all clusters.
This adapts thresholds to label-specific difficulty, yielding empirically improved conditional coverage without requiring predefined instance groups. 

The same partitioning-through-clustering strategy also applies when we partition the \textit{group space} into clusters via a learned clustering function $\tilde{h}: \mathcal{A} \to [K]$, and calibrate independent thresholds for each cluster of groups. We refer to these two methods as \textbf{Label-Clustered} and  \textbf{Group-Clustered} CP, respectively.

\textbf{Backward CP} \citep{gauthier2025backwardconformalprediction} reverses the usual prioritization: instead of fixing the coverage level $\alpha$ and accepting variable set sizes, it constrains the set size via a data-dependent rule $\mathcal{T}$ while providing a relaxed marginal coverage guarantee:
\begin{equation}\label{eq:backwards-coverage}
\mathbb{P}[y_{\mathrm{test}} \in \mathcal{C}(x_{\mathrm{test}})] \geq 1 - \mathbb{E}[\tilde{\alpha}],
\end{equation}
where the random variable $\tilde{\alpha} > 0$ is chosen to respect the size constraint. The prediction set is constructed using \emph{e-values} \citep{VOVK2025econformal} derived from the non-conformity scores; labels with sufficiently small e-values are included until the size constraint is reached.

Detailed mathematical formulations and pseudocode for Marginal, Mondrian, Label-Clustered, Group-Clustered, and Backward CP are provided in Appendix~\ref{app:conformal}.

\vspace{-4pt}
\section{Related Work}\label{sec:related}

The study of fairness in applications of CP is an emergent field, and several alternative directions have recently been introduced. Initially researchers adopted Equalized Coverage \citep{romano2020with, zhou2024} and suggested pursuing it in deployments of CP \citep{lu2022fair, zerva-martins-2024-conformalizing, garcia2025fair}. More recently this standard has been reexamined, with significant concerns being raised about its practical consequences \citep{cresswell2025conformal}. More broadly, researchers have applied existing group algorithmic fairness notions to prediction sets, including demographic parity \citep{liu2022fairnessCQR}, Equal Opportunity \citep{wang2024equal}, and others \citep{vadlamani2025a}. Individual fairness notions like counterfactual fairness \cite{kusner2017counterfactual} have also been extended to CP \citep{guldogan2025counterfactual}.

While these notions have been applied in various settings \citep{kuchibhotla2023nested, berk2023criminal, srinivasan2025fedcf}, the fairness definitions above pertain only to coverage and the construction of prediction sets, rather than impact in downstream tasks. 
Two exceptions are the work of  \citet{cresswell2025conformal} discussed above, and \citet{tasar2025} which defers decisions to an alternate process—such as a human-in-the-loop—unless the model expresses confidence via a singleton prediction set. While \citet{cresswell2025conformal} proposed Equalized Set Size as a fairness standard, \citet{tasar2025} proposed the \emph{deferral gap}—the difference in deferral rates across groups—as a substantive fairness metric. However, they only instantiated the alternate process through random class assignment which decouples the assessment of fairness and prediction set properties from downstream task performance. In contrast, we incorporate downstream usage directly into our definition and measurement of fairness.

\vspace{-4pt}
\section{Methodology}\label{sec:methodology}

Group-conditional coverage (\Cref{eq:group-coverage}) is a natural \emph{procedural} fairness goal for CP, but coverage alone does not fully characterize fairness in downstream decision-making. In particular, prediction sets with equal coverage may differ systematically across groups in size or informativeness, leading to unequal benefits when these sets are used by humans or automated decision rules \cite{cresswell2025conformal}. Our ultimate goal is to promote substantive fairness in downstream decision-making by using prediction sets. To this end, we develop a robust evaluation framework for assessing the substantive fairness of CP methods. Using this pipeline, we study how procedural fairness notions (i.e., Equalized Coverage and Equalized Set Size) affect substantive fairness, how substantive fairness changes with different prediction-set characteristics, and which CP methods are most effective for supporting equitable downstream performance.

As noted by \citet{cresswell2025conformal}, equalizing coverage can increase set size disparities, which may in turn amplify substantive unfairness. Motivated by this finding, we focus our analytical work on approaches that prioritize equalizing set size rather than equalizing coverage. Concretely, we consider Label-Clustered CP, discussed in \Cref{sec: advanced-CP-variants}, which mitigates set size disparity by clustering similar datapoints regardless of group. In this section we theoretically justify why Label-Clustered CP reduces set size gaps, then present our evaluation framework for assessing substantive fairness in CP.

\subsection{Label-Clustered CP Reduces Set Size Disparity}
\label{sec:clustered-fairness}

Let \(A \in \mathcal{A}\) be the protected attribute and consider groups \(a, b \in \mathcal{A}\). The expected set size disparity between groups is
\begin{equation}
    \Delta_{a,b} = \left| \mathbb{E}\big[|\mathcal{C}(X)|\mid A{=}a\big] - \mathbb{E}\big[|\mathcal{C}(X)|\mid A{=}b\big] \right|.
    \label{eq:set-size-disparity-clusteredCP}
\end{equation}
For Label-Clustered CP we derive an upper bound that decomposes $\Delta_{a,b}$ into three interpretable components which can explain why label clustering often empirically yields smaller $\Delta_{a,b}$ than Marginal or Mondrian CP. This bound makes explicit how the number of clusters $K$ affects method-driven components in $\Delta_{a,b}$.
\begin{theorem}[Label-Clustered CP set size disparity bound]
    \label{Theorem3.1}
    Fix any label-clustering map $h: \mathcal{Y} \to [K]$ and let $\mathcal{Y}_k := \{y \in \mathcal{Y}: h(y) = k \}$. Consider a label-clustered conformal set predictor $\mathcal{C}$ that uses cluster-specific thresholds. For any $y \in \mathcal{Y}, ~ k \in [K]$, and group $a \in \mathcal{A}$, define
    \begin{align}
        \mu_{k,a} &:= \mathbb{E}[|\mathcal{C}(X)| \mid h(Y) = k, A = a], \\
        r_{y,a} &:= \mathbb{E}[|\mathcal{C}(X)| \mid Y = y, A = a], \quad \text{\emph{and}} \\
        \epsilon_{k,a} &:= \max_{y \in \mathcal{Y}_k} r_{y,a} - \min_{y \in \mathcal{Y}_k} r_{y,a}.
    \end{align}
    \vspace{-2pt}
    Then, for any two groups $a, b$,
    \begin{align}       \label{eq:theorem1}
        \nonumber\!\!\Delta_{a,b} &\le \!\!\!\!\!\!\!\!\!\!\!\!\!\!\! \underbrace{\max_{k = 1,\ldots,K} \epsilon_{k,a}}_{\textbf{(I): Intra-cluster label heterogeneity}} \!\!\!\!\!\!\!\!\!\!\!
        + \ \underbrace{\big(\max_{k=1,\ldots,K} \mu_{k,a} - \min_{k = 1,\ldots,K} \mu_{k,a} \big)}_{\textbf{(II): Cross-cluster spread}} \\
        & \quad\quad + \underbrace{\big|\sum_{y \in \mathcal{Y}} \mathbb{P}(Y = y \mid  A = b) (r_{y,a} - r_{y,b})\big|}_{\textbf{(III): Intra-label cross-group disparity}}.
        \vspace{-6pt}
    \end{align}
\end{theorem}
\looseness=-1The proof of \Cref{Theorem3.1} is given in \Cref{app:clustering}, with a detailed justification of why Label-Clustered CP reduces set size disparity across groups. Here, we provide interpretations and implications of \Cref{Theorem3.1}. The quantities we define each break down labels, groups, and clusters in different ways: $\mu_{k,a}$ represents the expected set size of a given group in a given cluster (across labels), while $r_{y,a}$ looks at label $y$ within group $a$ (across clusters). $\epsilon_{k,a}$ is the spread of set size across labels in cluster $k$, conditioned on one group.

\Cref{Theorem3.1} highlights three drivers of set size disparity:\\
(I) \textbf{Intra-cluster label heterogeneity}: If clusters bring together labels with similar difficulty levels, then labels within each cluster tend to have similar expected set sizes, making each $\epsilon_{k,a}$ small. This explains why $K = 1$ (Marginal CP) can yield a large $\epsilon_{1,a}$ -- all labels are forced into a single cluster, so intra-cluster label heterogeneity can be high. \\
(II) \textbf{Cross-cluster spread}: Consider the case of $K = |\mathcal{Y}|$ where each label forms a cluster. Although the intra-cluster label heterogeneity is minimized ($\epsilon_{k,a}=0$), conformal thresholds become unstable for rare labels, yielding large disparity in set size between clusters. With a proper choice of $K$, Label-Clustered CP can make the expected set size more comparable across clusters while controlling $\epsilon_{k,a}$.\\
(III) \textbf{Intra-label cross-group disparity} captures set size disparity between groups but within labels. Compared to Mondrian CP, Label-Clustered CP avoids inflating this component because it uses shared thresholds across protected groups (within each label-cluster) and pools calibration data across groups, reducing variance and preventing artificial group differences; see \Cref{app:clustering} for detailed comparison.

Overall, the bound in \Cref{eq:theorem1} highlights two clustering-dependent drivers: intra-cluster label heterogeneity and cross-cluster spread. In our experiments we study the behaviour of these terms individually.

\subsection{LLM-in-the-loop Substantive Fairness Evaluation}
\label{sec:LLM_evaluator}
Evaluating the downstream impact of CP on decision-making typically requires expensive and difficult-to-scale human trials. To address this, we propose an \textbf{LLM-in-the-loop} evaluation framework which offers key advantages: (i) LLMs exhibit approximate i.i.d.\ behavior across evaluations, avoiding human fatigue, learning effects, and temporal drift which all increase variance; (ii) they are adaptable to heterogeneous tasks across diverse data modalities; (iii) they allow for scalable, robust statistical evaluation. Most importantly, we show that our LLM-in-the-loop evaluator reproduces the same qualitative ordering of substantive fairness metrics observed in prior human-in-the-loop experiments \citep{cresswell2025conformal}—particularly that Mondrian CP exhibits higher disparate impact than Marginal (see \Cref{subsec: comparability-human-in-the-loop} and \Cref{app: compare-human-LLM}). 
In this section, we define substantive fairness within this framework and detail our estimation procedure using Generalized Estimating Equations (GEE).

\textbf{Substantive fairness as decision improvement.}
We ground our definition of substantive fairness in the concrete benefit provided to the decision-maker. Let $\oAcc(x, \hat y, \mathcal{C}(x))$ denote the decision accuracy achieved by the agent that predicts $\hat y$ given input $x$ and prediction set $\mathcal{C}(x)$. We define the \emph{group-specific improvement} as the expected lift in utility provided by the CP method relative to a control baseline where the agent acts without a prediction set (i.e., $\mathcal{C}(x) = \varnothing$). For a protected group $a \in \mathcal{A}$, this is given by
\vspace{-2pt}
\begin{equation*}\label{eq:improvement}
  \delta_{t, a}
  := \mathbb{E}\bigl[
    \oAcc(x, \hat y, \mathcal{C}_{t}(x))
    - \oAcc(x, \hat y, \varnothing) 
    \mid x \in G_a
  \bigr],
  \vspace{-2pt}
\end{equation*}
where $t$ stands for a CP method (the experimental ``treatment'').
For substantive fairness we require that the improvement be consistent across groups, i.e., there is no disparate impact. Hence, we quantify unfairness as the maximum disparity between groups:
\vspace{-2pt}
\begin{equation}\label{eq:disparity}
  \Delta_{t}
  :=
  \max_{a,b \in \mathcal A}
  \bigl|
    \delta_{t,a} - \delta_{t,b}
  \bigr|.
  \vspace{-2pt}
\end{equation}
A disparity $\Delta_{t} \approx 0$ indicates that the CP method $t$ improves downstream decision-making equally for all groups.

\textbf{Estimation of $\Delta_{t}$ via Generalized Estimating Equations.}
Directly computing empirical averages for \Cref{eq:disparity} is prone to confounding factors, such as systematic variations in task difficulty and the agent's willingness to rely on the provided sets. To obtain robust, statistically valid estimates of $\Delta_{t}$, and to take into account the correlation among predictions made for the same task under the assistance of different CP sets, we employ a logistic GEE regression.

For any data modality we assume access to a predictive model used to generate conformal sets via a CP algorithm $t$ and fixed score function. We then provide an LLM (or multi-modal foundation model) with a description of the task, a test datapoint \(x_j\), corresponding prediction set $\mathcal{C}_{t}(x_j)$, and a statement of the coverage guarantee (i.e., $1-\alpha$). The LLM is used to generate \(M\) independent predictions $\hat y_{jt}^m$, $m \in [M]$. Then we define $R_{jt} := \frac{1}{M} \sum_{m=1}^{M} \mathbf{1}\{\hat{y}_{jt}^{m} = y_j\}$, the proportion of correctly predicted responses for $x_j$ with CP method \(t\). We model the probability of correctness with key covariates and clustering by task to account for intra-instance correlations, for which GEEs are suitable \cite{liang1986longitudinal}. The regression model is specified as
\vspace{-2pt}
\begin{equation}
    \text{logit}(\mathbb{E}[R_{jt}]) \sim \text{treat}_{t} \times \text{group}_{j} + \text{diff}_{j} + \text{adoption}_{jt}.
    \label{GEE_LLM}
    \vspace{-2pt}
\end{equation}
Here, $\text{group}_{j}$ is the group that $x_j$ belongs to, $\text{treat}_{t} \times \text{group}_{j}$ captures the interaction of interest; $\text{diff}_{j}$ approximates task difficulty (using Marginal CP set size); and $\text{adoption}_{jt}$ measures the proportion of the agent's predictions adopted from the provided set $\mathcal{C}_{t}(x_j)$. This adoption covariate is crucial for generalizing results, as it accounts for varying levels of faith the agent places in the CP sets it is shown. Additional explanation of the design considerations behind this GEE model is given in \Cref{app: LLM_evaluator_sub1}.

\textbf{Quantifying fairness with maxROR.}
From the fitted GEE model, we compute the marginal probability $p_{t,a}$ of a correct response for treatment $t$ and group $a$. We convert these probabilities into odds ratios (ORs) relative to the control baseline:
\vspace{-6pt}
\begin{equation}
    \textbf{OR}_{t, a} := \frac{p_{t,a}/(1-p_{t,a})}{p_{\text{control},a}/(1-p_{\text{control},a})}.
    \label{ORs}
    \vspace{-2pt}
\end{equation}
$\textbf{OR}_{t, a} > 1$ indicates that treatment $t$ improves the LLM's accuracy for group $a$, compared to the control. To measure the disparity of improvement across groups $a$ and $b$, we compute the ratio of odds ratios (ROR) and take the maximum over all pairs. This yields our primary metric for substantive fairness, the \textbf{maxROR}:
\vspace{-2pt}
\begin{equation}
    \textbf{maxROR}_{t} := \max_{a,b \in \mathcal{A}} \left( \frac{\textbf{OR}_{t,a}}{\textbf{OR}_{t,b}} \right) - 1.
    \label{maxROR_def}
    \vspace{-2pt}
\end{equation}
\textbf{maxROR} is a principled way of measuring disparity (\Cref{eq:disparity}) that accounts for factors such as the difference in difficulty between groups. A $\textbf{maxROR}_{t}$ close to zero implies no downstream disparate impact from the use of CP method $t$, while a value of 0.10 (10\%), for example, indicates that one group benefited 10\% more than another. We primarily report \textbf{maxROR} \% values from our LLM-in-the-loop evaluator to quantify substantive fairness. Additional technical details on measuring fairness with the evaluator are given in \Cref{app: LLM_evaluator_sub2}.

\section{Experimental Setup}\label{sec:experiment}
\subsection{Experimental Design}
\label{sec:exp_design}

Our experiments investigate the interplay of procedural and substantive fairness notions in CP. We employ the LLM-in-the-loop evaluator described in \Cref{sec:LLM_evaluator} to answer four core research questions:

\textbf{RQ1 Alignment:} \textit{Does the LLM-in-the-loop evaluator faithfully reflect decision-making behaviors observed in humans?}\\
We validate that our LLM-in-the-loop evaluator aligns with prior human-subject studies, showing that it is a meaningful proxy, enabling scalable substantive fairness evaluation.

\textbf{RQ2 Substantive Benchmarking:} \textit{Which CP methods achieve substantive fairness, while still being useful?}\\
We evaluate several CP methods to determine which is most fair in downstream tasks (lowest $\textbf{maxROR}$), with overall utility of the prediction sets in mind.

\textbf{RQ3 Metric Correlation:} \textit{Do procedural fairness metrics correlate with substantive fairness?}\\
We analyze the relationship between procedural notions (Equalized Coverage, Equalized Set Size) and our substantive metric ($\textbf{maxROR}$) to determine if procedural metrics can be diagnostic indicators of downstream fairness.

\textbf{RQ4 Theoretical Verification:} \textit{Can our theoretical analysis of Label-Clustered CP be experimentally verified?}\\
We validate our theoretical analysis of set size disparity for Label-Clustered CP (\Cref{sec:clustered-fairness}) through ablations and numerical studies.

In implementing the LLM-in-the-loop evaluator, we choose $M$, the number of independent LLM predictions made for each task--treatment pair, to balance estimation stability with computational cost. \Cref{app: computational_cost} reports $M$, total LLM predictions, and API cost or runtime for each experiment, showing that the evaluator remains low-cost relative to human-subject studies: \citet{cresswell2024cphuman} and \citet{cresswell2025conformal} required approximately \pounds 1500 in participant payments for 30k--42.5k human responses. 
the low cost of our estimator, as low as  \$1 for 60k predictions, enables broader comparisons across datasets, modalities, and CP methods. 
Further details on datasets, CP score functions, hyperparameter tuning, and prompt engineering for the LLM-in-the-loop are provided in \Cref{app:add-exp}. Our code implementing CP methods and the LLM evaluator on these tasks is available at this \href{https://github.com/layer6ai-labs/llm-in-the-loop-conformal-fairness}{Github repo}.

\subsection{Tasks, Datasets, and Models}
We evaluate our methods on four prediction tasks spanning vision, text, audio, and tabular modalities, using open-access datasets commonly studied in algorithmic fairness. In all settings, CP is applied to the outputs of task-specific base models to construct prediction sets, and a foundation model uses those sets as decision aids on the downstream task.

\textbf{Image Classification.}
We use the FACET dataset \citep{gustafson2023facet}, predicting one of 20 occupation classes from images. Age (Younger, Middle, Older, Unknown) defines the protected groups. Prediction sets are generated using a zero-shot CLIP ViT-L/14 model as the base model \citep{dosovitskiy2021an, radford2021learning}, while Qwen2.5-VL-7B-Instruct is used as the LLM-in-the-loop for its vision-language capabilities \cite{bai2025qwen2}.

\textbf{Text Classification.}
We consider occupation prediction on the BiosBias dataset \citep{dearteaga2019biasinbios}, restricted to the 10 most frequent classes with binary gender as the sensitive attribute. A linear classifier trained on frozen BERT representations \citep{devlin2019bert} acts as the base model, and GPT-4o-mini as the LLM-in-the-loop \cite{openai2024gpt4omini}.

\textbf{Audio Emotion Recognition.}
We use the RAVDESS dataset \citep{livingstone2018ravdess} to classify audio clips into eight emotion classes, with binary gender as the group attribute. Base predictions are obtained from a fine-tuned wav2vec~2.0 model \citep{baevski2020wav2vec}, and GPT-4o-audio-preview acts as the LLM-in-the-loop for its audio capabilities \cite{openai2026gpt4oaudiopreview}.

\textbf{Tabular Prediction.}
We predict income brackets on the ACSIncome dataset from Folktables \citep{ding2021retiring}, using race (aggregated) as the group attribute. An XGBoost classifier \citep{chen2016xgboost} is the base model, while Qwen2.5-7B is the LLM-in-the-loop \cite{yang2024qwen2}.

\begin{table}[t]
\centering
\caption{Base Model and CP Method Metrics on $\mathcal{D}_{\text{test}}$.}
\label{tab:base-model-metrics}
\footnotesize
\setlength{\tabcolsep}{3pt}
\renewcommand{\arraystretch}{0.9}
\begin{tabular}{lcc|lcc}
\toprule
Task & Acc & $\Delta_{\text{Acc}}$ & CP Method & Cvg & Size \\
\midrule

\multirow{5}{*}{FACET}
  & \multirow{5}{*}{70.0} & \multirow{5}{*}{22.2}
  & Marginal        & 89.9 & 2.62 \\
  &                 & 
  & Mondrian        & 89.9 & 2.66 \\
  &                 & 
  & Label-Clustered & 89.1 & 2.92 \\
  &                 & 
  & Group-Clustered & 89.3 & 2.51 \\
  &                 & 
  & Backward        & 90.1 & 3.50 \\
\midrule

\multirow{5}{*}{BiosBias}
  & \multirow{5}{*}{78.9} & \multirow{5}{*}{2.70}
  & Marginal        & 89.5 & 1.68 \\
  &                 & 
  & Mondrian        & 90.0 & 1.80 \\
  &                 & 
  & Label-Clustered & 90.3 & 1.75 \\
  &                 & 
  & Group-Clustered & 90.2 & 1.75 \\
  &                 & 
  & Backward        & 91.5 & 2.50 \\
\midrule

\multirow{5}{*}{RAVDESS}
  & \multirow{5}{*}{70.3} & \multirow{5}{*}{6.11}
  & Marginal        & 88.3 & 1.89 \\
  &                 & 
  & Mondrian        & 87.5 & 1.86 \\
  &                 & 
  & Label-Clustered & 87.8 & 1.92 \\
  &                 & 
  & Group-Clustered & 87.5 & 1.90 \\
  &                 & 
  & Backward        & 91.9 & 2.48 \\
\midrule

\multirow{5}{*}{ACSIncome}
  & \multirow{5}{*}{31.0} & \multirow{5}{*}{5.71}
  & Marginal        & 89.8 & 5.35 \\
  &                 & 
  & Mondrian        & 89.5 & 7.16 \\
  &                 & 
  & Label-Clustered & 89.9 & 5.33 \\
  &                 & 
  & Group-Clustered & 89.8 & 5.37 \\
  &                 & 
  & Backward        & 92.3 & 6.50 \\
\bottomrule
\end{tabular}
\end{table}

\begin{table}[t]
\centering
\caption{Human vs. LLM Evaluator Comparison (maxROR \%).}
\label{tab:human-llm-compare}
\begin{small}
\setlength{\tabcolsep}{4pt}
\begin{tabular}{lcc|cc}
\toprule
& \multicolumn{2}{c|}{\textbf{Human-in-the-loop}} & \multicolumn{2}{c}{\textbf{LLM-in-the-loop}} \\
\cmidrule(lr){2-3}\cmidrule(lr){4-5}
\textbf{Dataset} & \textbf{Marginal} & \textbf{Mondrian}
& \textbf{Marginal} & \textbf{Mondrian} \\
\midrule
FACET    & 26  & 51  & 9.0 & 38  \\
BiosBias & 12  & 33  & 6.9 & 8.1 \\
RAVDESS  & 1.0 & 28  & 11  & 79  \\
\bottomrule
\end{tabular}
\end{small}
\vspace{-4pt}
\end{table}

\Cref{tab:base-model-metrics} shows a summary of base model and CP metrics on the test set, including accuracy and the maximum accuracy gap between groups, $\Delta_\text{Acc}$. Cvg is the empirical coverage, and Size is the average set size, where $1-\alpha=0.9$. Metrics are computed on a single calibration-test split, not averaged across many.

\section{Results}\label{sec:results}

\subsection{RQ1: Validation of LLM-in-the-loop Evaluation}
\label{subsec: comparability-human-in-the-loop}
First, we validate that our LLM-in-the-loop evaluator reproduces a key \emph{qualitative pattern} of substantive fairness reported in prior human-in-the-loop experiments. Due to the cost of human evaluation, \citet{cresswell2025conformal} only evaluated two CP methods, Marginal and Mondrian, on three datasets, FACET, BiosBias, and RAVDESS. They found that Mondrian CP induced greater disparate impact on downstream prediction accuracy compared to Marginal.

In \Cref{tab:human-llm-compare} we compare \textbf{maxROR} measurements between the human experiment data collected by \citet{cresswell2025conformal}, and with our LLM evaluator. LLM-in-the-loop consistently reproduces the qualitative \textbf{maxROR} ordering for Marginal and Mondrian CP, with Mondrian showing greater unfairness across all three datasets. This consistency supports the use of our LLM-in-the-loop evaluator as a scalable proxy for diagnosing \emph{substantive fairness trends and rankings across CP methods}. Our intention is not to replace formal human-subject studies, but to provide a low-cost first-pass evaluator for identifying comparative patterns in how CP design choices affect substantive fairness, thereby helping guide where more targeted human evaluations should be invested. See \Cref{app: compare-human-LLM} for further calibration details of the LLM-in-the-loop evaluator.

\begin{figure}[t]
  \begin{center}    \centerline{\includegraphics[width=\columnwidth]{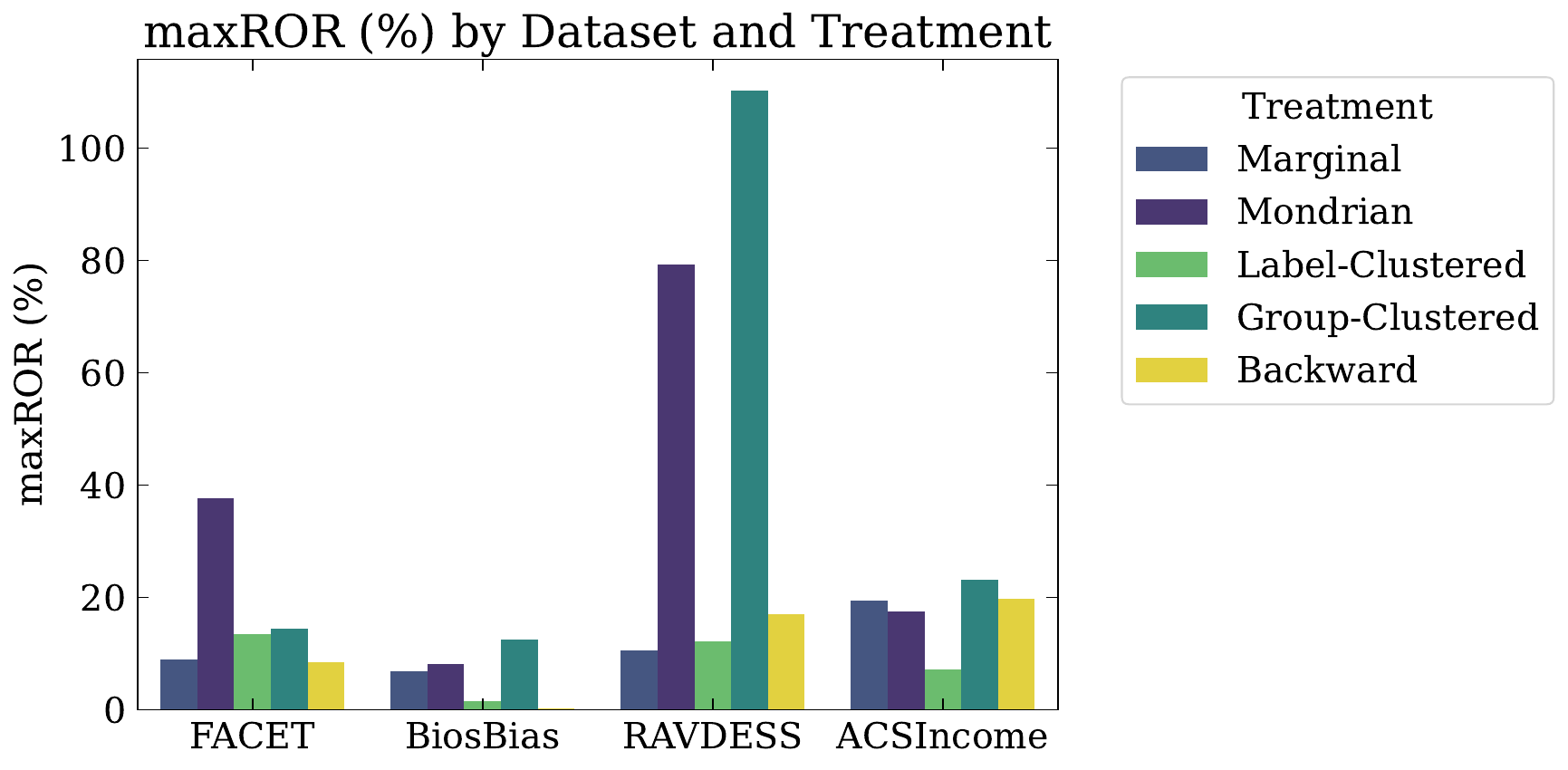}}
    \caption{
      \textbf{maxROR} (\%) of each CP method across four tasks. Lower is more substantively fair.
    }
    \label{fig:maxROR-by-datasets-treatments}
  \end{center}
  \vspace{-10pt}
\end{figure}

\subsection{RQ2: Substantive Fairness Benchmarking}
\label{subsec:rq2_results}

\begin{figure}[t]
  \begin{center}
\centerline{\includegraphics[width=0.99\columnwidth]{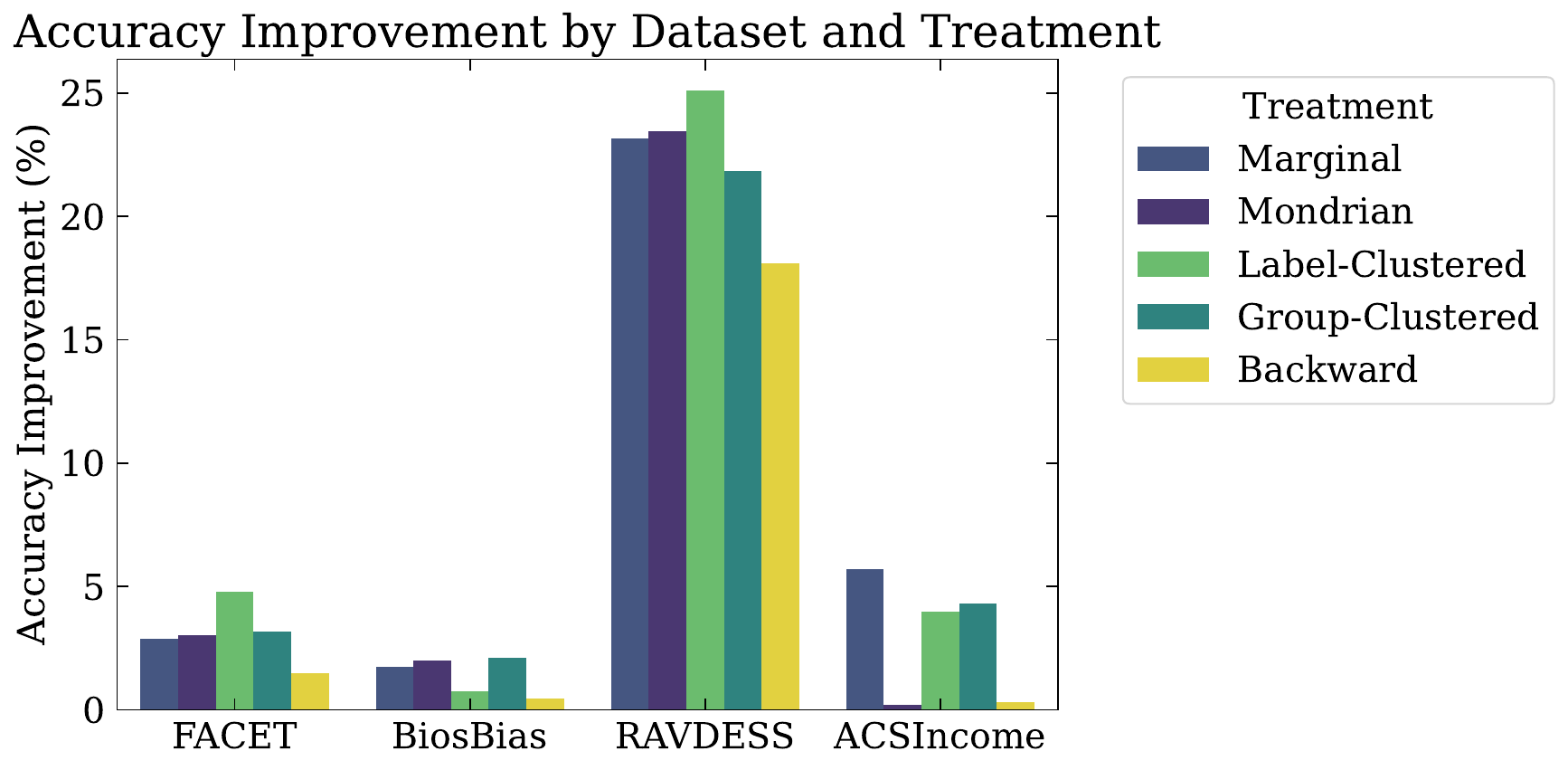}}
    \caption{
      Accuracy improvement (\%) relative to Control of each CP method, across four tasks. Higher is better.
    }
    \label{fig:accu-improv-by-datasets-treatments}
  \end{center}
  \vspace{-10pt}
\end{figure}

Having verified that our LLM-in-the-loop evaluator has similar qualitative behaviour to human decision-makers, we address the question: \textit{Which CP methods are most fair in downstream tasks?} We again measure the \textbf{maxROR} metric, but cover a wider variety of CP methods and datasets than prior research. In addition, sets should be helpful as a decision aid, so we also consider the overall accuracy of the LLM-in-the-loop on its task, relative to the control where no prediction set is provided. Our results in \Cref{fig:maxROR-by-datasets-treatments} identify \textbf{Backward and Label-Clustered CP as the most substantively fair} methods on average, but also that Label-Clustered CP is the more helpful of the two (\Cref{fig:accu-improv-by-datasets-treatments}).

On FACET and BiosBias, Backward CP achieved the lowest \textbf{maxROR}. However, Backward CP suffers from larger set size than other CP methods, partially due to its conservative empirical coverage (\Cref{tab:base-model-metrics}), and hence also is less helpful for the task as seen by lower accuracy improvement in all comparisons. Meanwhile, Label-Clustered CP offers a robust balance between efficiency and substantive fairness. Its \textbf{maxROR} was considerably lower than Backward on RAVDESS and ACSIncome, with much greater helpfulness to the decision maker. 

In contrast, Mondrian and Group-Clustered CP are never optimal in terms of \textbf{maxROR} and induced by far the most unfair outcomes for FACET and RAVDESS. For BiosBias they lead to the highest accuracy improvements, but clearly these improvements are not shared equally across groups in the data. Mondrian and Group-Clustered CP both pursue coverage parity through group-dependent calibration (at the level of individual groups for Mondrian CP and clusters of groups for Group-Clustered CP), but this coverage-oriented objective can be in tension with downstream equity.

As a robustness check, \Cref{app: group-adoption-interaction} shows that adding an adoption \(\times\) group interaction to the GEE leaves the \textbf{maxROR} rankings among CP methods unchanged, supporting that these downstream-fairness comparisons are not driven by group-varying adoption effects. Extended details on these experiments are in \Cref{app:subsec-evaluator-result}, including an ablation with a different LLM on BiosBias.

\subsection{RQ3: Procedural and Substantive Correlations}\label{sec:proc-sub-corr}

\begin{figure}[t]
  \centering
  \begin{subfigure}[t]{0.49\linewidth}
    \centering
    \includegraphics[width=\linewidth, trim={8 8 8 7}, clip]{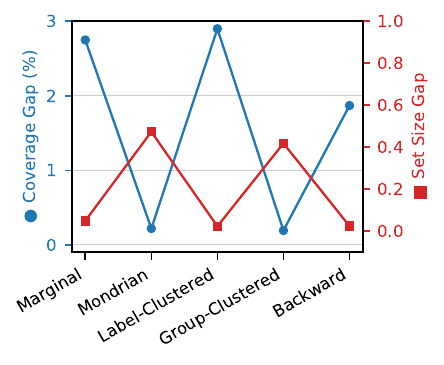}
    \caption{BiosBias}
    \label{fig: gap-bios-only}
  \end{subfigure}
  \hfill
  \begin{subfigure}[t]{0.49\linewidth}
    \centering
    \includegraphics[width=\linewidth, trim={8 8 8 7}, clip]{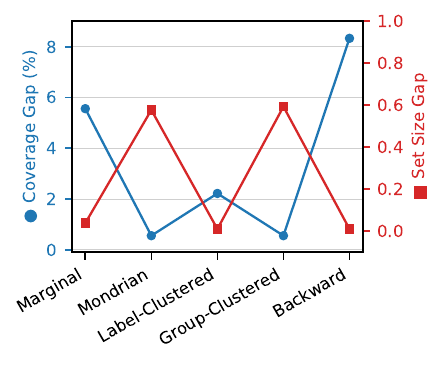}
    \caption{RAVDESS}
    \label{fig: gap-ravdess-only}
  \end{subfigure}
  \caption{ Coverage gap (blue dots, left axis) and set size gap (red squares, right axis) across CP methods. The two procedural fairness metrics are in direct tension. Corresponding plots for FACET and ACSIncome are in Appendix~\ref{app:cov-gap-plot}.
  }
  \label{fig:gap-four-separate}
\end{figure}

Next, we ask: \textit{Which procedural fairness metric correlates most strongly with substantive fairness?} Traditionally, researchers have focused on minimizing the coverage gap \cite{romano2020with}, with more recent studies recommending set size gap as an alternative \cite{cresswell2025conformal}. 
In the experiments, we report empirical plug-in estimates of the corresponding population-level quantities:
\begin{align}
    &\text{coverage gap} = \max_{a,b \in \mathcal{A}} |\mathbb{P}(y \in \mathcal{C}(x) \mid x \in G_a) \nonumber\\
    &\hspace{100pt} - \mathbb{P}(y \in \mathcal{C}(x) \mid x \in G_b)|, \\
    &\text{set size gap} = \max_{a,b \in \mathcal{A}} |\mathbb{E}[|\mathcal{C}(x)| \mid x \in G_a] \nonumber\\
    &\hspace{100pt} - \mathbb{E}[|\mathcal{C}(x)| \mid x \in G_b]|.
\end{align}
\Cref{fig:gap-four-separate} demonstrates that these two procedural metrics are in diametric opposition; CP methods optimize one at the expense of the other. Understanding which of these metrics correlates with substantive fairness enables its use as an early diagnostic signal of unfairness before expensive downstream deployments are undertaken.

\begin{figure}[t]
  \begin{center}
    \centerline{\includegraphics[width=\columnwidth]{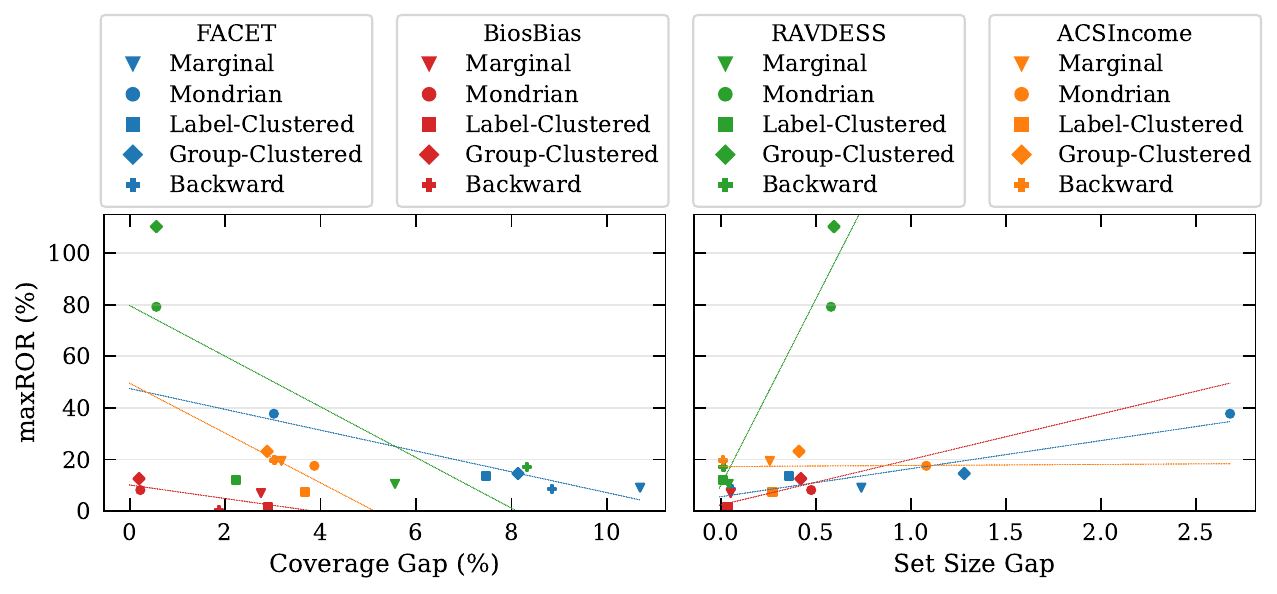}}
    \caption{
      \textbf{maxROR} (\%) compared to the coverage gap (Left) and set size gap (Right) between groups, across CP methods and datasets. Regression lines are fitted for each dataset individually to show trends. 
    }
    \label{fig:cov-size-gap-vs-maxROR}
  \end{center}
  \vskip -0.3in
\end{figure}

In \Cref{fig:cov-size-gap-vs-maxROR} we plot the procedural fairness metrics against our substantive metric, \textbf{maxROR} for each dataset and CP method. Since the metrics are on different scales between datasets, we also plot individual regression lines for the data from each dataset. We clearly see that all four regressions for the coverage gap have negative slope; decreasing the coverage gap (equalizing coverage between groups) leads to higher \textbf{maxROR} (greater unfairness). The set size gap data on the other hand shows positive slopes, such that decreasing it (equalizing set size) also decreases \textbf{maxROR}. From these consistent trends across datasets it is evident that Equalized Set Size as a procedural fairness notion is also aligned with substantive fairness goals of downstream equity, whereas Equalized Coverage is actively inequitable.

\subsection{RQ4: Effect of Label-Clustered CP on Set Size Gap}
\label{sec:exp-clustered-fairness}

Knowing that set size gap is a relevant predictor of downstream fairness, we revisit our theoretical analysis from \Cref{sec:clustered-fairness} and verify its insights experimentally. In \Cref{Theorem3.1} we decomposed the set size disparity $\Delta_{a,b}$ (\Cref{eq:disparity}) into three components that are affected by label clusters and groups in the data. We now examine the behaviour of the bound overall, the interplay of the three terms, and the effect of Label-Clustering CP's main hyperparameter—the number of clusters $K$.

\begin{figure}[t]
  \centering
  \begin{subfigure}[t]{0.49\columnwidth}
    \centering
    \includegraphics[width=\linewidth,height=0.85\linewidth]{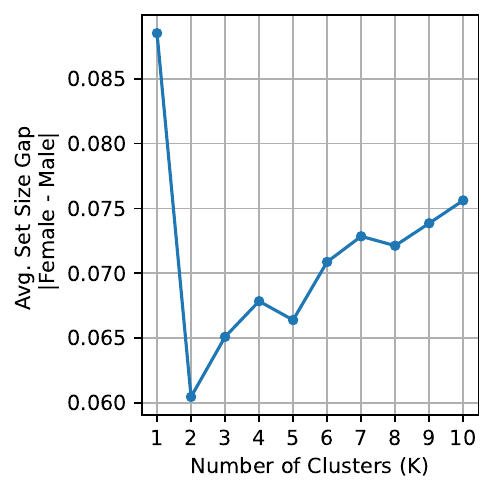}
    \caption{BiosBias}
  \end{subfigure}\hfill
  \begin{subfigure}[t]{0.49\columnwidth}
    \centering
    \includegraphics[width=\linewidth,height=0.85\linewidth]{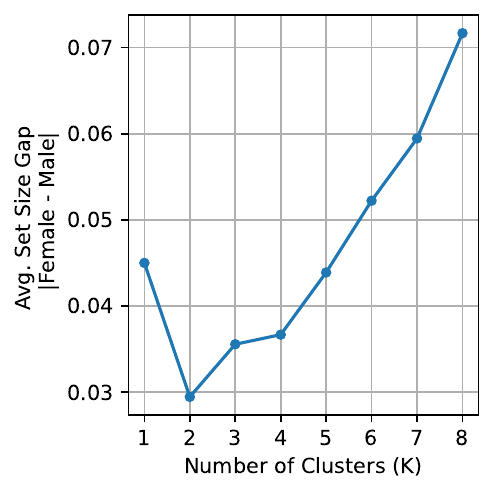}
    \caption{RAVDESS}
  \end{subfigure}
  \caption{Average prediction set size gap between Female and Male on the BiosBias and RAVDESS datasets over 10 random splits. The maximum standard error of the average set size gap is 0.016 in (a) and 0.010 in (b).}
  \label{fig:clusters-gap}
  \vspace{-8pt}
\end{figure}

First, using the BiosBias and RAVDESS datasets in \Cref{fig:clusters-gap}, we vary the number of clusters from $K=1$ (Marginal CP) up to the total number of classes $m$, and compute the average set size gap for Label-Clustered CP over 10 random calibration--test splits. The observed relationship between \(K\) and $\Delta_{a, b}$ exhibits a clear V-shaped pattern with a minimum of set size disparity at $K=2$, and sharp increase for $K=1$. This connects back to Label-Clustered CP's ability to reduce substantive unfairness compared to Marginal (\Cref{fig:maxROR-by-datasets-treatments}) by reducing the set size gap; clustering combines datapoints with similar labels regardless of group such that model confidence can be calibrated accurately within the clusters.

\Cref{app: maxROR-K-sensitivity} further examines whether the $K$-sensitivity of the set-size gap carries over to downstream fairness. As a single-split diagnostic on BiosBias, the \textbf{maxROR} trend for Label-Clustered CP qualitatively mirrors the set-size-gap pattern: $K = 1$ performs poorly, several intermediate values of $K$ substantially reduce \textbf{maxROR}, and larger $K$ values can increase the metric again. However, the value of $K$ minimizing \textbf{maxROR} need not coincide exactly with the value of $K$ minimizing set-size gap because \textbf{maxROR} reflects downstream decision behavior depending on factors beyond set-size gap alone.

\begin{figure}[t]
  \centering
  \begin{subfigure}[h]{0.49\columnwidth}
    \centering
    \includegraphics[width=\linewidth,height=0.9\linewidth]{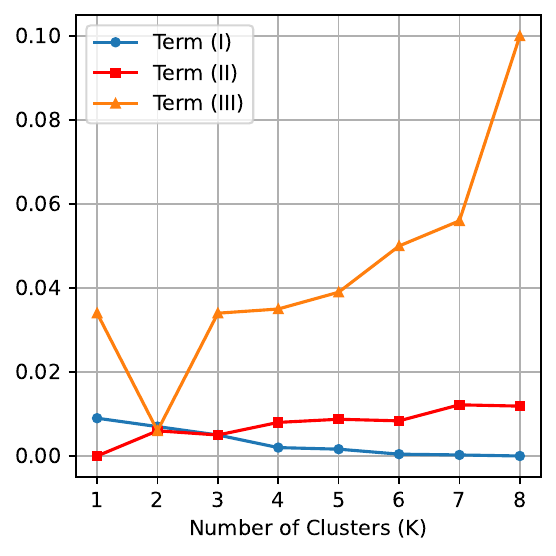}
    \caption{All three terms}
  \end{subfigure}\hfill
  \begin{subfigure}[h]{0.49\columnwidth}
    \centering
    \includegraphics[width=\linewidth,height=0.9\linewidth]{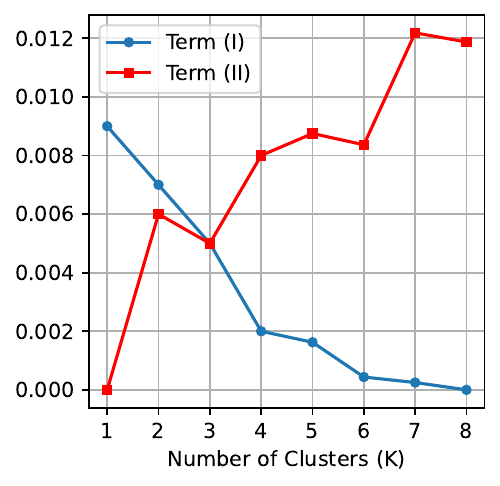}
    \caption{Terms I and II}
  \end{subfigure}
  \caption{Numerical computation of the three terms in \Cref{eq:theorem1} vs. number of clusters $K$ on RAVDESS.}
  \label{fig:theoretical-bound}
  \vspace{-6pt}
\end{figure}

We show more detail on the behaviour of the three terms separately for RAVDESS in \Cref{fig:theoretical-bound}. While term III is generally the largest and gives rise to the distinctive V shape with a minimum at $K=2$, the other terms' behavior closely aligns with the discussion in \Cref{sec:clustered-fairness}. When \(K=1\) (Marginal CP), term~II, the cross-cluster spread of expected set sizes, is of course minimized, but term~I remains large due to substantial label heterogeneity within the single cluster. Increasing \(K\) reduces intra-cluster label heterogeneity I, while the cross-cluster spread II increases as calibration becomes less stable for small clusters. 

\begin{figure}[t]
  \centering    \includegraphics[width=0.49\linewidth,height=0.49\linewidth]{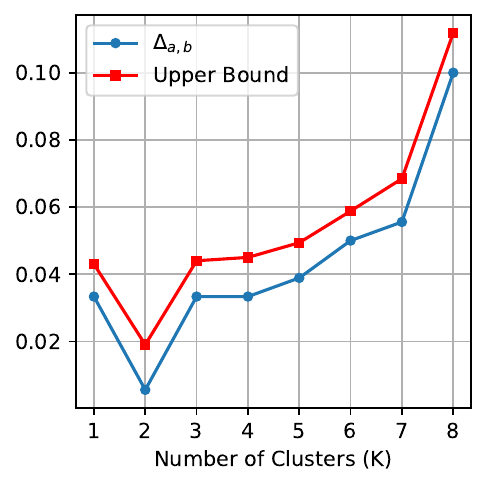}
  \caption{Numerical computation of $\Delta_{a,b}$ vs. the upper bound from \Cref{eq:theorem1} on RAVDESS with Label-Clustered CP. The bound is reasonably tight and faithfully reflects the shape of $\Delta_{a,b}$ as $K$ is varied.}
  \label{fig:tightness}
  \vspace{-8pt}
\end{figure}

Finally, in \Cref{fig:tightness} we demonstrate the tightness of the bound on RAVDESS by numerically computing $\Delta_{a,b}$ vs. the sum of all three terms. The bound is reasonably tight, demonstrating a regular and small bias, allowing us to rely on the interpretations of the three individual terms.

Overall, these experiments validate the theoretical statement that, with a carefully chosen number of clusters, Label-Clustered CP can more effectively balance label adaptivity and calibration stability to improve set size disparity, which correlates strongly with substantive fairness. Our decomposition gives insight into \emph{why} Label-Clustered CP is able to achieve better procedural fairness (Equalized Set Size) than other CP methods (\Cref{fig:gap-four-separate}), and by extension better substantive fairness (\Cref{fig:maxROR-by-datasets-treatments}).

\vspace{-4pt}
\subsection{Practical Guidelines}
\label{subsec:practical_guidelines}

Based on these theoretical and empirical findings, we offer the following recommendations for deploying CP in fairness-critical decision pipelines:

\textbf{Evaluate both procedural and substantive fairness:} Equality and equity are both noble pursuits, but can sometimes be at odds (\Cref{fig:cov-size-gap-vs-maxROR} left). Determine which criteria cannot be compromised on prior to building CP systems, and evaluate metrics reflecting both notions throughout development.

\textbf{Prioritize minimizing set size gaps:} Do not optimize for Equalized Coverage in isolation. Equalized Set Size correlates strongly with substantive fairness, whereas equalizing coverage tends to increase \textbf{maxROR} (\Cref{fig:cov-size-gap-vs-maxROR}). For Label-Clustered CP, the $K$-sensitivity analysis in \Cref{sec:exp-clustered-fairness} suggests first using the set-size gap as a low-cost diagnostic to narrow $K$ to a small set of promising candidates, then selecting among them using a downstream fairness metric when such data are available.
    
\textbf{Avoid demographic conditioning:} Methods that explicitly condition on the protected group (e.g., Mondrian, Group-Clustered) tend to amplify set size disparity to satisfy coverage constraints. Instead, conditioning on labels (Label-Clustered CP) calibrates thresholds within clusters of similar difficulty, which naturally balances sets without baking in group biases.

\vspace{-4pt}
\section{Conclusion}\label{sec:discussion}

In this work, we moved beyond the view of conformal prediction as a standalone procedure, and evaluated its impact on \emph{substantive fairness} in downstream decision-making. By designing a scalable LLM-in-the-loop evaluator, we demonstrated that the standard procedural fairness notion, Equalized Coverage, often fails to translate into equitable outcomes. Instead, our findings highlight that equalizing set size is the critical procedural lever that correlates with substantive fairness, with Label-Clustered CP achieving the most effective balance of utility and equity.

A promising avenue for future work is to deepen the causal analysis of these interactions. While our current study identifies strong correlations, explicitly controlling the \textbf{adoption rate} of the LLM evaluator (systematically varying how much the agent relies on the prediction set) would allow for a rigorous isolation of the causal effects of set properties on substantive decision outcomes.

\section*{Acknowledgements}

This work received partial support through a Mitacs Accelerate program co-funded by Mitacs and Layer 6 AI at TD. The work of M.A. is supported by NSERC grant RGPIN 2024-05640.

\section*{Impact Statement}

In this work we study the interactions between uncertainty quantification methods and fairness, pointing out a gap in the way fairness has been quantified in previous studies. The impact of our work is to raise awareness on issues of equity in machine learning, and as such we do not expect negative societal impacts to arise.

\clearpage

\bibliography{references}
\bibliographystyle{icml2026}

% APPENDIX
\newpage
\appendix
\renewcommand{\theequation}{A\arabic{equation}}
\setcounter{equation}{0}
\onecolumn
\addcontentsline{toc}{section}{Appendix} 
\part{Appendix} 
\parttoc 
\bigskip

\section{Justification that Label-Clustered CP Reduces Set Size Disparity}\label[appsec]{app:clustering}

In this appendix, we provide justification on how Label-Clustered CP reduces set size gap between protected \emph{groups}, 
even though it does calibration based on clusters of \emph{labels}. Let $A$ be the random variable of the protected groups, that is, the possible values of $A$ are all protected groups. Consider any two protected groups, say group $a$ and group $b$, we want to show that the gap between $\mathbb{E} [|\mathcal{C}(X)| \mid A = a]$ and $\mathbb{E} [|\mathcal{C}(X)| \mid A = b]$ can be reduced by implementing Label-Clustered CP, improving fairness between group $a$ and group $b$ in terms of prediction set size, especially compared to Mondrian CP. In what follows, $\mathcal{C}(X)$ denotes the prediction set constructed from Label-Clustered CP, and $|\mathcal{C}(X)|$ is the cardinality of the prediction set $\mathcal{C}(X)$.

\noindent \textbf{\large Proof of \Cref{Theorem3.1}}:
According to the law of total expectation, we have
\begin{align*}
    \mathbb{E}[|\mathcal{C}(X)| \mid A = a] &= \sum_{y \in \mathcal{Y}} \mathbb{P}(Y = y | A = a) ~\mathbb{E}[|\mathcal{C}(X)| \mid Y = y, A = a]\\
    \mathbb{E}[|\mathcal{C}(X)| \mid A = b] &= \sum_{y \in \mathcal{Y}} \mathbb{P}(Y = y | A = b) ~\mathbb{E}[|\mathcal{C}(X)| \mid Y = y, A = b]
\end{align*}
Let $\Delta_{a,b} = |\mathbb{E}[|\mathcal{C}(X)| \mid A = a] - \mathbb{E}[|\mathcal{C}(X)| \mid A = b]|$ be the disparity in expected set size, so we have
\begin{align}
    \Delta_{a,b} &= \Big| \sum_{y \in \mathcal{Y}} \mathbb{P}(Y = y | A = a) ~\mathbb{E}[|\mathcal{C}(X)| \mid Y = y, A = a] - \sum_{y \in \mathcal{Y}} \mathbb{P}(Y = y | A = b) ~\mathbb{E}[|\mathcal{C}(X)| \mid Y = y, A = b] \nonumber\\
    &\quad \quad + \sum_{y \in \mathcal{Y}} \mathbb{P}(Y = y | A = a) ~\mathbb{E}[|\mathcal{C}(X)| \mid Y = y, A = b] - \sum_{y \in \mathcal{Y}} \mathbb{P}(Y = y | A = a) ~\mathbb{E}[|\mathcal{C}(X)| \mid Y = y, A = b] \Big| \nonumber\\
    &= \Big| \underbrace{\sum_{y \in \mathcal{Y}} [\mathbb{P}(Y = y| A =a) - \mathbb{P}(Y = y| A =b)] ~\mathbb{E}[|\mathcal{C}(X)| \mid Y = y, A = b]}_{\text{(I)}} \label{eqI}\\
    &\quad \quad + \underbrace{\sum_{y \in \mathcal{Y}} \mathbb{P}(Y = y| A = a) \{\mathbb{E}[|\mathcal{C}(X)| \mid Y = y, A = a] - \mathbb{E}[|\mathcal{C}(X)| \mid Y = y, A = b] \}}_{\text{(II)}} \Big|
    \label{eqII}
\end{align}
We analyze term (I) in \Cref{eqI} and term (II) in \Cref{eqII} individually to investigate how Label-Clustered CP controls the gap of expected set size between group $a$ and group $b$. Recall the clustering function $h: \mathcal{Y} \to \{1, \ldots, K\}$ that maps each class $y \in \mathcal{Y}$ to one of the $K$ clusters based on score distributions of the labels. Throughout the arguments in this appendix, we treat $h$ as fixed by conditioning on the portion of data (clustering data set) used to learn it. Let $\mathcal{Y}_k := \{y \in \mathcal{Y}: h(y) = k \}$ be the set of labels in the $k$-th cluster for each $k = 1,\ldots,K$. 
Then, 
\begin{align*}
    (I) &= \sum_{y \in \mathcal{Y}} [\mathbb{P}(Y = y| A =a) - \mathbb{P}(Y = y| A =b)] ~\mathbb{E}[|\mathcal{C}(X)| \mid Y = y, A = b] \\
    &= \sum_{k=1}^{K} \sum_{y \in \mathcal{Y}_k} [\mathbb{P}(Y = y| A =a) - \mathbb{P}(Y = y| A =b)] ~\mathbb{E}[|\mathcal{C}(X)| \mid Y = y, A = b] \\
    &= \sum_{k=1}^{K} \sum_{y \in \mathcal{Y}_k} [\mathbb{P}(Y = y, h(Y) = h(y) | A =a) - \mathbb{P}(Y = y, h(Y) = h(y)| A =b)] \cdot \mathbb{E}[|\mathcal{C}(X)| \mid Y = y, A = b] \\
    &\hspace{100pt} (\text{because } \{Y = y\} \Rightarrow \{h(Y) = h(y)\}, \text{we have } \{Y = y\} \cap \{h(Y) = h(y)\} = \{Y = y\}) \\
    &= \sum_{k=1}^{K} \sum_{y \in \mathcal{Y}_k} [\mathbb{P}(Y = y, h(Y) = k | A =a) - \mathbb{P}(Y = y, h(Y) = k| A =b)] \cdot \mathbb{E}[|\mathcal{C}(X)| \mid Y = y, A = b]\\
    &= \sum_{k=1}^{K} \mathbb{P}(h(Y) = k | A = a) \sum_{y \in \mathcal{Y}_k} \mathbb{P}(Y = y | h(Y) = k, A = a) ~\mathbb{E}[|\mathcal{C}(X)| \mid Y = y, A = b] \\
    &\quad \quad \quad - \sum_{k=1}^{K} \mathbb{P}(h(Y) = k | A = b) \sum_{y \in \mathcal{Y}_k} \mathbb{P}(Y = y | h(Y) = k, A = b) ~\mathbb{E}[|\mathcal{C}(X)| \mid Y = y, A = b] 
\end{align*}
Now, define
\begin{align*}
    (I)_1 &:= \sum_{k=1}^{K} \mathbb{P}(h(Y) = k | A = a) \sum_{y \in \mathcal{Y}_k} \mathbb{P}(Y = y | h(Y) = k, A = a) ~\mathbb{E}[|\mathcal{C}(X)| \mid Y = y, A = b] \\
    &\quad \quad \quad - \sum_{k=1}^{K} \mathbb{P}(h(Y) = k | A = a) \sum_{y \in \mathcal{Y}_k} \mathbb{P}(Y = y | h(Y) = k, A = b) ~\mathbb{E}[|\mathcal{C}(X)| \mid Y = y, A = b]
\end{align*}
and
\begin{align*}
    (I)_2 &:= \sum_{k=1}^{K} \mathbb{P}(h(Y) = k | A = a) \sum_{y \in \mathcal{Y}_k} \mathbb{P}(Y = y | h(Y) = k, A = b) ~\mathbb{E}[|\mathcal{C}(X)| \mid Y = y, A = b] \\
    &\quad \quad \quad - \sum_{k=1}^{K} \mathbb{P}(h(Y) = k | A = b) \sum_{y \in \mathcal{Y}_k} \mathbb{P}(Y = y | h(Y) = k, A = b) ~\mathbb{E}[|\mathcal{C}(X)| \mid Y = y, A = b]
\end{align*}
Then, we have $(I) = (I)_1 + (I)_2$. We analyze $(I)_1$ and $(I)_2$ separately as follows.  

\noindent \textbf{1. Bound $|(I)_1|$:}

We can rewrite $(I)_1$ as
\begin{align}
    (I)_1 = \sum_{k=1}^{K} \mathbb{P}(h(Y) = k | A = a) \sum_{y \in \mathcal{Y}_k} \big[\mathbb{P}(Y = y | h(Y) = k, A = a) - \mathbb{P}(Y = y | h(Y) = k, A = b) \big] ~\mathbb{E}[|\mathcal{C}(X)| \mid Y = y, A = b]
    \label{I1}
\end{align}
First, fix a cluster $k$. Since $\sum_{y \in \mathcal{Y}_k} \mathbb{P}(Y = y | h(Y) = k, A = a) - \mathbb{P}(Y = y | h(Y) = k, A = b) = 0$, for any constant $c$, we have
\begin{align*}
    &\quad \sum_{y \in \mathcal{Y}_k} \big[\mathbb{P}(Y = y | h(Y) = k, A = a) - \mathbb{P}(Y = y | h(Y) = k, A = b) \big] ~\mathbb{E}[|\mathcal{C}(X)| \mid Y = y, A = b] \\
    &= \sum_{y \in \mathcal{Y}_k} \big[\mathbb{P}(Y = y | h(Y) = k, A = a) - \mathbb{P}(Y = y | h(Y) = k, A = b) \big] ~\left(\mathbb{E}[|\mathcal{C}(X)| \mid Y = y, A = b] - c \right)
\end{align*}
Applying this logic, if we define $c_k := \frac{1}{2} \left(\max_{y \in \mathcal{Y}_k} \mathbb{E}[|C(X)| \mid Y = y, A = b] + \min_{y \in \mathcal{Y}_k} \mathbb{E}[|C(X)| \mid Y = y, A = b] \right)$, then we have
\begin{align}
    &\quad \left| \sum_{y \in \mathcal{Y}_k} \big[\mathbb{P}(Y = y | h(Y) = k, A = a) - \mathbb{P}(Y = y | h(Y) = k, A = b) \big] ~\mathbb{E}[|\mathcal{C}(X)| \mid Y = y, A = b] \right| \nonumber\\
    &= \left| \sum_{y \in \mathcal{Y}_k} \big[\mathbb{P}(Y = y | h(Y) = k, A = a) - \mathbb{P}(Y = y | h(Y) = k, A = b) \big] ~\left(\mathbb{E}[|\mathcal{C}(X)| \mid Y = y, A = b] - c_k \right) \right| \nonumber\\
    &\le \Vert \mathbf{p}_{k,a} - \mathbf{p}_{k,b} \Vert_1 \cdot \max_{y \in \mathcal{Y}_k} \left|\mathbb{E}[|\mathcal{C}(X)| \mid Y = y, A = b] - c_k \right| ~~ \text{ by Hölder's inequality} \nonumber\\
    &= \Vert \mathbf{p}_{k,a} - \mathbf{p}_{k,b} \Vert_1 \cdot \frac{1}{2} \left(\max_{y \in \mathcal{Y}_k} \mathbb{E}[|C(X)| \mid Y = y, A = b] - \min_{y \in \mathcal{Y}_k} \mathbb{E}[|C(X)| \mid Y = y, A = b] \right) \nonumber\\
    &\le \frac{1}{2}(\Vert \mathbf{p}_{k,a} \Vert_1 + \Vert \mathbf{p}_{k,b} \Vert_1) \left(\max_{y \in \mathcal{Y}_k} \mathbb{E}[|C(X)| \mid Y = y, A = b] - \min_{y \in \mathcal{Y}_k} \mathbb{E}[|C(X)| \mid Y = y, A = b] \right) ~~\text{ by triangle inequality} \nonumber\\
    &= \frac{1}{2} \cdot 2 \left(\max_{y \in \mathcal{Y}_k} \mathbb{E}[|C(X)| \mid Y = y, A = b] - \min_{y \in \mathcal{Y}_k} \mathbb{E}[|C(X)| \mid Y = y, A = b] \right) \nonumber\\
    &= \max_{y \in \mathcal{Y}_k} \mathbb{E}[|C(X)| \mid Y = y, A = b] - \min_{y \in \mathcal{Y}_k} \mathbb{E}[|C(X)| \mid Y = y, A = b],
    \label{App:boundI1_1}
\end{align}
where $\mathbf{p}_{k,a}$ is the probability vector with each component being $\mathbb{P}(Y = y | h(Y) = k, A = a)$ for $y \in \mathcal{Y}_k$; similarly, $\mathbf{p}_{k,b}$ is the probability vector with each component being $\mathbb{P}(Y = y | h(Y) = k, A = b)$ for $y \in \mathcal{Y}_k$, so $\mathbf{p}_{k,a}$ and $\mathbf{p}_{k,b}$ are vectors of probabilities that depend on data distribution. 

For simplicity, for every $k \in \{1,\ldots,K \}$, define 
\begin{equation*}
    \epsilon_{k,b} := \max_{y \in \mathcal{Y}_k} \mathbb{E}[|C(X)| \mid Y = y, A = b] - \min_{y \in \mathcal{Y}_k} \mathbb{E}[|C(X)| \mid Y = y, A = b].
\end{equation*}
That is, $\epsilon_{k,b}$ is the spread of expected set size over the labels in cluster $k$, conditioning on group $b$. Then, we have
\begin{align*}
    |(I)_1| & \le \sum_{k=1}^{K} \mathbb{P}(h(Y) = k | A = a) \left| \sum_{y \in \mathcal{Y}_k} \big[\mathbb{P}(Y = y | h(Y) = k, A = a) - \mathbb{P}(Y = y | h(Y) = k, A = b) \big] ~\mathbb{E}[|\mathcal{C}(X)| \mid Y = y, A = b] \right| \\
    &\le \sum_{k=1}^{K} \mathbb{P}(h(Y) = k | A = a) \left(\max_{y \in \mathcal{Y}_k} \mathbb{E}[|C(X)| \mid Y = y, A = b] - \min_{y \in \mathcal{Y}_k} \mathbb{E}[|C(X)| \mid Y = y, A = b] \right) ~~\text{ by } \Cref{App:boundI1_1} \\
    &= \sum_{k=1}^{K} \epsilon_{k,b} ~\mathbb{P}(h(Y) = k | A = a) \\ 
    &\le \max_{k = 1,\ldots,K} \epsilon_{k,b}  \cdot \sum_{k=1}^{K}\mathbb{P}(h(Y) = k | A = a)\\
    &= \max_{k = 1,\ldots,K} \epsilon_{k,b} 
\end{align*}
The above derivation shows that $|(I)_1|$ is upper bounded by the maximum intra-cluster expected set size difference across labels. When $K = 1$, the Label-Clustered CP reduces to a special case, Marginal CP. In this case, all labels fall into one cluster, and the upper bound for $|(I)_1|$ becomes $\epsilon_{1,b}$. Because Marginal CP does not learn the cluster assignments according to score distributions of labels, instead forcing all labels into a single cluster, the $\epsilon_{1,b}$ term can be large due to label heterogeneity. In contrast, when $K = |\mathcal{Y}|$, each label forms a cluster, and $\epsilon_{k,b} = 0$ for all $k = 1,\ldots,K$, yielding $(I)_1 = 0$. Label-Clustered CP clusters labels using similarity of score distribution, which can be viewed as a proxy for label difficulty. Therefore, with a proper choice of $K$ and the associated cluster assignment, the labels in each cluster have similar score distributions and difficulty levels, yielding small $\epsilon_{k,b}$ for each $k$, which gives a tight upper bound for $|(I)_1|$. Therefore, Label-Clustered CP can effectively control $|(I)_1|$ by limiting label heterogeneity within each cluster.     

\noindent \textbf{2. Bound $|(I)_2|$:}

After combining the common terms, we get
\begin{align}
    (I)_2 = \sum_{k=1}^{K} \big(\mathbb{P}(h(Y) = k | A = a) - \mathbb{P}(h(Y) = k | A = b) \big) \sum_{y \in \mathcal{Y}_k} \mathbb{P}(Y = y | h(Y) = k, A = b) ~\mathbb{E}[|\mathcal{C}(X)| \mid Y = y, A = b]
    \label{boundI2}
\end{align}
We can simplify the $\sum_{y \in \mathcal{Y}_k} \mathbb{P}(Y = y | h(Y) = k, A = b) ~\mathbb{E}[|\mathcal{C}(X)| \mid Y = y, A = b]$ in \Cref{boundI2} as follows.
\small\begin{align*}
    \sum_{y \in \mathcal{Y}_k} \mathbb{P}(Y = y | h(Y) = k, A = b) \mathbb{E}[|\mathcal{C}(X)| \mid Y = y, A = b] &= \sum_{y \in \mathcal{Y}_k} \mathbb{P}(Y = y | h(Y) = k, A = b) ~\mathbb{E}[|\mathcal{C}(X)| \mid Y = y, h(Y) = k, A = b] \\
    &\hspace{95pt} (\text{because } Y = y \Rightarrow h(Y) = h(y) = k \text{ for } y \in \mathcal{Y}_k) \\
    &= \sum_{y \in \mathcal{Y}} \mathbb{P}(Y = y | h(Y) = k, A = b) ~\mathbb{E}[|\mathcal{C}(X)| \mid Y = y, h(Y) = k, A = b] \\
    &\hspace{50pt} (\text{because } \mathbb{P}(Y = y | h(Y) = k, A = b) = 0 \text{ for any } y \text{ not in } \mathcal{Y}_k)\\
    &= \mathbb{E}[|\mathcal{C}(X)| \mid h(Y) = k, A = b] ~~~~~~\text{(by the law of total expectation)}
\end{align*}
\normalsize
Then, plug in \Cref{boundI2}, we have
\begin{align*}
    (I)_2 &= \sum_{k=1}^{K} \big(\mathbb{P}(h(Y) = k | A = a) - \mathbb{P}(h(Y) = k | A = b) \big) ~\mathbb{E}[|\mathcal{C}(X)| \mid h(Y) = k, A = b]\\
    &= \sum_{k=1}^{K} \big(\mathbb{P}(h(Y) = k | A = a) - \mathbb{P}(h(Y) = k | A = b) \big) \big(\mathbb{E}[|\mathcal{C}(X)| \mid h(Y) = k, A = b] - c \big) ~~~~~\text{for any constant }c
\end{align*}
The last equality holds because 
\begin{align*}
    \sum_{k=1}^{K} c \big(\mathbb{P}(h(Y) = k | A = a) - \mathbb{P}(h(Y) = k | A = b) \big) &= c \Big(\sum_{k=1}^{K} \mathbb{P}(h(Y) = k | A = a) - \sum_{k=1}^{K} \mathbb{P}(h(Y) = k | A = b) \Big) \\
    &= c \cdot (1 - 1) = 0
\end{align*}
Define
\begin{align*}
    \mathbf{p}_a &:= [\mathbb{P}(h(Y) = 1 | A = a), \ldots, \mathbb{P}(h(Y) = K | A = a)]^T \\
    \mathbf{p}_b &:= [\mathbb{P}(h(Y) = 1 | A = b), \ldots, \mathbb{P}(h(Y) = K | A = b)]^T
\end{align*}
Then, we can bound $|(I)_2|$ as follows:
\begin{align}
    |(I)_2| &= \Big|\sum_{k=1}^{K} \big(\mathbb{P}(h(Y) = k | A = a) - \mathbb{P}(h(Y) = k | A = b) \big) \big(\mathbb{E}[|\mathcal{C}(X)| \mid h(Y) = k, A = b] - c \big) \Big| \nonumber\\
    &\le \Vert \mathbf{p}_a - \mathbf{p}_b \Vert_1 ~\max_{k = 1, \ldots, K} \big|\mathbb{E}[|\mathcal{C}(X)| \mid h(Y) = k, A = b] - c \big| ~~~\text{ by Hölder's inequality} \nonumber\\
    &= \frac{1}{2}\Vert \mathbf{p}_a - \mathbf{p}_b \Vert_1 \big(\max_{k=1,\ldots,K}\mathbb{E}[|\mathcal{C}(X)| \mid h(Y) = k, A = b] - \min_{k=1,\ldots,K}\mathbb{E}[|\mathcal{C}(X)| \mid h(Y) = k, A = b] \big),
    \label{spreadI2}
\end{align}
where the last equality holds when we choose 
\begin{equation*}
    c = \frac{1}{2}\big(\max_{k=1,\ldots,K}\mathbb{E}[|\mathcal{C}(X)| \mid h(Y) = k, A = b] + \min_{k=1,\ldots,K}\mathbb{E}[|\mathcal{C}(X)| \mid h(Y) = k, A = b] \big).
\end{equation*}
From \Cref{spreadI2}, we observe that the magnitude of $|(I)_2|$ depends on the product of: (1) the difference of cluster-membership distribution between the protected groups $a$ and $b$, and (2) the spread of $\mathbb{E}[|\mathcal{C}(X)| \mid h(Y) = k, A = b]$ over the clusters $k = 1,\ldots,K$. The factor (1), which shows up as $\Vert \mathbf{p}_a - \mathbf{p}_b \Vert_1$ in \Cref{spreadI2}, is induced by the correlation between $Y$ and $A$ (e.g., for the BiosBias data, there is correlation between occupation (label) and gender (protected group)). Since factor (1) is data-driven, and there is intrinsic correlation between $Y$ and $A$ in real-world data, we cannot control it directly by applying the Label-Clustered CP. Moreover, according to the triangle inequality, we always have $\Vert \mathbf{p}_a - \mathbf{p}_b \Vert_1 \le 2$. On the other hand, the factor (2), $\max_{k=1,\ldots,K}\mathbb{E}[|\mathcal{C}(X)| \mid h(Y) = k, A = b] - \min_{k=1,\ldots,K}\mathbb{E}[|\mathcal{C}(X)| \mid h(Y) = k, A = b]$ in \Cref{spreadI2}, is method-driven, and we can control it by choosing a proper number of clusters $K$. We provide two intuitive examples to illustrate how choice of $K$ can affect factor (2): the first one is when $K = 1$ in which case the Label-Clustered CP reduces to Marginal CP. In this case, we have factor (2) = 0, so the bound \Cref{spreadI2} becomes 0; the second example is another extreme case when $K = |\mathcal{Y}|$, that is, the number of clusters is exactly the number of possible labels, with each label forming its own cluster. In this case, for underrepresented labels with limited calibration data, we have large prediction sets for these labels, which increases set size gap among labels, resulting in a large difference in factor (2). 

In general, we do not want a large $K$. As indicated above, a large $K$ tends to increase the spread of set size across clusters. On the other hand, although the choice of $K = 1$ makes the upper bound in \Cref{spreadI2} vanish to 0, it boils down to Marginal CP, which, as discussed in bounding $|(I)_1|$, leads to large intra-cluster expected set size gap across labels. Moreover,  
as discussed in previous literature, Marginal CP can have significant disparity in terms of coverage across protected groups or labels \citep{vovk2005algorithmic, vovk2012conditional, foygelbarber2020limits}. In practice, we need to choose a proper $K$ when implementing the Label-Clustered CP to balance intra-cluster label homogeneity (for controlling $|(I)_1|$) and cross-cluster stability (for controlling $|(I)_2|$). With a suitable value of $K$ and its associated cluster assignments, we can bring both $|(I)_1|$ and $|(I)_2|$ to reasonably small values, achieving a tight bound for $|(I)|$. 

\noindent \textbf{3. How Label-Clustered CP helps to control $|(II)|$:}

Finally, we show how Label-Clustered CP can reduce $|(II)|$ in \Cref{eqII} compared to group-conditional CP. The term (II) is a weighted sum over $y \in \mathcal{Y}$ of the intra-label set size gap between group $a$ and group $b$. The differences in expected set size across protected groups after conditioning on the true label depend on how conformal thresholds are calibrated. The Mondrian CP estimates separate thresholds for each protected group, which can inflate $|(II)|$ for two reasons: (i) different thresholds impose different strictness levels across groups even within the same label, and (ii) in the case of imbalanced calibration data across groups (e.g., group $a$ has much more calibration data than group $b$ which has limited calibration data), 
splitting calibration dataset according to groups significantly reduces calibration set size of underrepresented groups, increasing quantile estimation variance and amplifying differences in the resulting set sizes among groups. In contrast, Label-Clustered CP mitigates the aforementioned inflation by using shared cluster thresholds (with each cluster threshold being the same across protected groups) estimated from pooled calibration data. This pooling stabilizes the calibration step and removes policy differences across groups induced by calibration, so the intra-label set size gap between groups is less amplified by threshold noise, typically resulting in a smaller $|(II)|$ than group-conditional CP. Below, we provide a mathematical proof to justify why Label-Clustered CP avoids inflating $|(II)|$ compared to Mondrian CP.

In what follows, a prediction set from the Label-Clustered CP is still denoted as $\mathcal{C}(X)$. For a learned clustering function $h: \mathcal{Y} \to \{1,\ldots,K\}$, let $\hat{q}_k$ be the conformal quantile for each cluster $k$, $k = 1,\ldots,K$. On the other hand, denote a prediction set constructed from Mondrian CP as $\mathcal{C}^{\text{group}}(X)$; let $\hat{q}^{\text{group}}_a$ and $\hat{q}^{\text{group}}_b$ be the conformal quantiles used in Mondrian CP for protected groups $a$ and $b$, respectively. Recall that the rule for constructing $\mathcal{C}(x)$ is
\begin{equation*}
    \mathcal{C}(x) = \{y \in \mathcal{Y}: s(x,y) \le \hat{q}_{h(y)} \}
\end{equation*}
and the rule for constructing $\mathcal{C}^{\text{group}}(x)$ is
\begin{equation*}
    \mathcal{C}^{\text{group}}(x) = \{y \in \mathcal{Y}: s(x,y) \le \hat{q}^{\text{group}}_{g(x)} \},
\end{equation*}
where $g: \mathcal{X} \to \mathcal{A}$ is the group assignment function. 

Now, for a fixed $y \in \mathcal{Y}$, a protected group $g \in \mathcal{A}$, and a threshold (quantile) $q$, define
\begin{equation}
    r_{y,g} (q) := \mathbb{E}[|\mathcal{C}_q(X)| \mid Y = y, A = g], 
    \label{appA: threshold_function}
\end{equation}
where $\mathcal{C}_q(\cdot)$ denotes the conformal prediction set obtained when the relevant rule uses threshold $q$ while holding everything else fixed. Furthermore, assume a mild regularity condition that changing the quantile threshold slightly cannot change the expected set size arbitrarily much. Mathematically, this assumption imposes a Lipschitz continuity on the function $r_{y,g}(\cdot)$, that is, there exists $0 \le L_{y,g} < \infty$ such that for all $q, q'$,
\begin{equation}
    \left|r_{y,g}(q) - r_{y,g}(q') \right| \le L_{y,g} |q - q'|.
    \label{appA: Lip_continuity_assump}
\end{equation}

Under Mondrian CP, consider any reference threshold $q^*$,
\begin{align}
    |(II)^{\text{group}}| &\le \sum_{y \in \mathcal{Y}} \mathbb{P}(Y = y| A = a) \left|\mathbb{E}[|\mathcal{C}^{\text{group}}(X)| \mid Y = y,A =a] - \mathbb{E}[|\mathcal{C}^{\text{group}}(X)| \mid Y = y,A =b]\right| \nonumber \\
    &= \sum_{y \in \mathcal{Y}} \mathbb{P}(Y = y| A = a) \left|r_{y,a}(\hat{q}^{\text{group}}_a) - r_{y,b}(\hat{q}^{\text{group}}_{b})\right| ~~\text{ by definition in \Cref{appA: threshold_function}} \nonumber \\
    &= \sum_{y \in \mathcal{Y}} \mathbb{P}(Y = y| A = a) \left|(r_{y,a}(q^*) - r_{y,b}(q^*)) + (r_{y,a}(\hat{q}^{\text{group}}_a) - r_{y,a}(q^*)) + (r_{y,b}(q^*) - r_{y,b}(\hat{q}^{\text{group}}_b)) \right| \nonumber \\
    & \le \sum_{y \in \mathcal{Y}} \mathbb{P}(Y = y| A = a) \left(|r_{y,a}(q^*) - r_{y,b}(q^*)| + |r_{y,a}(\hat{q}^{\text{group}}_a) - r_{y,a}(q^*)| + |r_{y,b}(q^*) - r_{y,b}(\hat{q}^{\text{group}}_b)| \right) \nonumber \\
    & \le \sum_{y \in \mathcal{Y}} \mathbb{P}(Y = y| A = a) \left(|r_{y,a}(q^*) - r_{y,b}(q^*)| + L_{y,a} |\hat{q}^{\text{group}}_a - q^*| + L_{y,b} |\hat{q}^{\text{group}}_b - q^*| \right) ~~\text{by \Cref{appA: Lip_continuity_assump}} \nonumber \\
    &= \underbrace{\sum_{y \in \mathcal{Y}} \mathbb{P}(Y = y| A = a)|r_{y,a}(q^*) - r_{y,b}(q^*)|}_{(II)_1} + \underbrace{\sum_{y \in \mathcal{Y}} \mathbb{P}(Y = y| A = a) \left(L_{y,a} |\hat{q}^{\text{group}}_a - q^*| + L_{y,b} |\hat{q}^{\text{group}}_b - q^*| \right)}_{(II)_2^{\text{group}}}
    \label{app: group-cond-term-II}
\end{align}
From the above derivation, term $(II)_1$ is data-driven, which comes from the difference in score distributions across protected groups conditional on $Y = y$. On the other hand, the term $(II)_2^{\text{group}}$ is induced by calibration of group-conditional CP which uses different group-specific thresholds $\hat{q}^{\text{group}}_a$ and $\hat{q}^{\text{group}}_b$. Moreover, term $(II)_2^{\text{group}}$ can be further inflated in the case that the calibration set is imbalanced across groups. For example, if group $b$ is underrepresented in the calibration set, then the quantile estimator $\hat{q}^{\text{group}}_b$ will have high bias and variance. Such noisy quantile estimation can lead to erratic behavior of set predictor, including large sets \citep{ding2023classconditional}.

Under Label-Clustered CP, following the same logic of deriving \Cref{app: group-cond-term-II}, we have
\begin{align}
    |(II)^{\text{cluster}}| &\le \sum_{y \in \mathcal{Y}} \mathbb{P}(Y = y| A = a) |r_{y,a}(\hat{q}_{h(y)}) - r_{y,b}(\hat{q}_{h(y)})| \nonumber \\
    &\le \sum_{y \in \mathcal{Y}} \mathbb{P}(Y = y | A = a) \left(|r_{y,a}(q^*) - r_{y,b}(q^*)| + |r_{y,a}(\hat{q}_{h(y)}) - r_{y,a}(q^*)| + |r_{y,b}(\hat{q}_{h(y)}) - r_{y,b}(q^*)| \right) \nonumber \\
    &\le \sum_{y \in \mathcal{Y}} \mathbb{P}(Y = y | A = a) \left(|r_{y,a}(q^*) - r_{y,b}(q^*)| + L_{y,a} |\hat{q}_{h(y)} - q^*| + L_{y,b} |\hat{q}_{h(y)} - q^*|\right) \nonumber \\
    &= \sum_{y \in \mathcal{Y}} \mathbb{P}(Y = y | A = a) |r_{y,a}(q^*) - r_{y,b}(q^*)| ~~\text{ if we set each reference $q^* = \hat{q}_{h(y)}$}
    \label{app: clustered-term-II}
\end{align}
Therefore, for Label-Clustered CP, the term $(II)_2^{\text{group}}$ induced by calibration disappears because the Label-Clustered CP uses the same threshold $\hat{q}_{h(y)}$ regardless of group. In contrast, in \Cref{app: group-cond-term-II}, $(II)_2^{\text{group}}$ cannot be eliminated due to different $\hat{q}^{\text{group}}_a$ and $\hat{q}^{\text{group}}_b$. This comparison shows that compared with group-conditional CP, which can inflate $|(II)|$ through split-calibration and group-specific quantile estimation, Label-Clustered CP enforces shared thresholds across groups within each label-cluster, thereby eliminating the inflation induced by calibration and leaving only the intrinsic intra-label cross-group disparity evaluated at a common threshold. 
\clearpage

\section{Conformal Prediction Algorithms}\label[appsec]{app:conformal}

\subsection{Clustered Conformal Prediction}

Clustered conformal prediction~\citep{ding2023classconditional} splits the calibration set into a clustering portion $\mathcal{D}_1$ (size $\lfloor \gamma n_{\mathrm{cal}} \rfloor$) and a calibration portion $\mathcal{D}_2$. A clustering function $h: \mathcal{Y} \to [K] \cup \{\mathrm{null}\}$ is learned on $\mathcal{D}_1$ (typically by embedding labels via their empirical score quantiles and applying $k$-means). Independent quantiles $\hat{q}_k$ are computed on $\mathcal{D}_2$ restricted to each cluster $k$ (with $\mathrm{null}$ using the full marginal $\mathcal{D}_2$).

The prediction set is
\[
\mathcal{C}(x_{\mathrm{test}}) = \{ y \in \mathcal{Y} : s(x_{\mathrm{test}}, y) \le \hat{q}_{h(y)} \}.
\]
Clustering strategies can be designed to promote fairness by grouping labels according to protected attributes (to support underrepresented groups), empirically identified unfair subpopulations, human-defined rules, or data-driven quantile-based approaches.

\subsection{Backward Conformal Prediction}

Backward conformal prediction~\citep{gauthier2025backwardconformalprediction} constrains prediction-set size via a rule $\mathcal{T}: (\mathcal{X} \times \mathcal{Y})^n \times \mathcal{X} \to \{1,\dots,|\mathcal{Y}|\}$, mapping calibration data and a test input to a maximum allowable size $l = \mathcal{T}(\mathcal{D}_{\mathrm{cal}}, x_{\mathrm{test}})$.

It relies on \emph{e-values}---nonnegative random variables $E$ with $\mathbb{E}[E] \leq 1$. For a positive score function $s > 0$, the test e-value for label $y$ is
\[
E_{\mathrm{test}}(y) = \frac{s(x_{\mathrm{test}}, y)}{\frac{1}{n+1} \Bigl( \sum_{i=1}^n S_i + s(x_{\mathrm{test}}, y) \Bigr)},
\]
where $S_i = s(x_i, y_i)$. The data-dependent level $\tilde{\alpha}$ is chosen as the smallest value such that the number of labels with $E_{\mathrm{test}}(y) < 1/\tilde{\alpha}$ does not exceed $l$:
\[
\tilde{\alpha} = \inf\!\Bigl\{\alpha \in (0,1] : \bigl|\{y : E_{\mathrm{test}}(y) < 1/\alpha\}\bigr| \le l \Bigr\}.
\]
The prediction set is
\[
\mathcal{C}(x_{\mathrm{test}}) = \{ y \in \mathcal{Y} : E_{\mathrm{test}}(y) < 1/\tilde{\alpha} \}.
\]
This satisfies $| \mathcal{C}(x_{\mathrm{test}}) | \le l$ and marginal coverage $\mathbb{P}\{y \in \mathcal{C}(x)\} \geq 1 - \mathbb{E}[\tilde{\alpha}]$.

In practice, $\mathbb{E}[\tilde{\alpha}]$ is estimated via leave-one-out estimator \(\hat\alpha^{\text{LOO}} = \frac{1}{n}\sum_{j=1}^n \tilde\alpha_j\), where each \(\tilde\alpha_j\) is computed by treating the \(j\)-th calibration point as a test observation.

\subsection{Pseudocode Implementations}
\begin{algorithm}[h]
\caption{Marginal (Split) Conformal Prediction}
\label{alg:split-cp}
\begin{algorithmic}[1]
\REQUIRE Calibration dataset $\mathcal{D}_{\mathrm{cal}} = \{(x_i,y_i)\}_{i=1}^{n_{\mathrm{cal}}}$, score function $s$, miscoverage level $\alpha$
\ENSURE Prediction set $\mathcal{C}_{\hat q_\alpha}(x)$

\STATE Compute calibration scores $S_i \gets s(x_i, y_i)$ for all $i \in [n_{\mathrm{cal}}]$
\STATE Compute $\tau_\alpha \gets \lceil (n_{\mathrm{cal}} + 1)(1-\alpha) \rceil / n_{\mathrm{cal}}$
\STATE Compute threshold $\hat q_\alpha \gets \operatorname{Quantile}_{\tau_\alpha}(S_1,\dots,S_{n_{\mathrm{cal}}})$
\STATE Define prediction set
\[
\mathcal{C}_{\hat q_\alpha}(x)
=
\{y \in \mathcal{Y} : s(x,y) \le \hat q_\alpha\}
\]
\end{algorithmic}
\end{algorithm}

\begin{algorithm}[h]
\caption{Mondrian (Group-Conditional) Conformal Prediction}
\label{alg:mondrian-cp}
\begin{algorithmic}[1]
\REQUIRE Calibration dataset $\mathcal{D}_{\mathrm{cal}}$, grouping function $g : \mathcal{X} \to \mathcal{A}$, score function $s$, level $\alpha$
\ENSURE Group-conditional prediction set $\mathcal{C}(x)$

\FOR{each group $a \in \mathcal{A}$}
    \STATE $\mathcal{I}_a \gets \{ i : g(x_i) = a \}$
    \STATE Compute scores $S_i \gets s(x_i,y_i)$ for $i \in \mathcal{I}_a$
    \STATE Compute
    \[
    \hat q_\alpha^{(a)} \gets 
    \operatorname{Quantile}_{\lceil (|\mathcal{I}_a|+1)(1-\alpha)\rceil / |\mathcal{I}_a|}
    \left(\{S_i\}_{i\in\mathcal{I}_a}\right)
    \]
\ENDFOR
\STATE Define prediction set
\[
\mathcal{C}(x)
=
\{y \in \mathcal{Y} : s(x,y) \le \hat q_\alpha^{(g(x))}\}
\]
\end{algorithmic}
\end{algorithm}

\begin{algorithm}[h]
\caption{Label-Clustered Conformal Prediction}
\label{alg:clustered-cp}
\begin{algorithmic}[1]
\REQUIRE Calibration data $\mathcal D_{\mathrm{cal}}=\{(X_i,Y_i)\}_{i=1}^{n_{\mathrm{cal}}}$, score function $s$, miscoverage level $\alpha$, split ratio $\gamma$
\ENSURE Prediction set $\mathcal C_{\mathrm{label-cluster}}(x)$

\STATE Select index set $I_1 \subset [n_{\mathrm{cal}}]$ with $|I_1|=\lfloor \gamma n_{\mathrm{cal}}\rfloor$
\STATE Define clustering set $\mathcal D_1=\{(X_i,Y_i): i\in I_1\}$ and calibration set $\mathcal D_2=\mathcal D_{\mathrm{cal}}\setminus\mathcal D_1$

\STATE Learn label clustering function
\[
 h:\mathcal Y \to [K]\cup\{\mathrm{null}\}
\]
using $\mathcal D_1$

\FOR{each cluster $k \in [K]\cup\{\mathrm{null}\}$}
    \STATE Define index set
    \[
    \mathcal I_k \gets \{ i \in \mathcal D_2 : h(Y_i)=k \}
    \]
    \STATE Compute scores $S_i \gets s(X_i,Y_i)$ for $i \in \mathcal I_k$
    \STATE Compute cluster quantile
    \[
    \hat q_k
    \gets
    \operatorname{Quantile}_{\lceil (|\mathcal I_k|+1)(1-\alpha)\rceil / |\mathcal I_k|}
    \left(\{S_i\}_{i\in\mathcal I_k}\right)
    \]
\ENDFOR

\STATE Construct prediction set
\[
\mathcal C_{\mathrm{label-cluster}}(x)
=
\{y \in \mathcal Y : s(x,y) \le \hat q_{h(y)}\}
\]
\end{algorithmic}
\end{algorithm}

\begin{algorithm}[h]
\caption{Group-Clustered Conformal Prediction}
\label{alg:grp-clustered-cp}
\begin{algorithmic}[1]
\REQUIRE Calibration data $\mathcal D_{\mathrm{cal}}=\{(X_i,Y_i)\}_{i=1}^{n_{\mathrm{cal}}}$, score function $s$, miscoverage level $\alpha$, split ratio $\gamma$
\ENSURE Prediction set $\mathcal C_{\mathrm{group-cluster}}(x)$

\STATE Select index set $I_1 \subset [n_{\mathrm{cal}}]$ with $|I_1|=\lfloor \gamma n_{\mathrm{cal}}\rfloor$
\STATE Define clustering set $\mathcal D_1=\{(X_i,Y_i): i\in I_1\}$ and calibration set $\mathcal D_2=\mathcal D_{\mathrm{cal}}\setminus\mathcal D_1$

\STATE Learn group clustering function
\[
 \tilde{h}:\mathcal A \to [K]\cup\{\mathrm{null}\}
\]
using $\mathcal D_1$

\FOR{each cluster $k \in [K]\cup\{\mathrm{null}\}$}
    \STATE Define index set
    \[
    \mathcal I_k \gets \{ i \in \mathcal D_2 : \tilde{h}(g(X_i)) =k \},
    \]
    where $g: \mathcal X \to \mathcal A$ is the group assignment function.
    \STATE Compute scores $S_i \gets s(X_i,Y_i)$ for $i \in \mathcal I_k$
    \STATE Compute cluster quantile
    \[
    \hat q_k
    \gets
    \operatorname{Quantile}_{\lceil (|\mathcal I_k|+1)(1-\alpha)\rceil / |\mathcal I_k|}
    \left(\{S_i\}_{i\in\mathcal I_k}\right)
    \]
\ENDFOR

\STATE Construct prediction set
\[
\mathcal C_{\mathrm{group-cluster}}(x)
=
\{y \in \mathcal Y : s(x,y) \le \hat q_{\tilde{h}(g(x))}\}
\]
\end{algorithmic}
\end{algorithm}

\begin{algorithm}[h]
\caption{Backward Conformal Prediction}
\label{alg:backward-cp}
\begin{algorithmic}[1]
\REQUIRE Calibration data $\{(X_i,Y_i)\}_{i=1}^n$, score function $s$, size constraint rule $\mathcal T$
\ENSURE Prediction set $\mathcal{C}_n^{\,\tilde\alpha}(x_{\mathrm{test}})$

\STATE Compute calibration scores $S_i \gets s(X_i,Y_i)$ for $i=1,\dots,n$

\FOR{each label $y \in \mathcal Y$}
    \STATE Compute test e-value
    \[
    E_{\mathrm{test}}(y)
    \gets
    \frac{s(x_{\mathrm{test}},y)}
    {\frac{1}{n+1}\Bigl(\sum_{i=1}^n S_i + s(x_{\mathrm{test}},y)\Bigr)}
    \]
\ENDFOR

\STATE Define data-dependent level
\[
\tilde\alpha
\gets
\inf\Bigl\{
\alpha \in (0,1):
\#\{y \in \mathcal Y : E_{\mathrm{test}}(y) < 1/\alpha\}
\le
\mathcal T\bigl((X_i,Y_i)_{i=1}^n, x_{\mathrm{test}}\bigr)
\Bigr\}
\]

\STATE Construct prediction set
\[
\mathcal{C}_n^{\,\tilde\alpha}(x_{\mathrm{test}})
=
\{y \in \mathcal Y : E_{\mathrm{test}}(y) < 1/\tilde\alpha\}
\]
\end{algorithmic}
\end{algorithm}
\clearpage

\section{Technical Details of the LLM-in-the-loop Evaluator}\label[appsec]{app:LLM_evaluator}
In this appendix, we provide details of constructing the logistic GEE model in \Cref{GEE_LLM} and computing the substantive fairness metric \textbf{maxROR} introduced in \Cref{sec:LLM_evaluator}. 

\subsection{Constructing GEE to Predict Downstream Prediction Accuracy}
\label[appsec]{app: LLM_evaluator_sub1}

To assess the effects of different prediction sets on human prediction accuracy and their disparity among protected groups, \citet{cresswell2025conformal} conducted randomized controlled trials with human decision makers, using generalized estimating equations (GEEs) to model accuracy against treatment (CP method), protected group, and task difficulty (approximated by marginal CP set size), then computing Odds Ratios (ORs) and Ratio of Odds Ratios (ROR) for treatment effects and disparities. In our evaluator for assessing substantive fairness, we consider using LLMs as downstream task predictors. For each task, an LLM is provided with the input $x$, a prediction set, and its coverage guarantee, outputting a label from all possible classes; in the control case, no set is provided.

For each $x_j \in \mathcal{D}_{\text{test}}$ and prediction set $\mathcal{C}_t(x_j)$ from treatment $t$ (Marginal, Mondrian, Label-Clustered, Group-Clustered, Backward; $t=1,\ldots,T$ including Control), the LLM makes $M$ independent predictions (to accommodate randomness of LLM responses for the same input from setting a non-zero temperature) based on $x_j$, $\mathcal{C}_t(x_j)$ and its coverage guarantee. The prompts used for describing the task and asking for LLM's prediction are provided in \Cref{app:add-exp}. Let $\hat{y}_{jt}^m$ be the $m$-th prediction and $R_{jt} = \frac{1}{M} \sum_{m=1}^M \mathbf{1}\{\hat{y}_{jt}^m = y_i\}$ be the empirical prediction accuracy. For a treatment $t$, the disparity of improvement in prediction accuracy (relative to Control) across protected groups can be estimated as
\begin{align}
  \hat{\Delta}_t & = \max_{a,b \in \mathcal{A}} \Bigg|\Big(\frac{1}{n_a}\sum_{j=1}^{n_a} R_{jt} - \frac{1}{n_a}\sum_{j=1}^{n_a} R_{j,\text{Control}} \Big) - \Big(\frac{1}{n_b}\sum_{j=1}^{n_b} R_{jt} - \frac{1}{n_b}\sum_{j=1}^{n_b} R_{j,\text{Control}} \Big) \Bigg|,
  \label{traditional_downstream_metric}
\end{align}
where $n_a, n_b$ are sizes of groups $a$ and $b$, respectively. However, this estimation can be misleading due to (i) neglecting confounding factors, such as systematic variations in task difficulty and the LLM's reliance on provided sets, and (ii) intra-task correlation of predictions across treatments (predictions made for the same task are based on the same $x$ and similar provided sets across treatments). We thus use logistic GEE regression, adjusting for covariates and clustering by task to account for the intra-task correlation.

In our LLM-in-the-loop setting, we observe that LLMs (especially the more capable ones) frequently constrain their answer to lie inside the provided prediction set. We therefore define an ``adoption'' indicator, $\text{adoption} = \mathbf{1}\text{\{the LLM's predicted label is contained in the provided set\}}$. Adoption captures the extent to which the prediction set is actually used as a decision aid, and it is strongly predictive of downstream correctness (see details in \Cref{app: compare-human-LLM}). Consequently, we treat ``adoption'' as an outcome-relevant covariate so that estimated treatment effects compare methods at comparable levels of reliance on the prediction set, rather than conflating treatment effects with shifts in how often the LLM follows the set. As a result, the following covariates are included in the logistic GEE model:
(i) $\text{treatment}_t$, the method used to construct the prediction set;
(ii) $\text{group}_j$, the protected group to which $x_j$ belongs;
(iii) $\text{diff}_{j}$, the difficulty of task $x_j$ approximated by the cardinality of the Marginal CP set;
(iv) $\text{adoption}_{j,t} := \frac{1}{M} \sum_{m=1}^{M} \mathbf{1}\{\hat{y}_{jt}^{m} \in \mathcal{C}_{t}(x_j)\}$, the proportion of predictions that fall within the provided set $\mathcal{C}_{t}(x_j)$ under treatment $t$ (with $\mathcal{C}_{\text{Control}}(x_j) = \varnothing$ yielding $\text{adoption}_{j,\text{Control}} = 0$ for each instance $x_j$).

Then, we fit a logistic GEE across all tasks to model the probability of correct prediction as a function of treatment, protected group, task difficulty, and adoption, using task-level clustering to account for intra-task correlation across treatments. As given in \Cref{sec:LLM_evaluator},  
the GEE is expressed as 
\begin{equation*}
    \text{logit} \left(\mathbb{E}[R_{jt}] \right) \sim \text{treatment}_{t} \times \text{group}_{j} + \text{diff}_{j} + \text{adoption}_{j,t},
\end{equation*}
for $j = 1,\ldots, N_{\text{test}}$ and $t = 1, \ldots, T$. The $\text{logit}(x) = \log \frac{x}{1-x}$, and the $\text{treatment}_{t} \times \text{group}_{j}$ means the interaction between $\text{treatment}_{t}$ and $\text{group}_{j}$.

When fitting the above GEE in \texttt{Python}, we set \texttt{cov\_struct = Exchangeable()}, which  assumes that all pairs of rows associated with the same task have the same correlation in their residuals after the mean is modeled. This is plausible because we have $T$ outcomes ($R_{j1}, \ldots, R_{jT}$) without natural ordering for $x_j$, and all these $T$ prediction outcomes share the same latent task difficulty and information, making a common intra-task correlation a reasonable condition. Even if there is heteroskedasticity within clusters, adding \texttt{.fit(cov\_type=`robust')} guarantees the consistency of standard error estimation when $N_{\text{test}}$ is large enough, protecting us against misspecifying the intra-task correlation. 

\subsection{Measuring Substantive Fairness from the LLM-in-the-loop Evaluator}
\label[appsec]{app: LLM_evaluator_sub2}

In what follows, we illustrate procedures on obtaining the $\textbf{maxROR}$ (\Cref{maxROR_def}) for different treatments as a measurement of substantive fairness from our LLM-in-the-loop evaluator. The $\textbf{maxROR}$ is obtained from the fitted GEE \Cref{GEE_LLM}, which provides model-based probabilities of correct prediction for each treatment and group.

First, define the notation of the estimated coefficients from fitting \eqref{GEE_LLM}: let 
\begin{itemize}
    \item $\hat{\beta}_0$ be the estimated intercept
    \item $\hat{\beta}_t^{\text{treatment}}$ be the estimated coefficient for treatment $t$
    \item $\hat{\beta}_a^{\text{group}}$ be the estimated coefficient for group $a$
    \item $\hat{\beta}_{t:a}$ be the estimated coefficient for the interaction of treatment $t$ and group $a$,
    \item $\hat{\beta}_{\text{diff}}$ be the estimated coefficient for difficulty
    \item $\hat{\beta}_{\text{adopt}}$ be the estimated coefficient for adoption
\end{itemize}
In model \eqref{GEE_LLM}, both treatment and group are categorical covariates, with control being the baseline category of treatment. Consider a non-control treatment $t$ and a group $a$, we define the model-based marginal probability of a correct response for treatment $t$ in group $a$ as
\begin{align}
    p_{t,a} &:= \frac{1}{N_{t,a}} \sum_{j \in \mathcal{I}_{t,a}} \text{logit}^{-1} \Big(\hat{\beta}_0 + \hat{\beta}_{t}^{\text{treatment}} + \hat{\beta}_{a}^{\text{group}} + 
    \hat{\beta}_{t:a} + \hat{\beta}_{\text{diff}} \text{diff}_{j} + \hat{\beta}_{\text{adopt}} \text{adoption}_{j,t} \Big),
    \label{app:pta_trt}
\end{align}
where $\mathcal{I}_{t,a}$ is the set of predictions in group $a$ that are applied treatment $t$, and $N_{t,a}$ is  the cardinality of the set $\mathcal{I}_{t,a}$. Thus, the $p_{t,a}$ defined in \eqref{app:pta_trt} is the model-based average probability that the LLM's prediction is correct, obtained by evaluating the fitted GEE at treatment $t$ and group $a$ while plugging in each task's difficulty and adoption rate under treatment $t$, and then averaging these predicted probabilities over all tasks in group $a$.

Similarly, for the baseline treatment ``control" and group $a$, we define 
\begin{align}
    p_{\text{control}, a} &:= \frac{1}{N_{\text{Control},a}} \sum_{j \in \mathcal{I}_{\text{Control},a}} \text{logit}^{-1} \Big(\hat{\beta}_0 + \hat{\beta}_{\text{Control}}^{\text{treatment}} + \hat{\beta}_{a}^{\text{group}} + 
    \hat{\beta}_{\text{Control}:a} + \hat{\beta}_{\text{diff}} \text{diff}_{j} + \hat{\beta}_{\text{adopt}} \text{adoption}_{j,\text{Control}} \Big) \nonumber\\
    &= \frac{1}{N_{\text{Control},a}} \sum_{j \in \mathcal{I}_{\text{Control},a}} \text{logit}^{-1} \left(\hat{\beta}_0 + \hat{\beta}_{a}^{\text{group}} + 
    \hat{\beta}_{\text{diff}} \text{diff}_{j} \right),
    \label{app:pta_control}
\end{align}
which is the model-based probability of a correct response under control in group $a$.

Then, for non-control treatment $t$ and every protected group $a$, the OR of $t$ versus control is given by 
\begin{equation*}
    \textbf{OR}_{t, a} := \frac{p_{t,a}/(1-p_{t,a})}{p_{\text{control},a}/(1-p_{\text{control},a})}
    \label{app:ORs}
\end{equation*}
For each protected group, the ORs assess how much more likely an LLM under treatment $t$ is to give the correct response than if it were in the control. If $\textbf{OR}_{t, a} > 1$, then for group $a$ the odds that the LLM produces a correct response under treatment $t$ are higher than them under the control, and if $\textbf{OR}_{t, a} < 1$ they are lower.  

The disparity of treatment effect on prediction accuracy is quantified by the ROR. For treatment $t$, the ROR between group $a$ and group $b$ is computed as
\begin{equation*}
    \textbf{ROR}_{t, a, b} := \frac{\textbf{OR}_{t,a}}{\textbf{OR}_{t,b}} - 1
    \label{app:RORs}
\end{equation*}
If $\textbf{ROR}_{t, a, b} \approx 0$, treatment $t$ provides the same treatment effect over the Control for group $a$ as it does for group $b$, indicating fairness of treatment effect from $t$ on prediction accuracy between the two groups. As mentioned in \Cref{sec:LLM_evaluator}, to compare the parity of conformal methods' impact on prediction accuracy, we compute and compare the \textbf{maxROR} 
\begin{equation*}
    \textbf{maxROR}_{t} := \max_{a,b \in \mathcal{A}} \textbf{ROR}_{t, a, b}
    \label{app: maxROR_def}
\end{equation*}
for each non-control treatment $t$.
For any two treatments $t_1$ and $t_2$, if $\textbf{maxROR}_{t_{1}} > \textbf{maxROR}_{t_{2}}$, then treatment $t_1$ induces a greater disparity in prediction accuracy across protected groups than treatment $t_2$. In this case, compared to $t_1$, treatment $t_2$ is preferred if the goal is to achieve fairness in downstream prediction accuracy under the assistance of prediction sets.

\clearpage

\section{Additional Experiment Details}\label[appsec]{app:add-exp}

\subsection{Dataset Details}
We consider four prediction tasks with open-access fairness datasets where algorithmic assistance may benefit human decision making. Across all tasks, we construct prediction sets using conformal prediction methods applied to the outputs of task-specific base models.

\paragraph{Image Classification.}
Image classification is widely used in high-stakes applications, including medical screening and surveillance, where biases may lead to serious societal consequences. Prior fairness research in visual domains has investigated facial recognition systems \citep{buolamwini2018gender}, medical image analysis \citep{drukker2023toward}, and methods for improving fairness in image classification models \citep{yang2024large}. To model a similar scenario, we use the \textbf{FACET} dataset \citep{gustafson2023facet}, which consists of images of people labeled by occupation and grouped by age.
We retain the 20 most common occupation classes: [\textit{Backpacker, Boatman, Computer User, Craftsman, Farmer, Guard, Guitarist, Gymnast, Hairdresser, Horse Rider, Laborer, Officer, Motorcyclist, Painter, Repairman, Salesperson, Singer, Skateboarder, Speaker, Tennis Player}], and split the data into calibration (\(\mathcal{D}_{\text{cal}}\)), calibration-validation (\(\mathcal{D}_{\text{calval}}\)), and test (\(\mathcal{D}_{\text{test}}\)) sets, stratified by class.
Age annotations are provided in four predefined groups: \emph{Younger}, \emph{Middle}, \emph{Older}, and \emph{Unknown}; see \Cref{table-facet-group-counts} for group distributions in \(\mathcal{D}_{\text{calval}}\), \(\mathcal{D}_{\text{cal}}\) and \(\mathcal{D}_{\text{test}}\). We employ CLIP ViT-L/14 \citep{radford2021learning} as a zero-shot image classifier to generate class scores, to which conformal prediction is applied. FACET is distributed under Meta's FACET usage agreement (custom license) and is intended for evaluation only; using FACET annotations for training is prohibited.

\paragraph{Text Classification.}
Text classification is commonly used to organize large volumes of text input, for example in hiring or recruitment, where demographic biases may arise. As a surrogate task, we use the \textbf{BiosBias} dataset \citep{dearteaga2019biasinbios}, which contains personal biographies labeled by occupation and grouped by binary gender.
We select the 10 most frequent occupations [\textit{Professor, Physician, Photographer, Journalist, Psychologist, Teacher, Dentist, Surgeon, Painter, and Model}] and partition the dataset into \(\mathcal{D}_{\text{train}}\), \(\mathcal{D}_{\text{val}}\), \(\mathcal{D}_{\text{cal}}\), \(\mathcal{D}_{\text{calval}}\), and \(\mathcal{D}_{\text{test}}\), ensuring class balance across splits. See \Cref{table-biosbias-group-counts} for binary group distributions in \(\mathcal{D}_{\text{calval}}\), \(\mathcal{D}_{\text{cal}}\) and \(\mathcal{D}_{\text{test}}\). 
A pre-trained BERT model \citep{devlin2019bert} is used to generate text representations, on which we train a linear classifier. Conformal prediction is applied using the classifier's output scores. The code and data-generation pipeline for BiosBias is released under the MIT License.

\paragraph{Audio Emotion Recognition.}
Emotion recognition arises naturally in human communication, though emotional expression can vary across speakers from different demographic groups. We use the \textbf{RAVDESS} dataset \citep{livingstone2018ravdess}, which contains audio recordings of professional actors expressing eight emotions [\textit{Happy, Angry, Calm, Fearful, Neutral, Disgust, Sad and Surprised}] using identical short phrases, with speakers grouped by binary gender.
The dataset is split into \(\mathcal{D}_{\text{cal}}\), \(\mathcal{D}_{\text{calval}}\), and \(\mathcal{D}_{\text{test}}\), stratified by emotion class and gender (see \Cref{table-ravdess-group-counts}).
We adopt a fine-tuned wav2vec~2.0 model \citep{baevski2020wav2vec,grosman2021xlsr53-large-english,wiam2023xlsr53-large-english} for emotion classification. RAVDESS is released under the Creative Commons Attribution-NonCommercial-ShareAlike 4.0 International license (CC BY-NC-SA 4.0).

\paragraph{Tabular Data Prediction.}
Tabular prediction tasks arise in many real-world decision-making settings, including banking, insurance underwriting, credit risk assessment, and public policy, where structured demographic and socioeconomic features are used to inform consequential decisions. We consider a tabular prediction setting using the \textbf{ACSIncome} dataset from the Folktables benchmark \citep{ding2021retiring}, which is derived from the 2023 U.S.\ Census data. Race is treated as the sensitive attribute for group-based evaluation. We re-grouped the race attribute to five categories [\textit{White alone, Black or African American alone, Asian alone, Two or More Races, All Other Races (Aggregated)}].
The task is to predict the income level of an individual among 10 predefined income brackets [\textit{104 - 9000, 9000 - 20000, 20000 - 30000, 30000 - 38800, 38800 - 48450, 48450 - 60000, 60000 - 75000, 75000 - 96900, 96900 - 140000, 140000 - 1672000}], using features such as education, employment, and household characteristics.
We partition the dataset into calibration (\(\mathcal{D}_{\text{cal}}\)), calibration-validation (\(\mathcal{D}_{\text{calval}}\)), and test (\(\mathcal{D}_{\text{test}}\)) splits, with stratification by income bracket. Refer to \Cref{table-acsincome-group-counts} for group distributions in \(\mathcal{D}_{\text{calval}}\), \(\mathcal{D}_{\text{cal}}\) and \(\mathcal{D}_{\text{test}}\).  
For classification, we employ an XGBoost model \citep{chen2016xgboost}, and apply conformal prediction to the model's output scores. ACSIncome is derived from the U.S. Census Bureau’s American Community Survey (ACS) Public Use Microdata Sample (PUMS); use of the underlying ACS PUMS data is governed by the Census Bureau’s terms of service.

\begin{table}[t]
\centering
\begin{minipage}{0.30\textwidth}
  \centering
  \caption{FACET Group Counts.}
  \label{table-facet-group-counts}
  \begin{center}
    \begin{small}
      \begin{sc}
        \begin{tabular}{lccr}
          \toprule
          Group     &  \(\mathcal{D}_{\text{calval}}\) &  \(\mathcal{D}_{\text{cal}}\) &   \(\mathcal{D}_{\text{test}}\)      \\
          \midrule
          Younger & 254 & 711 & 276 \\
          Middle & 772 & 2144 & 729\\
          Older & 103 & 299 & 91 \\
          Unknown & 271 & 846 & 304 \\
          \midrule
          Total & 1400 & 4000 & 1400\\
          \bottomrule
        \end{tabular}
      \end{sc}
    \end{small}
  \end{center}
\end{minipage}
\hfill
\begin{minipage}{0.30\textwidth}
  \centering
  \caption{BiosBias Group Counts.}
  \label{table-biosbias-group-counts}
  \begin{center}
    \begin{small}
      \begin{sc}
        \begin{tabular}{lccr}
          \toprule
          Group     &  \(\mathcal{D}_{\text{calval}}\) &  \(\mathcal{D}_{\text{cal}}\) &   \(\mathcal{D}_{\text{test}}\)      \\
          \midrule
          Female & 2424 & 4887 & 969 \\
          Male & 2576 & 5113 & 1031\\
          \midrule
          Total & 5000 & 10000 & 2000\\
          \bottomrule
        \end{tabular}
      \end{sc}
    \end{small}
  \end{center}
\end{minipage}
\hfill
\begin{minipage}{0.30\textwidth}
     \caption{RAVDESS Group Counts.}
  \label{table-ravdess-group-counts}
  \begin{center}
    \begin{small}
      \begin{sc}
        \begin{tabular}{lccr}
          \toprule
          Group     &  \(\mathcal{D}_{\text{calval}}\) &  \(\mathcal{D}_{\text{cal}}\) &   \(\mathcal{D}_{\text{test}}\)      \\
          \midrule
          Female & 120 & 420 & 180 \\
          Male & 120 & 420 & 180\\
          \midrule
          Total & 240 & 840 & 360\\
          \bottomrule
        \end{tabular}
      \end{sc}
    \end{small}
  \end{center}
\end{minipage}
\end{table}

\begin{table}[t]
  \caption{ACSIncome Group Counts.}
  \label{table-acsincome-group-counts}
  \begin{center}
    \begin{small}
      \begin{sc}
        \begin{tabular}{lccr}
          \toprule
          Group     &  \(\mathcal{D}_{\text{calval}}\) &  \(\mathcal{D}_{\text{cal}}\) &   \(\mathcal{D}_{\text{test}}\)      \\
          \midrule
          White & 6655 & 13487 & 6637 \\
          Black or African American & 842 & 1560 & 819\\
          Asian & 662 & 1362 & 663 \\
          Two or More Races & 1134 & 2164 & 1117 \\
          All Other Races & 707 & 1427 & 764\\
          \midrule
          Total & 10000 & 20000 & 10000\\
          \bottomrule
        \end{tabular}
      \end{sc}
    \end{small}
  \end{center}
  \vskip -0.1in
\end{table}

\subsection{Score Functions Used for Conformal Prediction}
The Marginal, Mondrian, Label-Clustered and Group-Clustered CP considered in this paper are implemented using one of two nonconformity score functions: \textbf{RAPS} \cite{angelopoulos2021uncertainty} or \textbf{SAPS} \cite{huang2024saps}. A nonconformity score $s:\mathcal{X}\times\mathcal{Y}\to\mathbb{R}$ assigns a scalar value $s(x,y)$ to each candidate label $y$ for an input $x$, where smaller values indicate that $y$ is more compatible with $x$ under the base model.
Different CP procedures in our experiments differ only in how these scores are calibrated (e.g., global vs.\ group-wise vs.\ cluster-wise calibration), but they all use the same underlying score definitions below.

\paragraph{Notation.}
Let $\mathcal{Y}=\{1,\dots,L\}$ be the label set. For an input $x$, let $f(x)\in\mathbb{R}^L$ denote the model logits and define temperature-scaled softmax probabilities
\[
p_y(x) \;=\; \frac{\exp(f_y(x)/T)}{\sum_{y'\in\mathcal{Y}}\exp(f_{y'}(x)/T)},\qquad T>0.
\]
Let $\pi_x(1),\dots,\pi_x(L)$ be labels sorted so that $p_{\pi_x(1)}(x)\ge p_{\pi_x(2)}(x)\ge\cdots\ge p_{\pi_x(L)}(x)$, and define the rank
\[
o_x(y) \;:=\; \min\{k:\ \pi_x(k)=y\}.
\]
We also use an independent randomization variable $u\sim\mathrm{Unif}[0,1]$ for tie-breaking.

\paragraph{RAPS.}
RAPS (Regularized Adaptive Prediction Sets \citep{angelopoulos2021uncertainty}) combines a randomized cumulative-mass term with an explicit penalty on lower-ranked labels. Define the cumulative probability mass strictly above $y$ by
\[
\rho_x(y) \;:=\; \sum_{k=1}^{o_x(y)-1} p_{\pi_x(k)}(x).
\]
Given hyperparameters $\lambda\ge 0$ and $k_{\mathrm{reg}}\in\{1,\dots,L\}$, the RAPS nonconformity score is
\[
s_{\mathrm{RAPS}}(x,y)
\;:=\;
\rho_x(y)\;+\;u\,p_y(x)\;+\;\lambda\bigl(o_x(y)-k_{\mathrm{reg}}\bigr)_+,
\qquad (a)_+ := \max\{a,0\}.
\]

\paragraph{SAPS.}
SAPS (Sorted Adaptive Prediction Sets \citep{huang2024saps}) is a rank-based score that retains the ordering information while
reducing dependence on small tail probabilities. Let $p_{\max}(x):=\max_{y'}p_{y'}(x)=p_{\pi_x(1)}(x)$.
With hyperparameter $\lambda\ge 0$, define
\[
s_{\mathrm{SAPS}}(x,y)
\;:=\;
\begin{cases}
u\,p_y(x), & \text{if } o_x(y)=1,\\[3pt]
p_{\max}(x)\;+\;\lambda\bigl(o_x(y)-2+u\bigr), & \text{if } o_x(y)\ge 2.
\end{cases}
\]

\subsection{Hyperparameters}

\paragraph{Tuning hyperparameters in RAPS and SAPS. } To construct CP sets, we split the data into three disjoint parts: a tuning set $\mathcal{D}_{\text{calval}}$, a calibration set $\mathcal{D}_{\text{cal}}$, and a test set $\mathcal{D}_{\text{test}}$. Hyperparameters in the score functions are selected using $\mathcal{D}_{\text{calval}}$. After tuning, conformal thresholds are computed on $\mathcal{D}_{\text{cal}}$ with the chosen hyperparameters. Prediction sets used in the LLM-in-the-loop evaluator for the downstream task are then obtained on $\mathcal{D}_{\text{test}}$.

We tune hyperparameters by using Bayesian optimization via the Optuna library \citep{akiba2019optuna}, coupled with the TPESampler for efficient search over 50 iterations to minimize average set size. For each set of candidate hyperparameters, we (i) compute the conformal threshold(s) $\hat{q}$ from $\mathcal{D}_{\text{cal}}$, (ii) form prediction sets according to the rules of CP methods on $\mathcal{D}_{\text{calval}}$, and (iii) score the hyperparameters candidate by the average set size on $\mathcal{D}_{\text{calval}}$. We select the optimal hyperparameters as the minimizer of average set size. For Mondrian, Label-Clustered and Group-Clustered CP, the same set of hyperparameters is used across all groups/clusters during tuning, while the final conformal thresholds are calibrated separately within each group/cluster. \Cref{tab:hyperparams} presents the final hyperparameters after the tuning procedure.    

\begin{table}[H]
\centering
\caption{Hyperparameter Settings for Each Dataset After Tuning}
\label{tab:hyperparams}
\footnotesize
\setlength{\tabcolsep}{3pt}
\renewcommand{\arraystretch}{0.9}
\begin{sc}
\begin{tabular}{llccc ccc ccc ccc}
\toprule
\multirow{2}{*}{Dataset} & \multirow{2}{*}{Score Function}
& \multicolumn{3}{c}{Marginal} & \multicolumn{3}{c}{Mondrian} & \multicolumn{3}{c}{Label-Clustered} & \multicolumn{3}{c}{Group-Clustered} \\
\cmidrule(lr){3-5}\cmidrule(lr){6-8}\cmidrule(lr){9-11}\cmidrule(lr){12-14}
& & $T$ & $\lambda$ & $k_{\mathrm{reg}}$ & $T$ & $\lambda$ & $k_{\mathrm{reg}}$ & $T$ & $\lambda$ & $k_{\mathrm{reg}}$ & $T$ & $\lambda$ & $k_{\mathrm{reg}}$\\
\midrule
FACET   & RAPS & 0.53 & 0.07 & 4 & 0.30 & 0.16 & 4 & 0.51 & 1.38 & 4 & 0.30 & 0.16 & 4  \\
BiosBias& SAPS & 0.56 & 0.20 &-- & 0.40 & 0.20 & --& 0.74& 0.28& --& 0.47& 0.19&--\\
RAVDESS & RAPS & 0.16 & 1.61 & 3 & 0.15 & 0.69 & 3 & 0.17 & 0.35 & 3 & 0.16 & 0.51 & 3 \\
ACSIncome & RAPS & 0.09 & 0.05 & 4 & 0.10 & 0.05 & 4 & 0.09 & 0.05 & 4 & 0.09 & 0.05 & 4 \\
\bottomrule
\end{tabular}
\end{sc}
\end{table}

\paragraph{Hyperparameters in Clustered CP. } As described in \Cref{app:conformal}, we use a proportion parameter $\gamma$ to determine the size of the clustering subset, $\lfloor \gamma n_{\mathrm{cal}} \rfloor$, which is used to learn the cluster assignments. In our experiments, we set $\gamma = 0.3$, which provides sufficient data to estimate stable cluster structure while leaving enough observations in the remaining calibration set to estimate conformal thresholds. Following \citet{ding2023classconditional}, we use $\{0.5, 0.6, 0.7, 0.8, 0.9 \} \cup \{1 -\alpha\}$-quantiles of a score distribution of class/group as the embedding vector for clustering. 

To ensure these quantile features are well-defined, we set $n_{\alpha} = (1/\alpha) - 1$ (e.g., $n_{\alpha} = 9$ when $\alpha = 0.1$), the minimum sample size for which the empirical $(1-\alpha)$-quantile is finite. Any class/group with at most $n_\alpha$ observations in the clustering subset is assigned to a \emph{null} cluster; the remaining classes/groups are embedded via the above quantiles and clustered using $k$-means.

\paragraph{Hyperparameters and implementation of Backward CP. } In our implementation of Backward CP, we do not use RAPS or SAPS score functions, instead, we use the cross-entropy loss as the score to compute the e-value. Let $f(x) \in \mathbb{R}^{|\mathcal{Y}|}$ denote model logits and $p(y|x) = \text{softmax}(f(x))_y$. We use the cross-entropy loss
\begin{equation*}
    s_{\text{NLL}}(x,y) = -\log \left(p(y|x) + \epsilon \right),
\end{equation*}
with $\epsilon > 0$ for numeric stability so the score is well-defined even when $p(y|x) = 0$. In our experiment, we set $\epsilon = 1e-4$. Given calibration data $\{(X_i, Y_i)\}_{i=1}^{n}$, define $S_i = s_{\text{NLL}}(X_i, Y_i)$. For a test input $x_{\text{test}}$ and candidate label $y$, compute the e-value
\begin{equation*}
    E_{\text{test}} (y) = \frac{(n+1)s_{\text{NLL}}(x_{\text{test}}, y)}{\sum_{i=1}^{n} S_i + s_{\text{NLL}}(x_{\text{test}}, y)}.
\end{equation*}
For any $\alpha \in (0, 1)$, the prediction set is 
\begin{equation*}
    \mathcal{C}_n^{\alpha} (x_{\text{test}}) = \{y \in \mathcal{Y}: E_{\text{test}} (y) < 1/\alpha \},
\end{equation*}
using a strict inequality to break ties.

Backward CP selects a data-dependent level $\tilde{\alpha}(x_\text{test})$ to satisfy a maximum set size constraint $\mathcal{T}$. In our implementation, the target size is 
\begin{equation*}
    \mathcal{T} = \lceil \text{average set size from Marginal CP} \rceil + \text{offset}.
\end{equation*}
We choose $\tilde{\alpha}$ as the smallest $\alpha \in [\epsilon', 1 - \epsilon']$ (set $\epsilon' = 1e-4$) such that $|\mathcal{C}_n^{\alpha}| \le \mathcal{T}$. The $\tilde{\alpha}$ is computed by binary search up to a tolerance parameter (which we set to $1e-7$). If the resulting set is empty, we increase $\mathcal{T}$ by one and repeat to ensure a nonempty prediction set. Moreover, the offset starts from 0, and we increase it iteratively until the empirical coverage meets the target $1 - \alpha$ up to a fixed maximum number of iterations, which we set to be 10. 

\subsection{Prompts Used for LLM-in-the-loop Evaluator}

For LLM-based downstream prediction tasks, we use different LLMs depending on the modality and dataset: Qwen2.5-VL-7B-Instruct (\url{https://huggingface.co/Qwen/Qwen2.5-VL-7B-Instruct}) for FACET (vision-language), GPT-4o-mini (\url{https://platform.openai.com/docs/models/gpt-4o-mini}) and Llama-3.2-3B (\url{https://huggingface.co/meta-llama/Llama-3.2-3B}) for BiosBias (text), GPT-4o-audio-preview (\url{https://platform.openai.com/docs/models/gpt-4o-audio-preview}) for RAVDESS (audio), and Qwen2.5-7B (\url{https://huggingface.co/Qwen/Qwen2.5-7B}) for ACSIncome (tabular and text-only prompting). These choice reflect modality compatibility and were held fixed across CP methods within each dataset to ensure fair comparisons.

The prompt templates used in our experiments are as follows:

\noindent \textbf{1. FACET}

\begin{promptbox}
You are given an image of a person and a list of possible classes: [{all_options}].
Select exactly one class name from this list. Respond with only the class name. Do not 
provide explanations or additional text.

image attached
The following shorter list contains classes that are more likely to be correct:
[{options}]
Based on the image, which class is the person most likely to belong to?
\end{promptbox}

Control:

\begin{promptbox}
You are given an image of a person and a list of possible classes: [{all_options}].
Select exactly one class name from this list. Respond with only the class name. Do not 
provide explanations or additional text.

image attached
Based on the image, which class is the person most likely to belong to?
\end{promptbox}

\noindent \textbf{2. BiosBias}

\begin{promptbox}
You are an HR expert. Given a short biography, select the correct occupation from the 
provided list: [{all_options}].
Respond with only the occupation name from this list. Do not explain your reasoning.

Biography:
{prompt}
The following shorter list contains occupations that are more likely to be correct:
[{options}]
Based on the biography above, which occupation is the best match?
\end{promptbox}

Control:

\begin{promptbox}
You are an HR expert. Given a short biography, select the correct occupation from the 
provided list: [{all_options}].
Respond with only the occupation name from this list. Do not explain your reasoning.

Biography:
{prompt}
Based on the biography above, which occupation is the best match?
\end{promptbox}

\noindent \textbf{3. RAVDESS}

\begin{promptbox}
You are an expert in emotion classification from audio. 
Instructions:
  - Focus on HOW they speak, not WHAT they say
  - Listen for vocal tone, pitch patterns, and energy
  - Classify which emotion does the audio convey

Choose one emotion from this list: [{all_options}]
Answer with ONE word from the list only. 

Prediction set from a classifier with {coverage_info} confidence to contain the 
true answer: [{options}]

Listen carefully to the speaker's voice. Analyze:
  1. Vocal tone
  2. Pitch patterns (high/low, rising/falling)
  3. Speaking energy
  4. Emotional expression in delivery
  
Based on these cues in the audio, which emotion best matches the speaker's vocal expression?
\end{promptbox}

Control:

\begin{promptbox}
You are an expert in emotion classification from audio. 
Instructions:
  - Focus on HOW they speak, not WHAT they say
  - Listen for vocal tone, pitch patterns, and energy
  - Classify which emotion does the audio convey

Choose one emotion from this list: [{all_options}]
Answer with ONE word from the list only. 

Listen carefully to the speaker's voice. Analyze:
  1. Vocal tone
  2. Pitch patterns (high/low, rising/falling)
  3. Speaking energy
  4. Emotional expression in delivery
  
Based on these cues in the audio, which emotion best matches the speaker's vocal expression?
Choose from: [{all_options}]
\end{promptbox}

\noindent \textbf{4. ACSIncome}

\begin{promptbox}
You are a labor economics expert. Given a structured demographic and employment profile 
from a census survey, select the correct income bracket from the provided list: 
[{all_options}]. 
Respond with only the income bracket label. Do not explain your reasoning.

Profile: The following is a structured demographic and employment profile derived from a 
census survey.
Age: {AGEP}
Class of worker: {COW}
Educational attainment: {SCHL}
Marital status: {MAR}
Occupation code: {OCCP}
Place of birth: {POBP}
Employment status of parent: {ESP}
Relationship to household reference person: {RELSHIPP}
Usual hours worked per week (past 12 months): {WKHP}
Sex: {SEX}
Race: {RAC1P}
The following income brackets are more likely to be correct: [{options}]
Based on the profile above, which income bracket is the best match?
\end{promptbox}

Control:

\begin{promptbox}
You are a labor economics expert. Given a structured demographic and employment profile 
from a census survey, select the correct income bracket from the provided list: 
[{all_options}]. 
Respond with only the income bracket label. Do not explain your reasoning.

Profile: The following is a structured demographic and employment profile derived from a 
census survey.
Age: {AGEP}
Class of worker: {COW}
Educational attainment: {SCHL}
Marital status: {MAR}
Occupation code: {OCCP}
Place of birth: {POBP}
Employment status of parent: {ESP}
Relationship to household reference person: {RELSHIPP}
Usual hours worked per week (past 12 months): {WKHP}
Sex: {SEX}
Race: {RAC1P}
Based on the profile above, which income bracket is the best match?
\end{promptbox}

\subsection{Computational Cost and Choice of $M$}
\label[appsec]{app: computational_cost}

Our LLM-in-the-loop evaluator queries the LLM $M$ independent times for each task instance and treatment, and then uses the resulting empirical accuracy $R_{jt}$ in the GEE-based estimation procedure described in \Cref{sec:LLM_evaluator} and \Cref{app:LLM_evaluator}. Increasing $M$ reduces the estimation variability induced by stochastic LLM responses, but also increases API cost or wall-clock runtime. We therefore selected $M$ by balancing statistical stability and computational cost across evaluation settings. In general, we used $M = 5$ for API-based models and up to $M = 8$ for open-source models when computationally feasible. For ACSIncome, which has a substantially larger number of test tasks (as reported in \Cref{table-acsincome-group-counts}), we used $M = 4$, which still resulted in 240000 total LLM predictions.

\Cref{tab:computational-cost} summarizes the number of independent predictions $M$ and computational cost for each LLM-in-the-loop experiment. For API-based models, we report monetary cost in USD. For open-source models run locally, we report wall-clock GPU time spent on a single Nvidia RTX 5060Ti 16GB GPU. These results show that the proposed evaluator is substantially more scalable than human-subject experiments while still allowing repeated independent predictions for each task instance. Although a larger $M$ would further reduce stochastic variability, our bootstrap results in \Cref{app:bootstrap-results} provide additional evidence that the qualitative conclusions are stable under resampling. In particular, the bootstrap standard errors for \textbf{maxROR} are small. Thus, while $M$ is necessarily chosen subject to computational constraints, the design of repeated prediction together with the bootstrap analysis provides a practical robustness check for the reported downstream-fairness trends among CP methods.

\begin{table}[H]
\centering
\small
\setlength{\tabcolsep}{9pt}
\caption{
Computational cost of the LLM-in-the-loop evaluation. 
The column ``Total predictions'' counts the total number of LLM predictions across task instances and treatments, including the $M$ repeated predictions per instance-treatment pair. 
For API-based models, cost is reported in USD; for open-source models run locally, we report elapsed wall-clock runtime on a Nvidia RTX 5060Ti 16GB GPU. 
}
\label{tab:computational-cost}
\begin{tabular}{llccc}
\toprule
Dataset & LLM-in-the-loop & $M$ & Total predictions & Cost/wall-clock time \\
\midrule
FACET & Qwen2.5-VL-7B-Instruct & 8 & 67200 & 30 h 56 min \\
BiosBias & GPT-4o-mini & 5 & 60000 & \$0.82 \\
BiosBias & Llama-3.2-3B & 8 & 96000 & 1 h 52 min \\
RAVDESS & GPT-4o-audio-preview & 5 & 10800 & \$16 \\
ACSIncome & Qwen2.5-7B & 4 & 240000 & 22 h 50 min \\
\bottomrule
\end{tabular}
\end{table}

\clearpage

\section{Details and Results from the LLM-in-the-loop Evaluator}\label[appsec]{app:details-evaluator}

\subsection{Comparing Human-in-the-loop and LLM-in-the-loop Evaluators}
\label[appsec]{app: compare-human-LLM}

We present a detailed comparison between our LLM-in-the-loop evaluator and the prior human-in-the-loop evaluator \citep{cresswell2025conformal} to show the validity of using the LLM-in-the-loop evaluator as a scalable proxy for assessing substantive fairness of CP methods. Moreover, we use our experimental results to illustrate how we calibrate the LLM-in-the-loop evaluator so that it captures the key behaviors of substantive fairness discovered in the prior study with human subjects.

\paragraph{Adoption-rate-difference in human vs. LLM.} 
The following \Cref{adoption_comp_FACET}, \Cref{adoption_comp_BiosBias}, and \Cref{adoption_comp_RAVDESS} present the adoption rate (proportion of predicted labels contained in the provided prediction set) of the prior human subjects and the LLMs used in our experiments. Here, we consider the intersection of the CP methods (Marginal and Mondrian) and datasets (FACET, BiosBias, RAVDESS) between the ones used by \citet{cresswell2025conformal} and in our experiments. As observed from the tables, overall, LLMs have higher adoption rates compared to human subjects, indicating that LLMs tend to closely follow the prompt and rely on the provided prediction set when making decisions. Inearly experiments, we found that this tendency of high reliance on provided sets is more prominent when using more capable LLMs. 

\paragraph{Calibrate the LLM-in-the-loop evaluator.} 
As LLMs are more dependent on the provided prediction set when outputting a response, accuracy differs systematically across adoption status, and invalid responses (answers that are outside of the label space) often occur when the LLM does not pick a label from the provided set (see \Cref{table-cond-prob-adoption}). Given that adoption is strongly predictive of downstream correctness from LLM, we treat adoption as an outcome-relevant covariate so that estimated treatment effects compare methods at comparable levels of reliance on the prediction set. 

Here, we include an example of $\textbf{maxROR}$ computed from the GEE without the adoption covariate to illustrate that omitting adoption can yield misleading conclusions on substantive fairness, because treatments and groups may be compared at different (and uneven) levels of reliance on the provided set. As what described in \Cref{sec:LLM_evaluator}, for each test $x_j$ in BiosBias and treatment $t$, we ask LLM to predict $y_j$ based on $x_j$, $\mathcal{C}_t(x_j)$, and the stated coverage guarantee $M$ times (in the control case, no prediction set is provided). Then, we compute the $R_{jt}$, which is the proportion of correct responses for $x_j$ under treatment $t$ out of the $M$ predictions, and fit to the GEE proposed by \citet{cresswell2025conformal}:
\begin{equation}
    \text{logit}(\mathbb{E}[R_{jt}]) \sim \text{treatment}_t \times \text{group}_j + \text{diff}_j
    \label{GEE2025}
\end{equation}
The \textbf{OR} and \textbf{maxROR} are computed based on the fitted \Cref{GEE2025} according to the same logic described in \Cref{app: LLM_evaluator_sub2}. The results are shown in \Cref{counter-example-GEE}. As we can see, in this case, $\textbf{maxROR}_{\text{Marginal}} > \textbf{maxROR}_{\text{Mondrian}}$, which does not align with the behavior discovered by \citet{cresswell2025conformal}. This is because the LLM's adoption rate for the Male group under the Marginal treatment is extremely high (98.20\%; see \Cref{adoption_comp_BiosBias}). As a result, predictions for the Male group benefit disproportionately from the Marginal treatment, inflating $\textbf{OR}_{\text{Marginal, Male}}$ and hence $\textbf{maxROR}_{\text{Marginal}}$. This counterexample motivates including ``adoption'' as a covariate in the GEE.

\paragraph{Alignment with human-in-the-loop experiments.}
To make our evaluator reflect the substantive-fairness pattern discovered by \citet{cresswell2025conformal} while respecting the different experimental design, we treat each task instance as the clustering unit in the GEE \Cref{GEE_LLM} with adoption as a covariate.
From the results shown in \Cref{app:subsec-evaluator-result} (also summarized in \Cref{tab:human-llm-compare}), we see that the LLM-in-the-loop evaluator outputs $\textbf{maxROR}_{\text{Marginal}} < \textbf{maxROR}_{\text{Mondrian}}$ in our experiments on FACET, BiosBias, and RAVDESS. Moreover, the $\textbf{maxROR}$ shows a pattern that it is small when the set size disparity between sensitive groups is small (e.g., in the case of Marginal, Label-Clustered, or Backward), and the $\textbf{maxROR}$ is large for Mondrian and Group-Clustered CP, which have large set size disparity due to equalizing coverage across groups. This trend of $\textbf{maxROR}$ reflects the downstream prediction behavior discovered by \citet{cresswell2025conformal}, validating that the LLM-in-the-loop evaluator captures key properties of substantive fairness revealed in the prior human-in-the-loop study, and supporting the use of the LLM-in-the-loop evaluator as a proxy for assessing substantive fairness across CP methods. 
\begin{table}[H]
  \caption{Adoption rate of human vs. LLM (Qwen2.5-VL-7B-Instruct) for FACET}
  \label{adoption_comp_FACET}
  \begin{center}
    \begin{small}
      \begin{sc}
        \begin{tabular}{lccc}
          \toprule
          Treatment     &  Group  & Human adoption\%   & LLM adoption\% \\
          \midrule
          \multirow{4}{8em}{Marginal} & Middle & 92.82 &  96.52\\
          & Older & 91.66 & 90.66 \\
          & Unknown & 96.76 & 99.01\\
          & Younger & 94.79 & 98.23\\
          \midrule
          \multirow{4}{8em}{Mondrian} & Middle & 93.22 &  96.42\\
          & Older & 92.91 &  92.17\\
          & Unknown & 96.76 & 98.68\\
          & Younger & 96.66 & 98.05\\
          \bottomrule
        \end{tabular}
      \end{sc}
    \end{small}
  \end{center}
  \vskip -0.1in
\end{table}

\begin{table}[H]
  \caption{Adoption rate of human vs. LLM (GPT-4o-mini) for BiosBias}
  \label{adoption_comp_BiosBias}
  \begin{center}
    \begin{small}
      \begin{sc}
        \begin{tabular}{lccc}
          \toprule
          Treatment     &  Group  & Human adoption\%   & LLM adoption\% \\
          \midrule
          \multirow{2}{8em}{Marginal} & Female & 92.31 & 95.22 \\
          & Male & 93.87 & 98.20 \\
          \midrule
          \multirow{2}{8em}{Mondrian} & Female & 92.30 & 92.07 \\
          & Male & 94.48 & 97.36 \\
          \bottomrule
        \end{tabular}
      \end{sc}
    \end{small}
  \end{center}
  \vskip -0.1in
\end{table}

\begin{table}[H]
  \caption{Adoption rate of human vs. LLM (GPT-4o-audio-preview) for RAVDESS}
  \label{adoption_comp_RAVDESS}
  \begin{center}
    \begin{small}
      \begin{sc}
        \begin{tabular}{lccc}
          \toprule
          Treatment     &  Group  & Human adoption\%   & LLM adoption\% \\
          \midrule
          \multirow{2}{8em}{Marginal} & Female & 90.67 & 91.89 \\
          & Male & 87.97 & 92.44 \\
          \midrule
          \multirow{2}{8em}{Mondrian} & Female & 91.95 & 90.00 \\
          & Male & 92.20 &  94.11\\
          \bottomrule
        \end{tabular}
      \end{sc}
    \end{small}
  \end{center}
  \vskip -0.1in
\end{table}

\begin{table}[H]
  \caption{Empirical probabilities of correct response and invalid response conditioning on adoption status for FACET, BiosBias, and RAVDESS datasets. As we can see, the $\mathbb{P}(\text{correct} | \text{adoption} = 1)$ is higher than $\mathbb{P}(\text{correct} | \text{adoption} = 0)$, especially for the challenging FACET and RAVDESS tasks.}
  \label{table-cond-prob-adoption}
  \begin{center}
    \begin{small}
      \begin{sc}
        \begin{tabular}{lccc}
          \toprule
              &  FACET  & BiosBias   & RAVDESS \\
          \midrule
          $\mathbb{P}(\text{correct} \mid \text{adoption} = 1)$ & 78.40\% & 80.64\% & 46.58\% \\
          $\mathbb{P}(\text{correct} \mid \text{adoption} = 0)$ & 37.99\% & 76.74\% & 0.00\% \\
          $\mathbb{P}(\,\,\,\text{invalid} \mid \text{adoption} = 0)$ & 25.14\% & 9.30\% & 41.75\% \\
          \bottomrule
        \end{tabular}
      \end{sc}
    \end{small}
  \end{center}
  \vskip -0.1in
\end{table}

\begin{table}[H]
  \caption{The \textbf{OR} and \textbf{maxROR} computed from fitting a GEE without the ``adoption'' covariate for BiosBias. As shown on the table, $\textbf{OR}_{\text{Marginal}, \text{Male}}$ has a relatively large value from the relatively high adoption rate, resulting in larger $\textbf{maxROR}_{\text{Marginal}}$ than $\textbf{maxROR}_{\text{Mondrian}}$.}
  \label{counter-example-GEE}
  \begin{center}
    \begin{small}
      \begin{sc}
        \begin{tabular}{lcccr}
          \toprule
           Group   &  $\textbf{OR}_{\text{marginal}}$  & $\textbf{OR}_{\text{Mondrian}}$   &  $\textbf{maxROR}_{\text{marginal}} (\%)$  & $\textbf{maxROR}_{\text{Mondrian}} (\%)$ \\
          \midrule
          Female & 1.026 & 1.219 & \multirow{2}{4em}{27.4} & \multirow{2}{4em}{9.9} \\
          Male & 1.308 & 1.110 \\
          \bottomrule
        \end{tabular}
      \end{sc}
    \end{small}
  \end{center}
  \vskip -0.1in
\end{table}

\subsection{Robustness to Group-Varying Adoption Effects}
\label[appsec]{app: group-adoption-interaction}

The previous subsection motivates the inclusion of adoption in the GEE model: because LLMs tend to rely strongly on the provided prediction set, and because correctness differs systematically across adoption status, adoption is an outcome-relevant covariate for comparing CP methods at comparable levels of reliance on the prediction set. Here, we further examine whether this adjustment is sensitive to group-varying adoption effects.

A potential concern is that if adoption varies systematically across groups in a way related to set quality, adjusting for adoption could introduce collider-related distortion. Our goal is not to interpret the adoption coefficient causally, but to compare CP methods while accounting for the degree to which the LLM relies on the provided set.

Define the following two GEE specifications. The first is the model \Cref{GEE_LLM} used in the main experiments:
\begin{equation}
\mathrm{Model~1:}\qquad
\mathrm{logit}\{\mathbb{E}[R_{jt}]\}
\sim
\mathrm{treat}_t \times \mathrm{group}_j + \mathrm{diff}_j + \mathrm{adoption}_{jt}.
\end{equation}
The second augments this model with an adoption-by-group interaction:
\begin{equation}
\mathrm{Model~2:}\qquad
\mathrm{logit}\{\mathbb{E}[R_{jt}]\}
\sim
\mathrm{treat}_t \times \mathrm{group}_j + \mathrm{diff}_j
+ \mathrm{adoption}_{jt} \times \mathrm{group}_j.
\end{equation}

To assess whether our substantive-fairness conclusions are sensitive to group-varying adoption effects, we conduct two complementary robustness checks. First, we fit the expanded Model~2 and test whether these interaction coefficients are equal to zero using Wald $z$-tests, with significance assessed at level $0.05$. Second, because our main conclusions are based on \textbf{maxROR} rather than on coefficient-level interpretation, we recompute \textbf{maxROR} under the augmented model (using the same marginalization procedure described in \Cref{app: LLM_evaluator_sub2}) and compare them with those obtained from fitting the Model~1. This second check directly evaluates whether allowing the association between adoption and correctness to vary by group changes the reported downstream-fairness comparisons among CP methods.

\Cref{tab:adoption-interaction-pvalues} reports the $p$-values from the Wald $z$-tests, and \Cref{tab:adoption-interaction-maxror} reports the comparison of \textbf{maxROR} between the two GEE specifications. On BiosBias and FACET, none of the adoption-by-group interaction terms is statistically significant at the $0.05$ level. As for the \textbf{maxROR} comparisons of BiosBias and FACET, the values are nearly identical, with only small numerical differences that do not change the ranking or qualitative comparison among CP methods. For ACSIncome, the expanded Model~2 fit produced reasonable Wald $p$-values for the adoption-by-group interaction terms, but some corresponding coefficient estimates and standard errors were large, likely due to sparse subgroup cells or near-separation under the richer interaction structure. We therefore do not interpret the ACSIncome coefficient-level tests as primary evidence. Importantly, the \textbf{maxROR} results remain essentially unchanged between Model~1 and Model~2, so the substantive-fairness conclusions are stable for ACSIncome as well. For RAVDESS, the augmented model did not fit stably, which is unsurprising given the much smaller effective test set size of 360 (as reported in \Cref{table-ravdess-group-counts}), so we omit RAVDESS from this robustness check. 

Taken together, the coefficient-level tests and the \textbf{maxROR} comparisons provide complementary evidence that our main downstream-fairness conclusions are not driven by a group-varying adoption effect. Overall, these results support the robustness of our reported substantive-fairness trends among CP methods to the adoption specification.

\begin{table}[H]
\centering
\small
\setlength{\tabcolsep}{9pt}
\caption{
Wald-test $p$-values for adoption-by-group interaction terms in Model~2. 
The null hypothesis is that the corresponding adoption-by-group interaction coefficient equals zero. 
Statistical significance is assessed at level $0.05$.
Baseline group of FACET, BiosBias, and ACSIncome is Middle, Female, and All Other Races (Aggregated), respectively.
}
\label{tab:adoption-interaction-pvalues}
\begin{tabular}{llc}
\toprule
Dataset & Interaction term & $p$-value \\
\midrule
FACET & adoption \(\times\) Older & .896 \\
      & adoption \(\times\) Unknown & .454 \\
      & adoption \(\times\) Younger & .812 \\
\midrule
BiosBias & adoption \(\times\) Male & .425 \\
\midrule
ACSIncome & adoption \(\times\) Asian alone & .128 \\
          & adoption \(\times\) Black or African American alone & .824\(^\dagger\) \\
          & adoption \(\times\) Two or More Races & .552\(^\dagger\) \\
          & adoption \(\times\) White alone & .119 \\
\bottomrule
\end{tabular}

\vspace{2pt}
\begin{minipage}{0.95\linewidth}
\footnotesize
\(^\dagger\) For ACSIncome, some adoption-by-group interaction coefficients had extremely large estimates and standard errors, suggesting numerical instability caused by sparse subgroup cells under the augmented interaction model. We therefore do not rely on coefficient-level inference for ACSIncome; instead, we use the \textbf{maxROR} comparison in \Cref{tab:adoption-interaction-maxror} as the primary robustness check.
\end{minipage}
\end{table}

\begin{table}[H]
\centering
\small
\setlength{\tabcolsep}{5pt}
\caption{
Comparison of \textbf{maxROR} (\%) computed from Model~1 and Model~2. 
Model~1 is the GEE used in the main experiments, with adoption included as a main effect. 
Model~2 additionally includes an adoption-by-group interaction. 
The \textbf{maxROR} values are nearly unchanged, indicating that our main substantive-fairness conclusions are not driven by collider-related distortion from the adoption specification. 
}
\label{tab:adoption-interaction-maxror}
\begin{sc}
\begin{tabular}{llcc}
\toprule
Dataset & Treatment & Model~1 \textbf{maxROR} (\%) & Model~2 \textbf{maxROR} (\%) \\
\midrule
FACET & Marginal & 9.0 & 9.1 \\
      & Mondrian & 37.7 & 37.4 \\
      & Label-Clustered & 13.5 & 14.0 \\
      & Group-Clustered & 14.5 & 14.3 \\
      & Backward & 8.5 & 8.4 \\
\midrule
BiosBias & Marginal & 6.9 & 6.9 \\
        & Mondrian & 8.1 & 8.1 \\
        & Label-Clustered & 1.6 & 1.6 \\
        & Group-Clustered & 12.5 & 12.4 \\
        & Backward & 0.3 & 0.3 \\
\midrule
ACSIncome & Marginal & 19.4 & 19.4 \\
          & Mondrian & 17.5 & 17.2 \\
          & Label-Clustered & 7.2 & 7.2 \\
          & Group-Clustered & 23.1 & 23.2 \\
          & Backward & 19.7 & 19.7 \\
\bottomrule
\end{tabular}
\end{sc}
\end{table}

\subsection{LLM-in-the-loop Evaluator Results on Different Tasks}
\label[appsec]{app:subsec-evaluator-result}

In \Cref{table-bios-llm-4omini-L3} through \Cref{table-acs-llm} we show more detailed statistics from LLM-in-the-loop experiments of \Cref{subsec:rq2_results} across datasets and CP methods. In particular, we provide one ablation on BiosBias where our standard LLM GPT-4o-mini was replaced with Llama-3.2-3B \cite{grattafiori2024llama}, demonstrating that qualitatively similar results can be obtained from distinct LLMs on the same task.

\begin{table}[H]
  \caption{Accuracy and fairness result on BiosBias experiment with \textit{GPT-4o-mini} as the LLM-in-the-loop.}
  \label{table-bios-llm-4omini-L3}
  \begin{center}
    \begin{small}
      \begin{sc}
        \begin{tabular}{lcccccr}
          \toprule
          Treatment     & Cvg (Gap)\%   & Size (Gap)  & Accuracy (Gap)\%       & maxROR\%         \\
          \midrule
          Control       &             &             &  78.96 (2.86)                &                \\
          Marginal      & 89.5 (2.75)     & 1.68 (.050)         &  80.71 (3.75)    & 6.9                 \\
          Mondrian      & 90.0 (.220)     & 1.80 (.474)         &  80.96 (3.90)    & 8.1                  \\
          Label-Clustered  & 90.2 (2.80) &  1.81 (.033)         &  79.72 (\textbf{2.55})  & \textbf{1.6}   \\
          Group-Clustered  & 90.2 (.193)  &  1.75 (.419)         &  \textbf{81.05} (4.53) & 12.5             \\
          Backward      & 91.5 (1.87)  &  2.50 (.025) &  79.41 (\textbf{2.87})              & \textbf{0.3}                   \\
          \bottomrule
        \end{tabular}
      \end{sc}
    \end{small}
  \end{center}
  \vskip -0.1in
\end{table}

\begin{table}[H]
  \caption{Accuracy and fairness result on BiosBias experiment with \textit{Llama-3.2-3B} as the LLM-in-the-loop.}
  \label{table-bios-llm-3b}
  \begin{center}
    \begin{small}
      \begin{sc}
        \begin{tabular}{lcccccr}
          \toprule
          Treatment     & Cvg (Gap)\%   & Size (Gap)  & Accuracy (Gap)\%        & maxROR\%         \\
          \midrule
          Control       &             &             & 67.54 (4.97)          &                \\
          Marginal      & 89.7 (2.76) & 1.69 (.027) & \textbf{76.03} (3.28) & 50.0          \\
          Mondrian      & 89.1 (0.00)  & 1.70 (.435) & 75.98 (5.03)          & 66.3          \\
          Label-Clustered & 90.1 (.019) & 1.78 (.067) & 74.83 (\textbf{1.55})          & \textbf{36.0} \\
          Group-Clustered & 90.3 (.010) & 1.74 (.343) & 75.74 (4.30)          & 59.3          \\
          Backward      & 91.5 (.018) & 2.49 (.016) & 71.19 (\textbf{1.38}) & \textbf{36.2}          \\
          \bottomrule
        \end{tabular}
      \end{sc}
    \end{small}
  \end{center}
  \vskip -0.1in
\end{table}

\begin{table}[H]
  \caption{Accuracy and fairness result on RAVDESS experiment with \textit{GPT-4o-audio-preview} as the LLM-in-the-loop.}
  \label{table-ravdess-4o-L2}
  \begin{center}
    \begin{small}
      \begin{sc}
        \begin{tabular}{lcccccr}
          \toprule
          Treatment     & Cvg (Gap)\%   & Size (Gap)  & Accuracy (Gap)\% & maxROR\% \\
          \midrule
          Control       &             &             &  21.11 (0.67)            &        \\
          Marginal      & 88.33 (5.56) & 1.89 (.039) &  44.28 (\textbf{1.00})     &  \textbf{10.5}      \\
          Mondrian   & 87.50 (.556) & 1.86 (.578)  &  44.56 (12.0)     &  79.2      \\
          Label-Clustered & 87.78 (2.22) & 1.92 (.011) &  \textbf{46.22} (\textbf{1.33})     &  \textbf{12.1}      \\
          Group-Clustered & 87.50 (.556) & 1.90 (.594) &  42.94 (15.44)     &  110.3      \\
          Backward      & 91.94 (8.33) & 2.48 (.011) &  39.22 (2.22)      &  17.0      \\
          \bottomrule
        \end{tabular}
      \end{sc}
    \end{small}
  \end{center}
  \vskip -0.1in
\end{table}

\begin{table}[H]
  \caption{Accuracy and fairness result on FACET experiment with \textit{Qwen2.5-VL-7B-Instruct} as the LLM-in-the-loop.}
  \label{table-facet-llm}
  \begin{center}
    \begin{small}
      \begin{sc}
        \begin{tabular}{lcccccr}
          \toprule
          Treatment     & Cvg (Gap)\%   & Size (Gap)  & Accuracy (Gap)\%                  & maxROR\%         \\
          \midrule
          Control       &             &             & 74.04 (18.00)                   &                \\
          Marginal      & 89.9 (10.7) & 2.62 (.738) & 76.91 (\textbf{16.27})                   & \textbf{9.0}          \\
          Mondrian      & 89.9 (3.02) & 2.69 (2.68) & 77.06 (21.71)                   & 37.7          \\
          Label-Clustered & 89.1 (7.46) & 2.92 (.356) & \textbf{78.81} (\textbf{14.47}) & 13.5          \\
          Group-Clustered & 89.1 (8.14) & 2.50 (1.28) & 77.22 (18.05)                   & 14.5          \\
          Backward      & 90.3 (8.85) & 3.50 (.053) & 75.54 (18.39)                   & \textbf{8.5} \\
          \bottomrule
        \end{tabular}
      \end{sc}
    \end{small}
  \end{center}
  \vskip -0.1in
\end{table}

\begin{table}[H]
  \caption{Accuracy and fairness result on ACSIncome experiment with \textit{Qwen2.5-7B} as the LLM-in-the-loop.}
  \label{table-acs-llm}
  \begin{center}
    \begin{small}
      \begin{sc}
        \begin{tabular}{lcccccr}
          \toprule
          Treatment     & Cvg (Gap)\%   & Size (Gap)  & Accuracy (Gap)\%        & maxROR\%         \\
          \midrule
          Control       &             &             & 14.70 (2.22)          &                \\
          Marginal      & 89.8 (3.18) & 5.35 (.256) & \textbf{20.41} (3.15) & 19.4          \\
          Mondrian      & 89.5 (3.87) & 7.16 (1.08) & 14.91 (\textbf{1.59}) & 17.5          \\
          Label-Clustered & 89.9 (3.67) & 5.33 (.270) & 18.68 (\textbf{2.88}) & \textbf{7.2} \\
          Group-Clustered & 89.8 (2.88) & 5.37 (.410) & 19.02 (5.05)          & 23.1          \\
          Backward      & 92.3 (3.03) & 6.50 (.010) & 15.02 (4.37)          & 19.7          \\
          \bottomrule
        \end{tabular}
      \end{sc}
    \end{small}
  \end{center}
  \vskip -0.1in
\end{table}

\subsection{Bootstrap Results}
\label[appsec]{app:bootstrap-results}

To provide uncertainty estimates on the \textbf{maxROR} metric, we performed bootstrap sampling over the LLM's task predictions, with results shown in \Cref{table-bios-llm-4omini-L3-bootstrap} through \Cref{table-acs-llm-bootstrap}.

\begin{table}[H]
  \caption{Mean accuracy (gap) and mean $\pm$ one standard error
           for maxROR over 1,000 resamples for BiosBias experiment on \Cref{table-bios-llm-4omini-L3}}
  \label{table-bios-llm-4omini-L3-bootstrap}
  \begin{center}
    \begin{small}
      \begin{sc}
        \begin{tabular}{lcccr}
          \toprule
          Treatment     & Accuracy (Gap)\%                              & maxROR\%                   \\
          \midrule
          Control       &  79.01 (2.84) &                           \\
          Marginal      &  80.79 (3.69)          &  9.0 $\pm$ .22                      \\
          Mondrian      &  81.06 (3.87)          &  9.7 $\pm$ .22          \\
          Label-Clustered &  79.78 (\textbf{2.54}) &  \textbf{6.1}  $\pm$ .16            \\
          Group-Clustered &  81.15 (4.49) &  13.1 $\pm$ .26        \\
          Backward      &  79.49 (\textbf{2.82}) &  \textbf{6.5} $\pm$ .16                           \\
          \bottomrule
        \end{tabular}
      \end{sc}
    \end{small}
  \end{center}
  \vskip -0.1in
\end{table}

\begin{table}[H]
  \caption{Mean accuracy (gap) and mean $\pm$ one standard error
           for maxROR over 1,000 resamples for RAVDESS experiment on \Cref{table-ravdess-4o-L2}}
  \label{table-ravdess-4o-L2-bootstrap}
  \begin{center}
    \begin{small}
      \begin{sc}
        \begin{tabular}{lcccr}
          \toprule
          Treatment     & Accuracy (Gap)\%                              & maxROR\%                  \\
          \midrule
          Control       & 21.07 (.840) &                           \\
          Marginal      & 44.26 (\textbf{.820})          & 24.6  $\pm$ 1.1                     \\
          Mondrian      & 44.56 (11.79)          & 70.0  $\pm$ 1.2          \\
          Label-Clustered & 46.09  (\textbf{1.29}) &  \textbf{23.4}  $\pm$ .78            \\
          Group-Clustered &  42.89 (15.25) & 92.5  $\pm$ 1.5           \\
          Backward      & 39.28 (2.17) & \textbf{18.4} $\pm$ .49                       \\
          \bottomrule
        \end{tabular}
      \end{sc}
    \end{small}
  \end{center}
  \vskip -0.1in
\end{table}

\begin{table}[H]
  \caption{Mean accuracy (gap) and mean $\pm$ one standard error
           for maxROR over 1,000 resamples for FACET experiment on \Cref{table-facet-llm}}
  \label{table-facet-llm-bootstrap}
  \begin{center}
    \begin{small}
      \begin{sc}
        \begin{tabular}{lcccr}
          \toprule
          Treatment     & Accuracy (Gap)\%                              & maxROR\%                  \\
          \midrule
          Control       & 73.64 (18.17) &                           \\
          Marginal      & 76.69 (\textbf{16.33})          & \textbf{23.4}  $\pm$ .38                  \\
          Mondrian      & 76.26 (21.76)          &  42.3 $\pm$ .60          \\
          Label-Clustered &  78.86 (\textbf{14.64}) & 28.5   $\pm$ .44        \\
          Group-Clustered &  77.04 (18.07) & 30.8  $\pm$ .47       \\
          Backward      & 75.07 (18.43) & \textbf{21.6} $\pm$ .35  \\
          \bottomrule
        \end{tabular}
      \end{sc}
    \end{small}
  \end{center}
  \vskip -0.1in
\end{table}

\begin{table}[H]
  \caption{Mean accuracy (gap) and mean $\pm$ one standard error
           for maxROR over 1,000 resamples for ACSIncome experiment on \Cref{table-acs-llm}}
  \label{table-acs-llm-bootstrap}
  \begin{center}
    \begin{small}
      \begin{sc}
        \begin{tabular}{lcccr}
          \toprule
          Treatment     & Accuracy (Gap)\%                              & maxROR\%                  \\
          \midrule
          Control       & 14.55 (2.27) &                           \\
          Marginal      & 20.29 (3.22)          &  32.6 $\pm$ .45  \\
          Mondrian      & 14.67 (\textbf{1.55})          & 30.9 $\pm$ .45          \\
          Label-Clustered &  18.38 (\textbf{2.90}) &  \textbf{23.4} $\pm$ .34        \\
          Group-Clustered &  18.19 (5.06) &  32.2 $\pm$  .43      \\
          Backward      & 15.10 (4.35) & 33.2 $\pm$ .45  \\
          \bottomrule
        \end{tabular}
      \end{sc}
    \end{small}
  \end{center}
  \vskip -0.1in
\end{table}

\clearpage

\section{Additional Tables and Plots}\label[appsec]{app:additional-plots}

\subsection{CP and LLM-in-the-loop Metrics by Group}

For further insights, in this section we provide additional tables and plots that break down our experimental data showing statistics conditional on group variables.

\begin{table}[H]
  \caption{Continuation of Table \ref{table-bios-llm-4omini-L3}. Results on BiosBias experiment for each group.}
  \label{table-bios-llm-4omini-L3-group}
  \begin{center}
    \begin{small}
      \begin{sc}
        \begin{tabular}{lccccr}
          \toprule
          Treatment     &  Group  & Cvg\%   & Size & Singleton\% & Accuracy\% \\
          \midrule
          \multirow{2}{8em}{Control} & Female & & & &  80.43           \\
          & Male & & & & 77.58 \\
          \midrule
          \multirow{2}{8em}{Marginal} & Female & 90.92 & 1.65 & 62.33 & 82.64       \\
          & Male & 88.17 & 1.70 & 60.14 & 78.89\\
          \midrule
          \multirow{2}{8em}{Mondrian} & Female & 89.89 & 1.56 & 65.33 &  82.97      \\
          & Male & 90.11 & 2.03 & 29.00 & 79.07\\
          \midrule
          \multirow{2}{9em}{Label-Clustered (K = 3)} & Female & 91.64 & 1.79 & 44.58 & 81.03     \\
          & Male & 88.85 & 1.82 & 43.16 & 78.49\\
          \midrule
          \multirow{2}{9em}{Group-Clustered (K = 2)} & Female & 90.30 & 1.53 & 69.66 &  83.38    \\
          & Male & 90.11 & 1.95 & 40.93 & 78.86 \\
          \midrule
          \multirow{2}{8em}{Backward} & Female & 92.47 & 2.49 & 0.00 &  80.89     \\
          & Male & 90.59 & 2.51 & 0.00 & 78.02\\
          \bottomrule
        \end{tabular}
      \end{sc}
    \end{small}
  \end{center}
  \vskip -0.1in
\end{table}

\begin{figure}[H]
  \centering

  \begin{subfigure}[t]{0.49\linewidth}
    \centering
    \includegraphics[width=\linewidth, height=0.22\textheight]{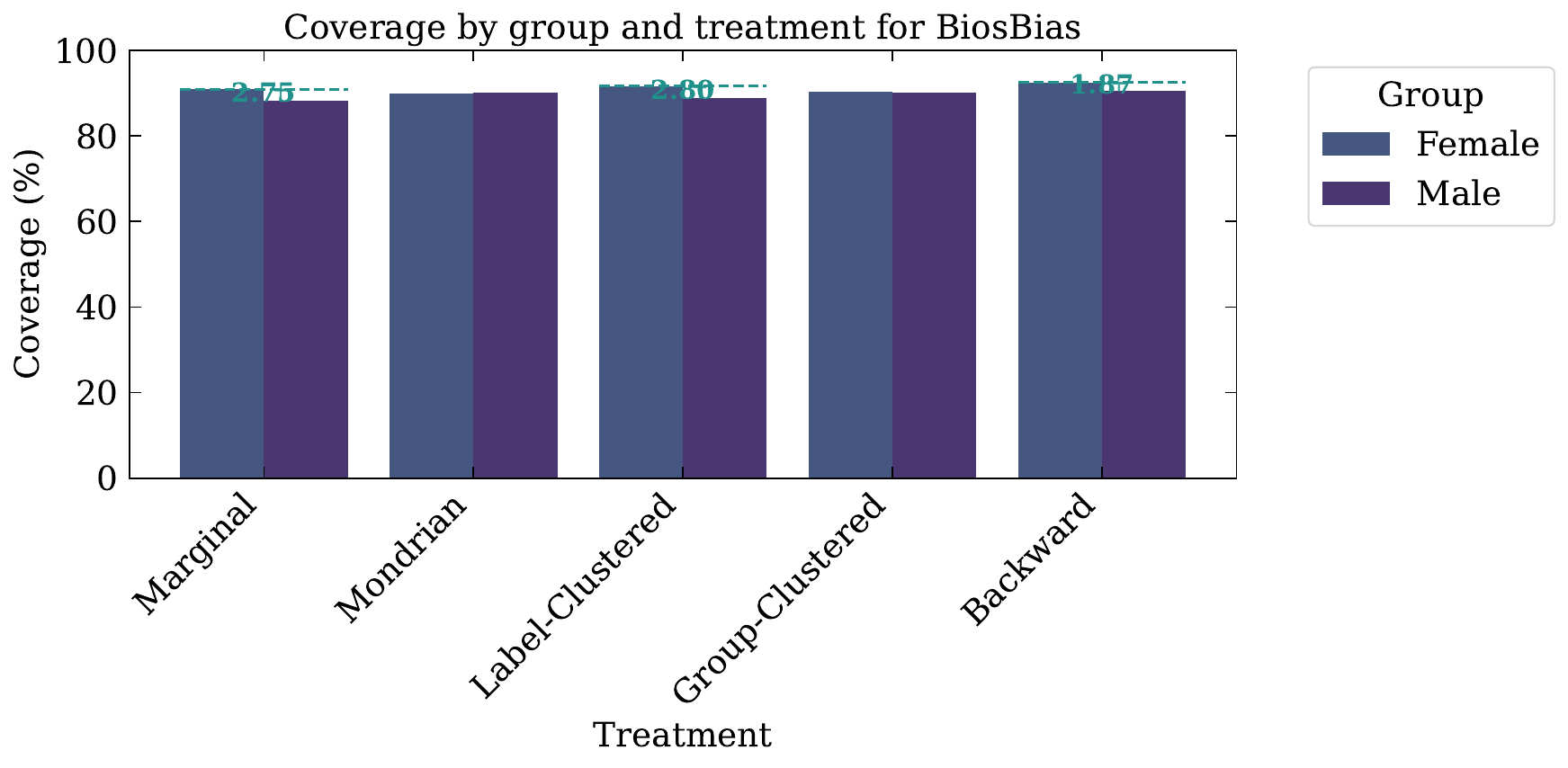}
    \caption{Coverage by group and treatment}
    \label{fig: BiosBias_4omini_L3_coverage}
  \end{subfigure}
  \hfill
  \begin{subfigure}[t]{0.49\linewidth}
    \centering
    \includegraphics[width=\linewidth, height=0.22\textheight]{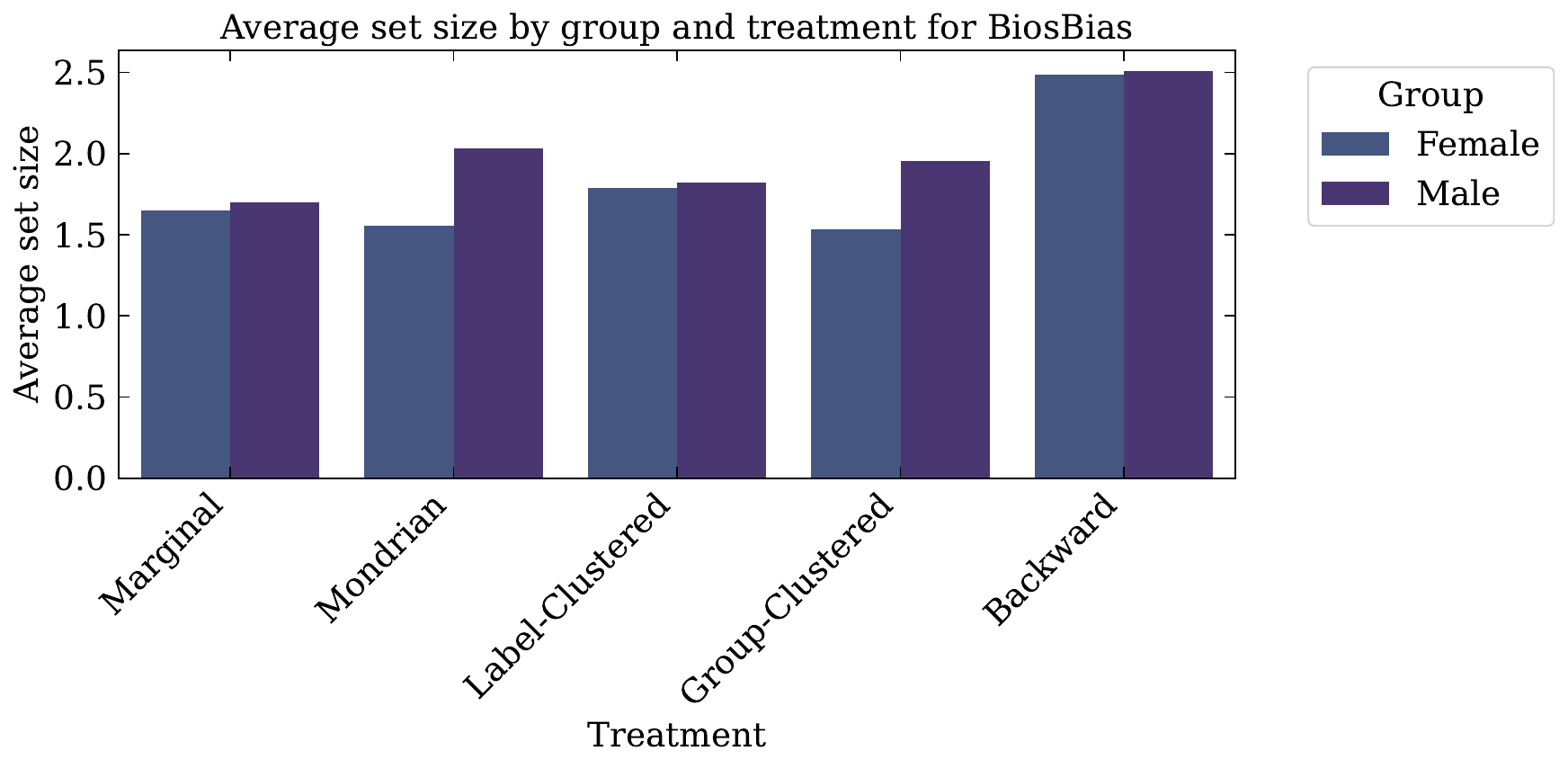}
    \caption{Average set size by group and treatment}
    \label{fig: BiosBias_4omini_L3_size}
  \end{subfigure}

 \vspace{0.3em}

  \begin{subfigure}[t]{0.49\linewidth}
    \centering
    \includegraphics[width=\linewidth, height=0.22\textheight]{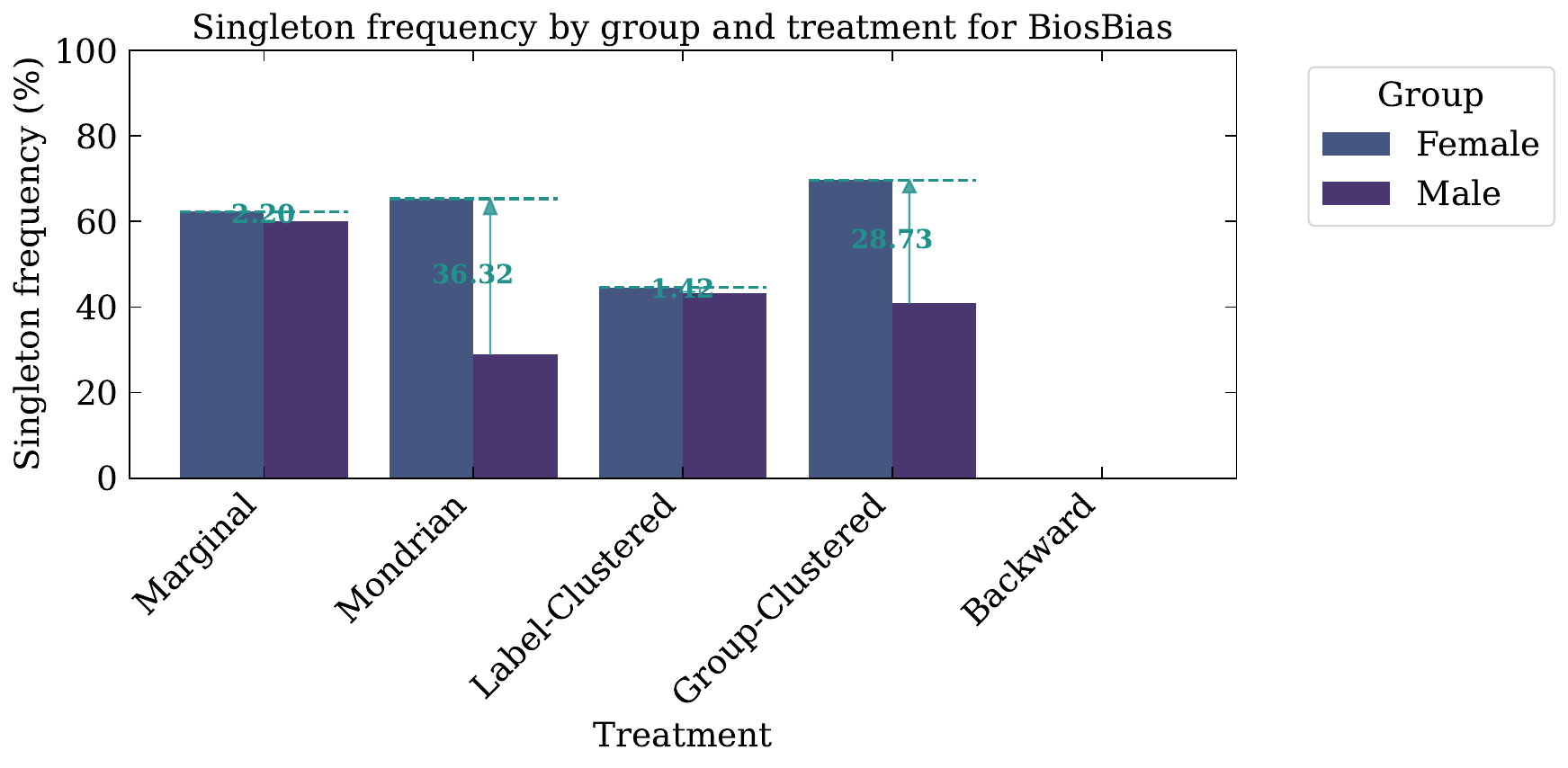}
    \caption{Singleton rate by group and treatment}
    \label{fig: BiosBias_4omini_L3_singleton}
  \end{subfigure}
  \hfill
  \begin{subfigure}[t]{0.49\linewidth}
    \centering
    \includegraphics[width=\linewidth, height=0.22\textheight]{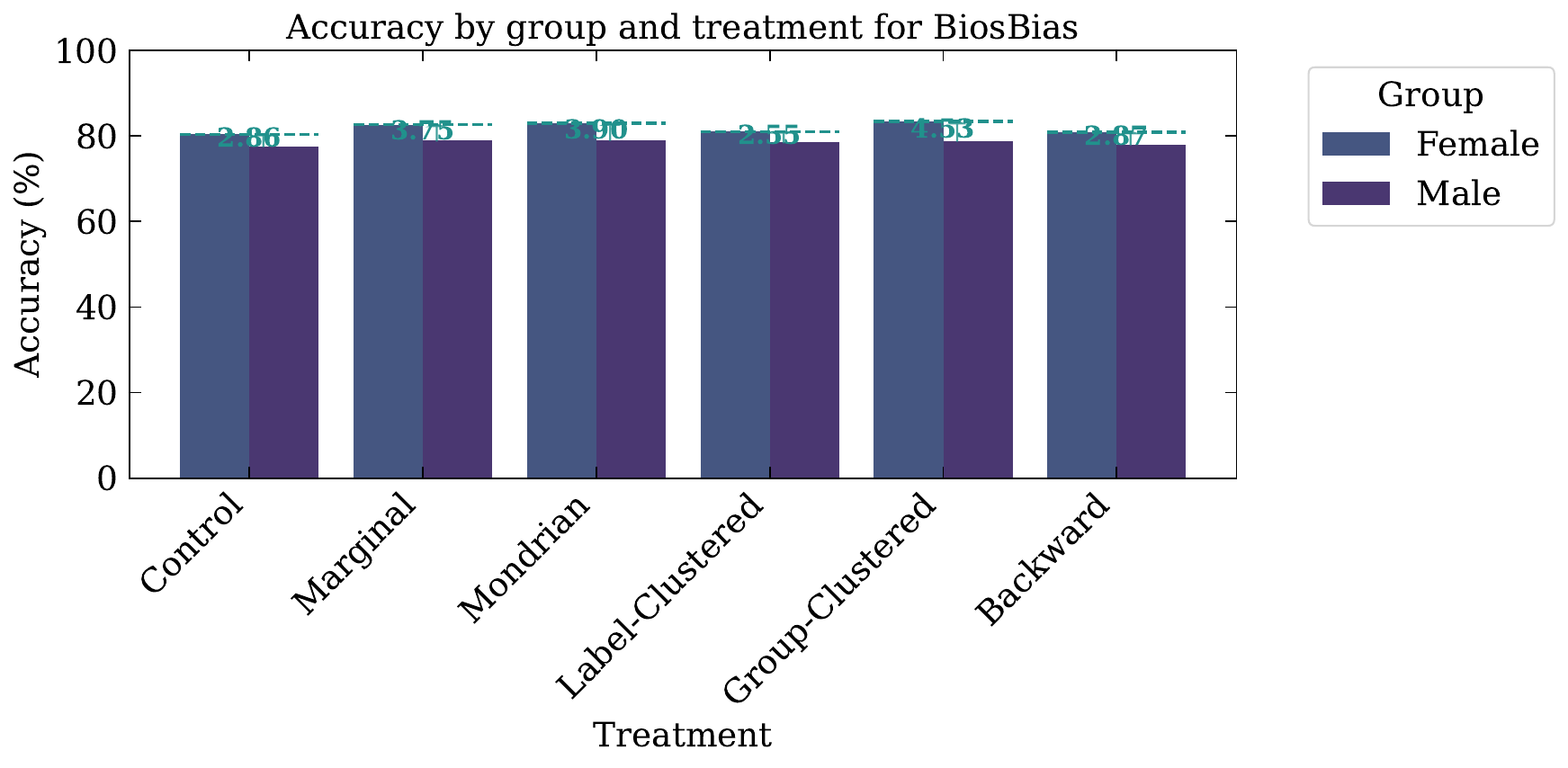}
    \caption{Accuracy rate by group and treatment}
    \label{fig: BiosBias_4omini_L3_acc}
  \end{subfigure}

  \caption{Experiment results of BiosBias with $K = 3$ in label-clustered CP and $K = 2$ in group-clustered CP. LLM-in-the-loop: GPT-4o-mini.}
  \label{fig: BiosBias_4omini_L3}
\end{figure}

%%%%%%%%%%%%%%%%RAVDESS%%%%%%%%%%%%%%%%%%%%%%%%%%%%%%%%
\begin{table}[H]
  \caption{Continuation of Table \ref{table-ravdess-4o-L2}. Results on RAVDESS experiment for each group.}
  \label{table-ravdess-4o-L2-group}
  \begin{center}
    \begin{small}
      \begin{sc}
        \begin{tabular}{lccccr}
          \toprule
          Treatment     &  Group  & Cvg\%   & Size & Singleton\% & Accuracy\% \\
          \midrule
          \multirow{2}{8em}{Control} & Female & & & & 20.78            \\
          & Male & & & & 21.44 \\
          \midrule
          \multirow{2}{8em}{Marginal} & Female & 91.11 & 1.91 & 37.78 & 44.78       \\
          & Male & 85.56 & 1.87 & 40.56 & 43.78\\
          \midrule
          \multirow{2}{8em}{Mondrian} & Female & 87.78 & 1.57 & 57.78 & 50.56       \\
          & Male & 87.22 & 2.15 & 22.22 & 38.56\\
          \midrule
          \multirow{2}{9em}{Label-Clustered (K = 2)} & Female & 88.89 & 1.92 & 30.56 & 46.89     \\
          & Male & 86.67 & 1.93 & 32.22 & 45.56\\
          \midrule
          \multirow{2}{9em}{Group-Clustered (K = 2)} & Female & 87.22 & 1.61 & 56.11 & 50.67     \\
          & Male & 87.78 & 2.20 & 20.00 & 35.22 \\
          \midrule
          \multirow{2}{8em}{Backward} & Female & 96.11 & 2.47 & 0.00 & 40.33      \\
          & Male & 87.78 & 2.48 & 0.00 & 38.11\\
          \bottomrule
        \end{tabular}
      \end{sc}
    \end{small}
  \end{center}
  \vskip -0.1in
\end{table}

\begin{figure}[H]
  \centering

  \begin{subfigure}[t]{0.49\linewidth}
    \centering
    \includegraphics[width=\linewidth, height=0.22\textheight]{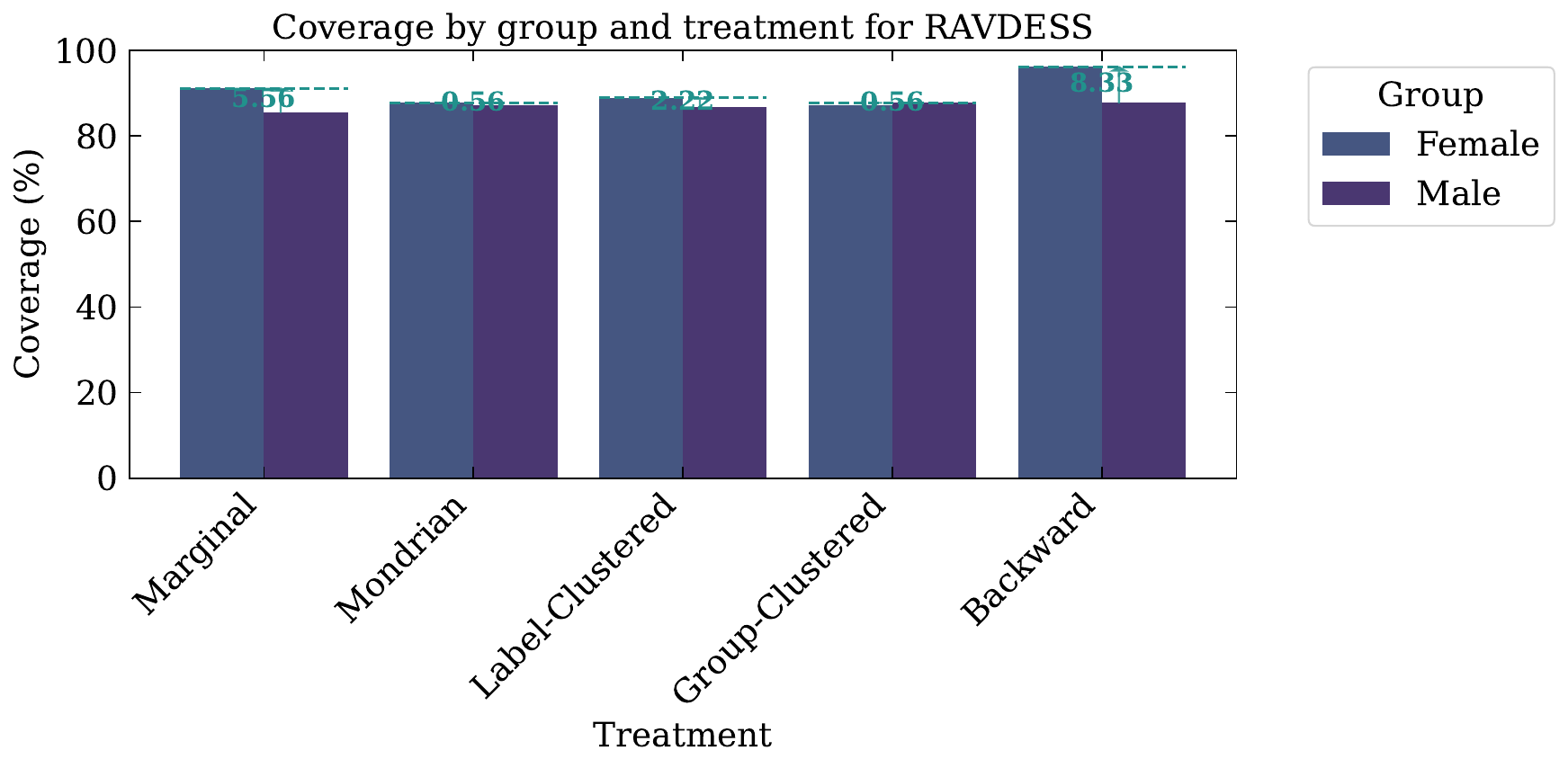}
    \caption{Coverage by group and treatment}
    \label{fig: RAVDESS_GPT-4o-L2_coverage}
  \end{subfigure}
  \hfill
  \begin{subfigure}[t]{0.49\linewidth}
    \centering
    \includegraphics[width=\linewidth, height=0.22\textheight]{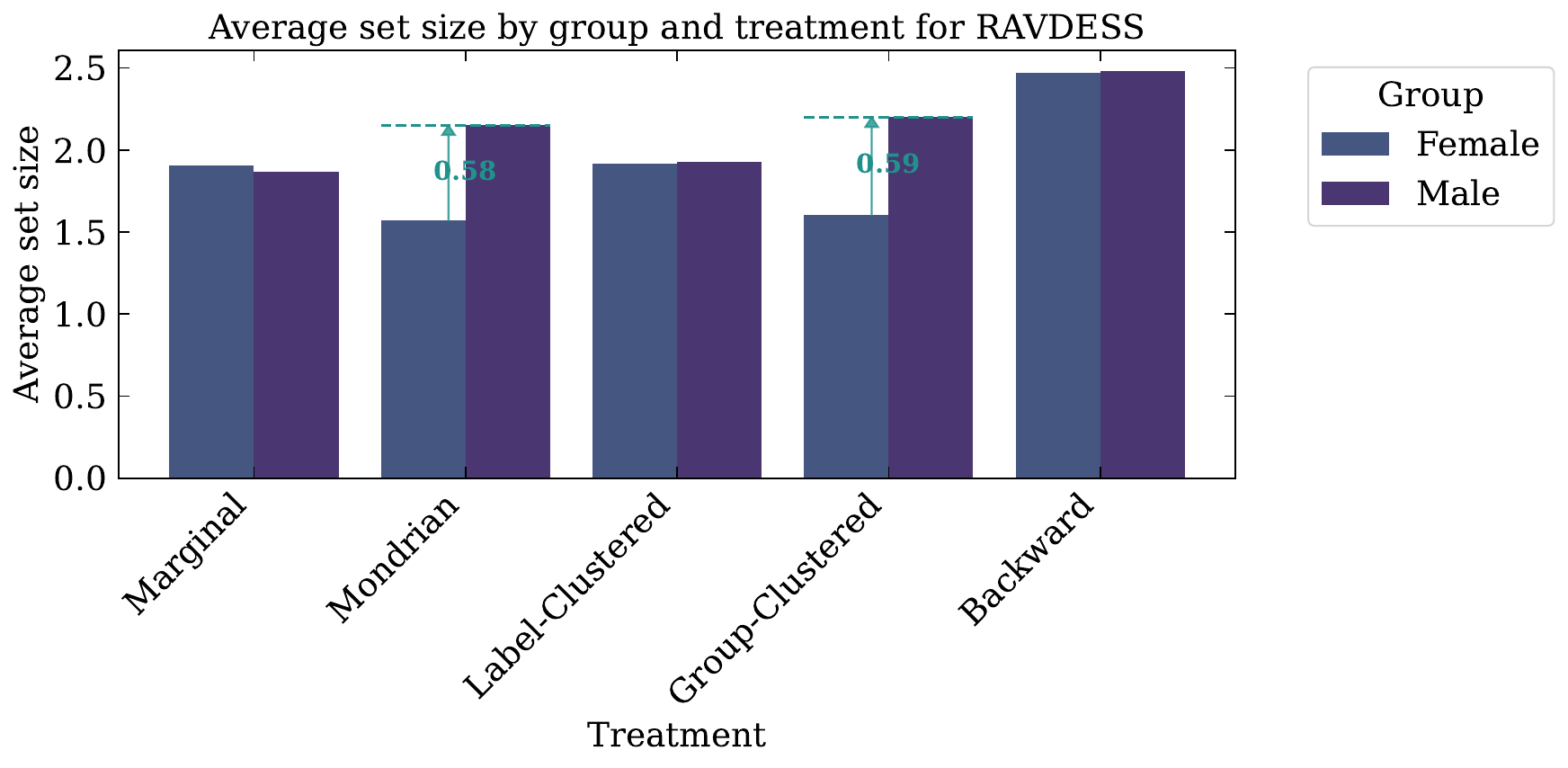}
    \caption{Average set size by group and treatment}
    \label{fig: RAVDESS_GPT-4o-L2_size}
  \end{subfigure}

 \vspace{0.3em}

  \begin{subfigure}[t]{0.49\linewidth}
    \centering
    \includegraphics[width=\linewidth, height=0.22\textheight]{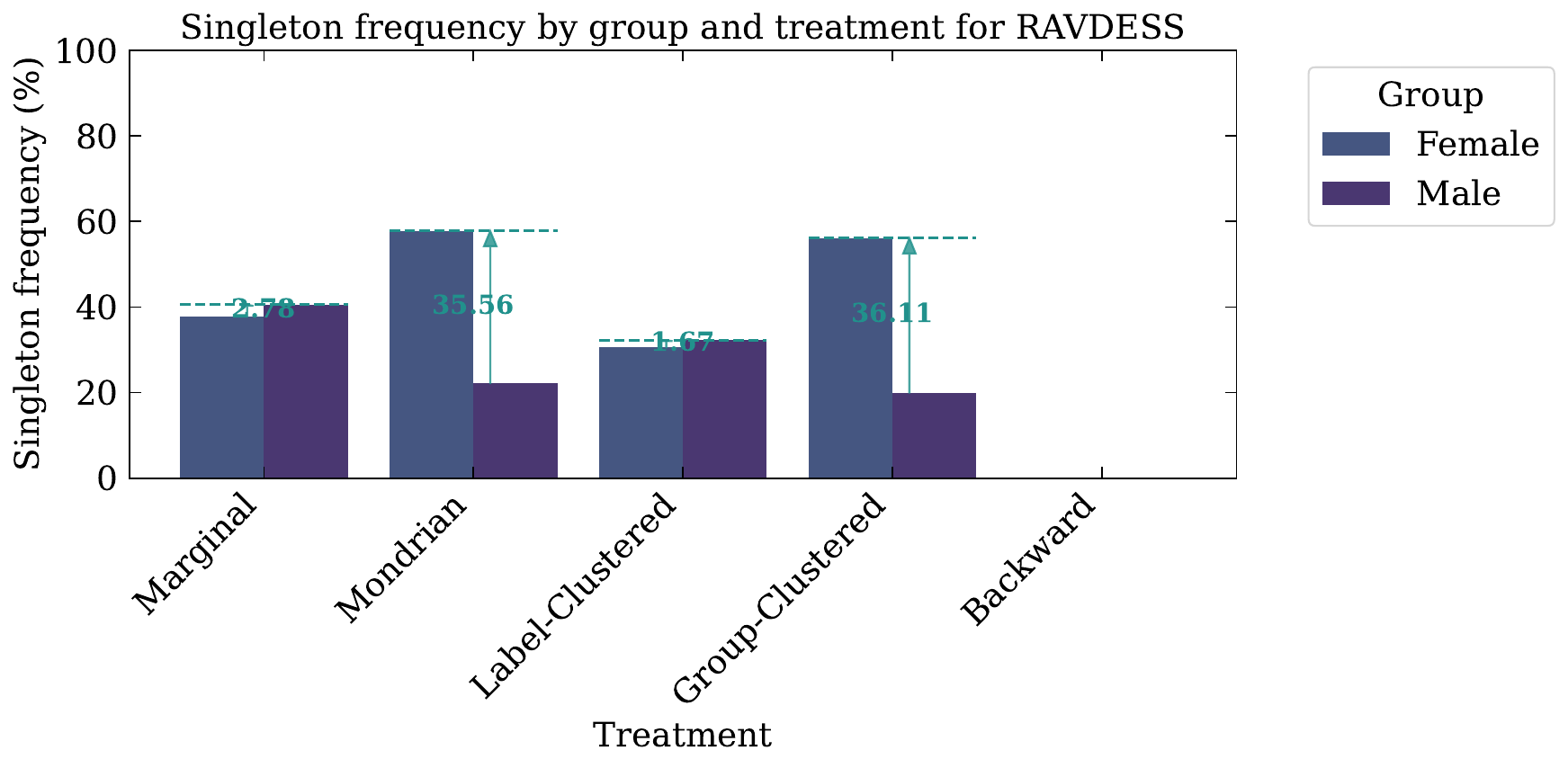}
    \caption{Singleton rate by group and treatment}
    \label{fig: RAVDESS_GPT-4o-L2_singleton}
  \end{subfigure}
  \hfill
  \begin{subfigure}[t]{0.49\linewidth}
    \centering
    \includegraphics[width=\linewidth, height=0.22\textheight]{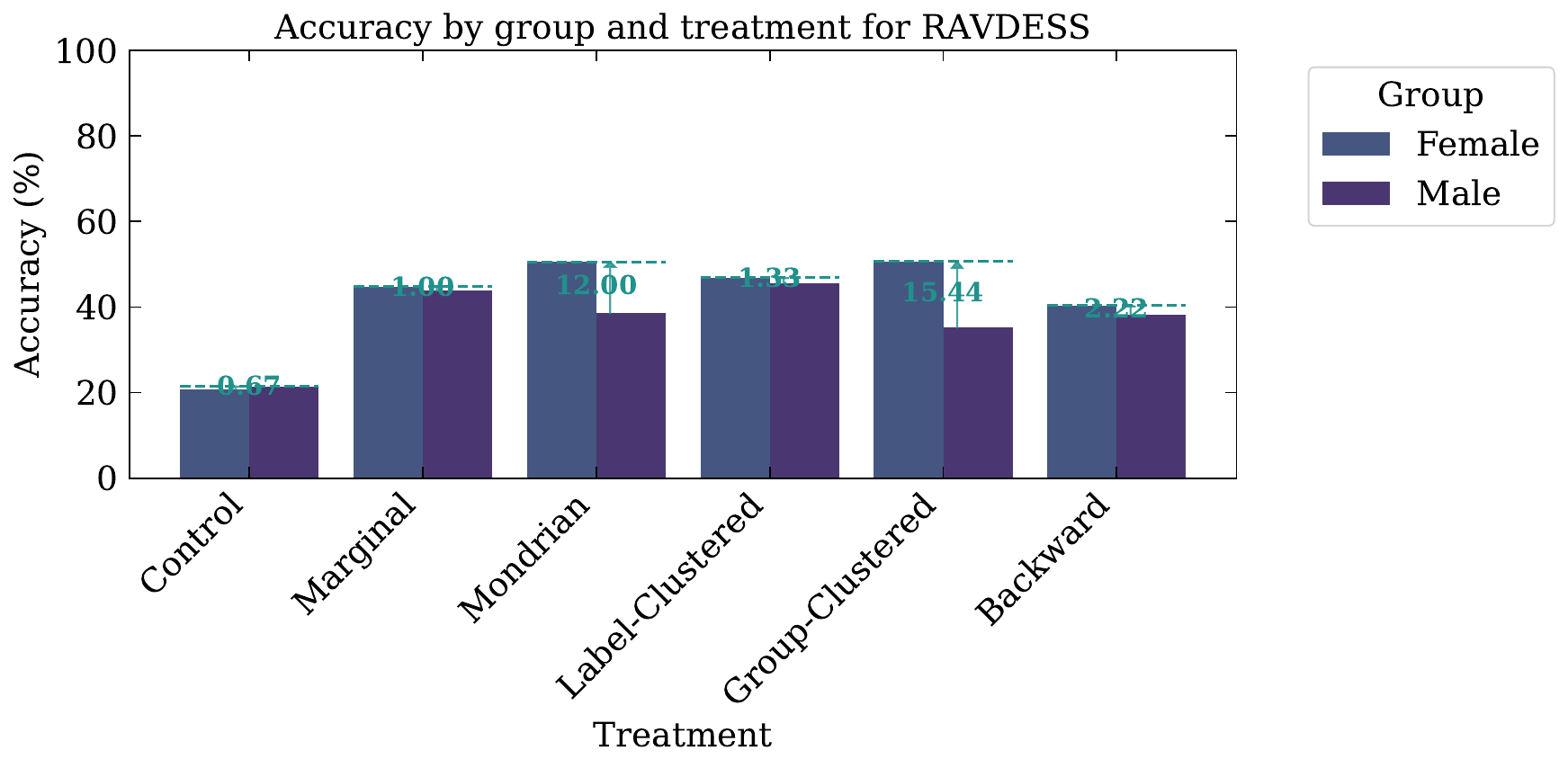}
    \caption{Accuracy rate by group and treatment}
    \label{fig: RAVDESS_GPT-4o-L2_acc}
  \end{subfigure}

  \caption{Experiment results of RAVDESS with $K = 2$ in both label-clustered and group-clustered CP. LLM-in-the-loop: GPT-4o-audio-preview.}
  \label{fig: RAVDESS_GPT-4o-L2}
\end{figure}

%%%%%%%%%%%%%%FACET%%%%%%%%%%%%%%%%%%%%%%%%%%%%%%%%%%%%%%%%%%%%%%%%%%%%%%
\begin{table}[H]
  \caption{Continuation of Table \ref{table-facet-llm}. Result on FACET experiment for each group.}
  \label{table-facet-llm-group}
  \begin{center}
    \begin{small}
      \begin{sc}
        \begin{tabular}{lccccr}
          \toprule
          Treatment     &  Group  & Cvg\%   & Size & Singleton\% & Accuracy\% \\
          \midrule
          \multirow{4}{8em}{Control} & Middle & & & &    71.40        \\
          & Older & & & & 65.25 \\
          & Unknown & & & & 74.63 \\
          & Younger & & & & 83.24 \\
          \midrule
          \multirow{4}{8em}{Marginal} & Middle & 88.61 & 2.72 & 29.22 &   74.49         \\
          & Older & 84.62 & 2.91 & 18.68 & 69.51 \\
          & Unknown & 89.47 & 2.72 & 28.95 & 76.89 \\
          & Younger & 95.29 & 2.17 & 50.36 & 85.78 \\
          \midrule
          \multirow{4}{8em}{Mondrian} & Middle & 89.16 & 2.93 & 24.14 &   74.21         \\
          & Older & 89.01 & 4.31 & 0.00 &  65.38\\
          & Unknown & 89.80 & 2.60 & 30.26 & 78.29 \\
          & Younger & 92.03 & 1.62 & 67.39 & 87.09 \\
          \midrule
          \multirow{4}{9em}{Label-Clustered (K = 2)} & Middle & 87.11 & 2.91 & 11.11 & 76.37           \\
          & Older & 90.11 & 3.12 & 5.49 & 72.94 \\
          & Unknown & 88.49 & 3.00 & 8.88 & 78.62 \\
          & Younger & 94.57 & 2.76 & 9.78 & 87.41 \\
          \midrule
          \multirow{4}{9em}{Group-Clustered (K = 2)} & Middle & 89.03 & 2.87 & 25.10 & 74.02           \\
          & Older & 84.62 & 3.04 & 15.38 & 68.41 \\
          & Unknown & 87.50 & 2.11 & 44.41 & 79.15 \\
          & Younger & 92.75 & 1.76 & 63.04 & 86.46 \\
          \midrule
          \multirow{4}{8em}{Backward} & Middle & 88.75 & 3.51 & 0.00 &  72.81          \\
          & Older & 85.71 & 3.53 & 0.00 & 66.07 \\
          & Unknown & 91.45 & 3.49 & 0.00 & 76.81 \\
          & Younger & 94.57 & 3.47 & 0.00 &  84.47\\
          \bottomrule
        \end{tabular}
      \end{sc}
    \end{small}
  \end{center}
  \vskip -0.1in
\end{table}

%%%%%%%%

\begin{figure}[H]
  \centering

  \begin{subfigure}[t]{0.49\linewidth}
    \centering
    \includegraphics[width=\linewidth, height=0.22\textheight]{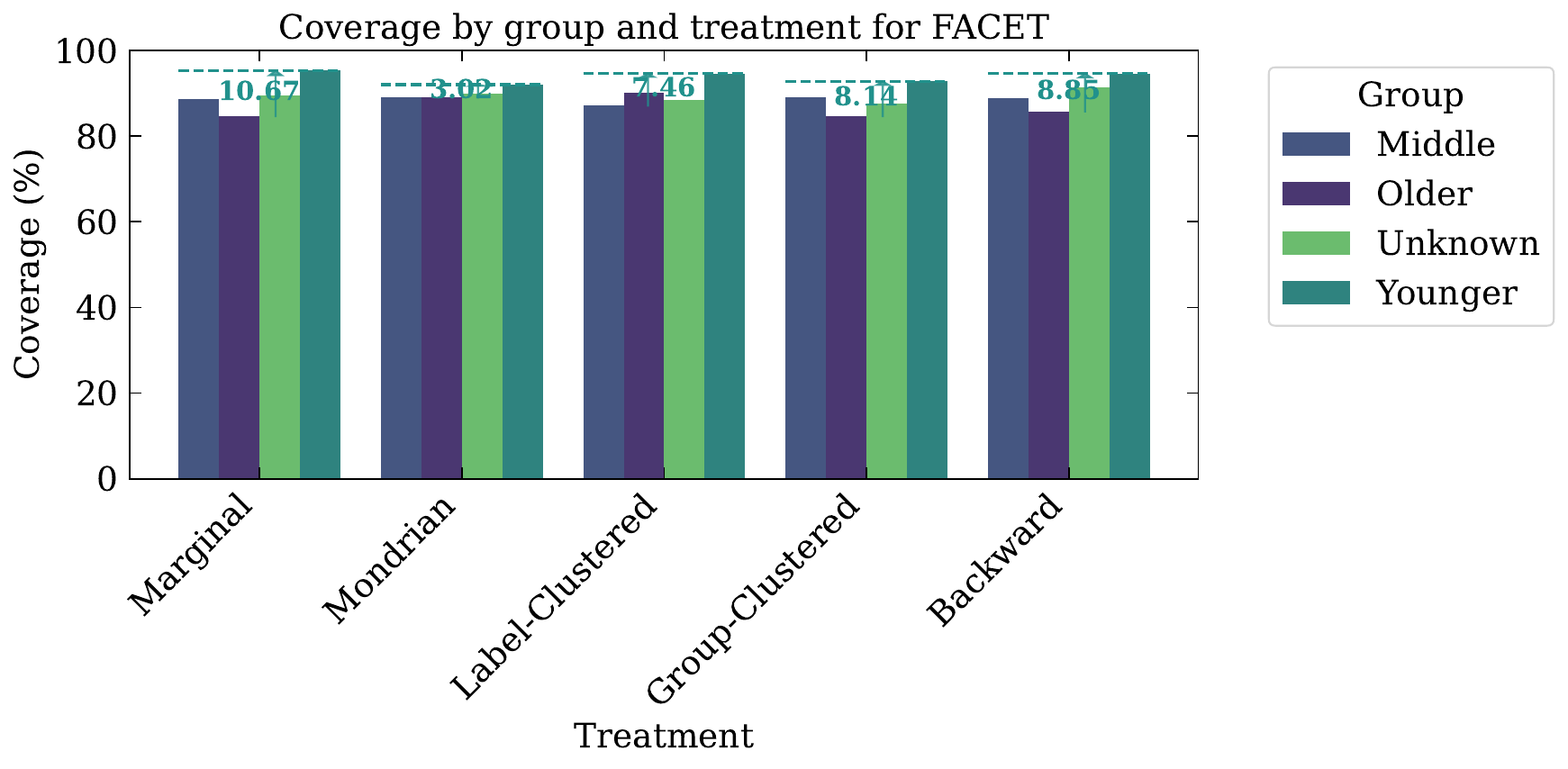}
    \caption{Coverage by group and treatment}
    \label{fig: FACET_coverage}
  \end{subfigure}
  \hfill
  \begin{subfigure}[t]{0.49\linewidth}
    \centering
    \includegraphics[width=\linewidth, height=0.22\textheight]{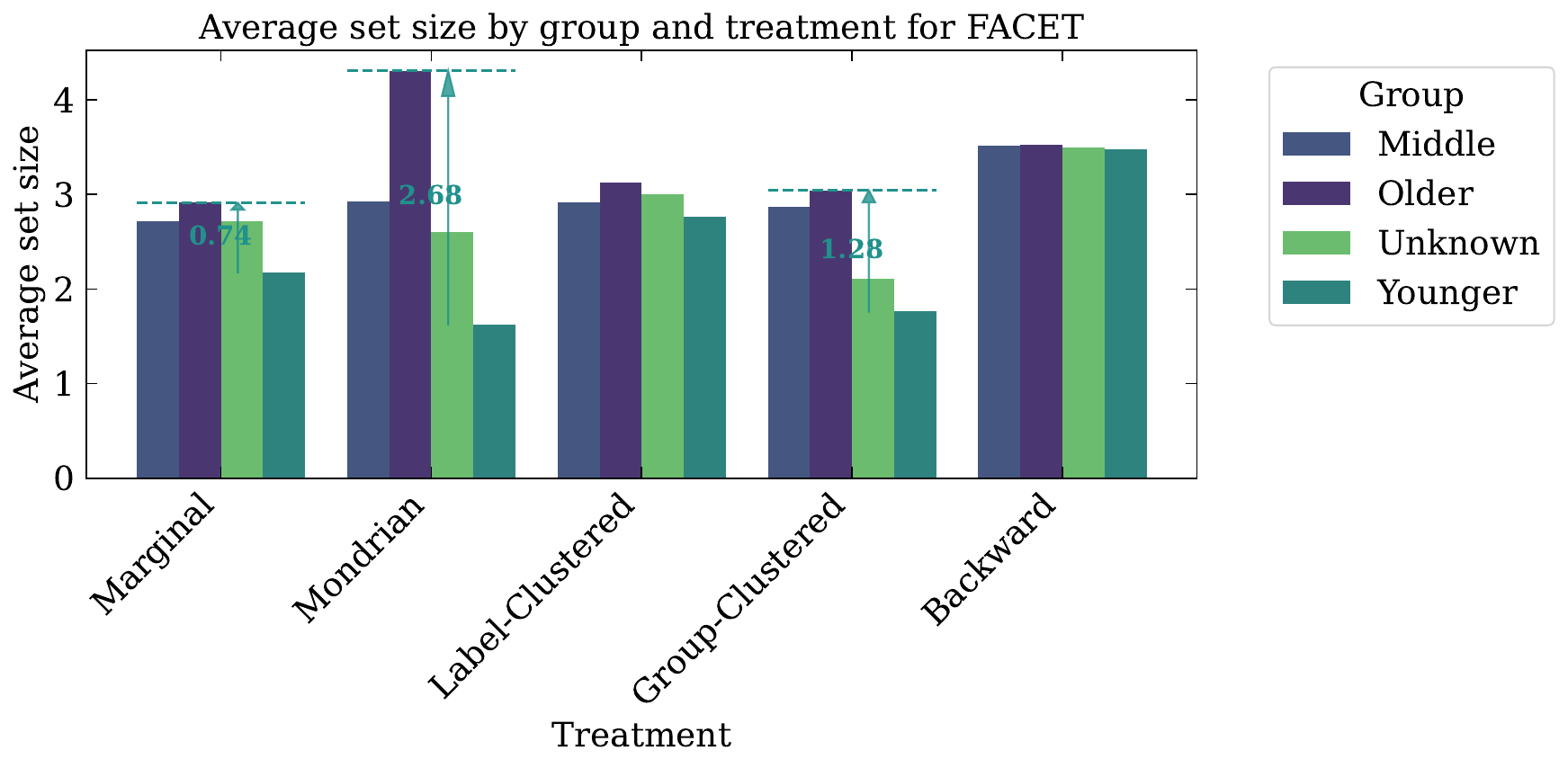}
    \caption{Average set size by group and treatment}
    \label{fig: FACET_size}
  \end{subfigure}
  
  \vspace{0.3em}
  
  \begin{subfigure}[t]{0.49\linewidth}
    \centering
    \includegraphics[width=\linewidth, height=0.22\textheight]{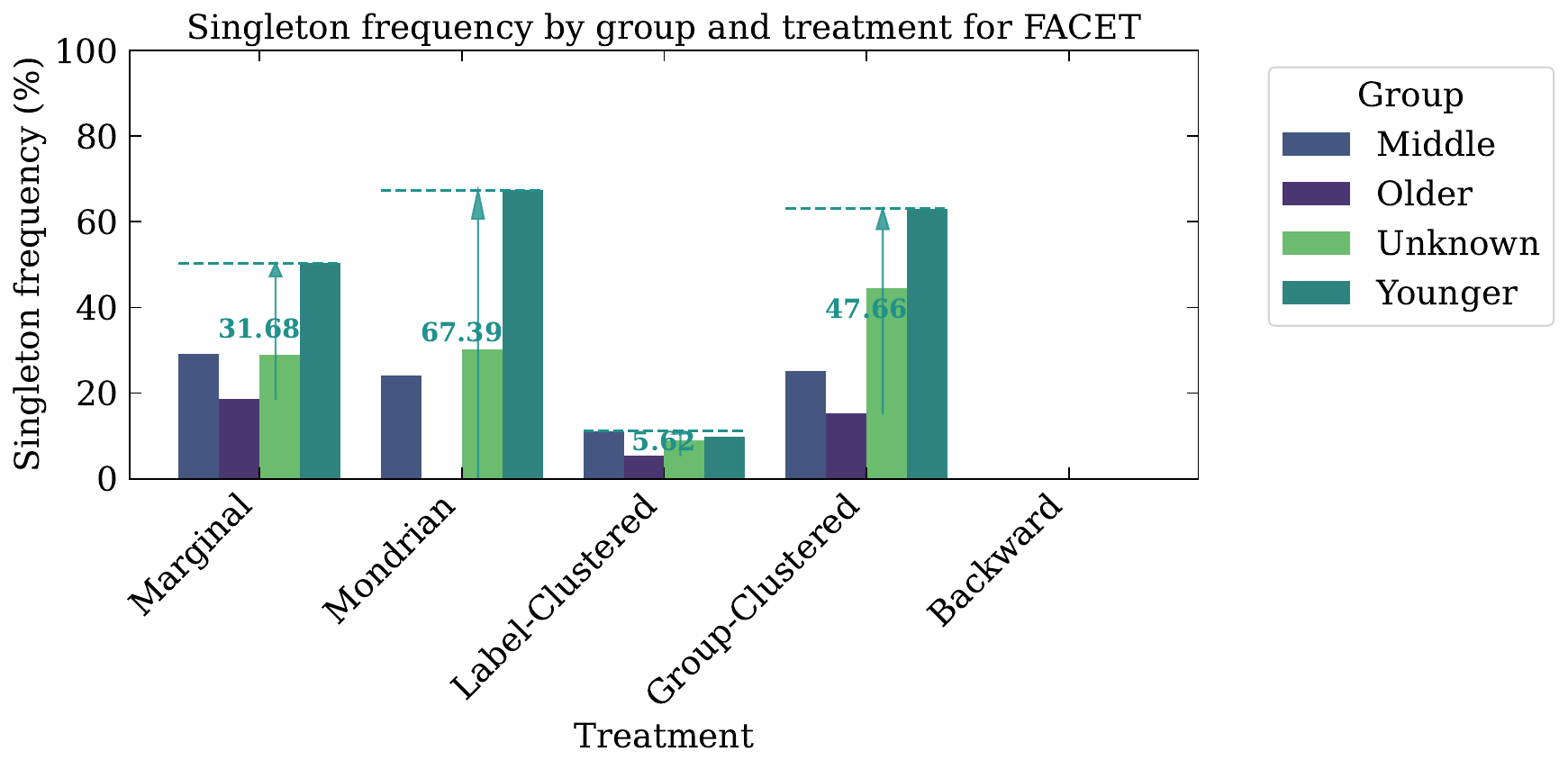}
    \caption{Singleton rate by group and treatment}
    \label{fig: FACET_singleton}
  \end{subfigure}
  \hfill
  \begin{subfigure}[t]{0.49\linewidth}
    \centering
    \includegraphics[width=\linewidth, height=0.22\textheight]{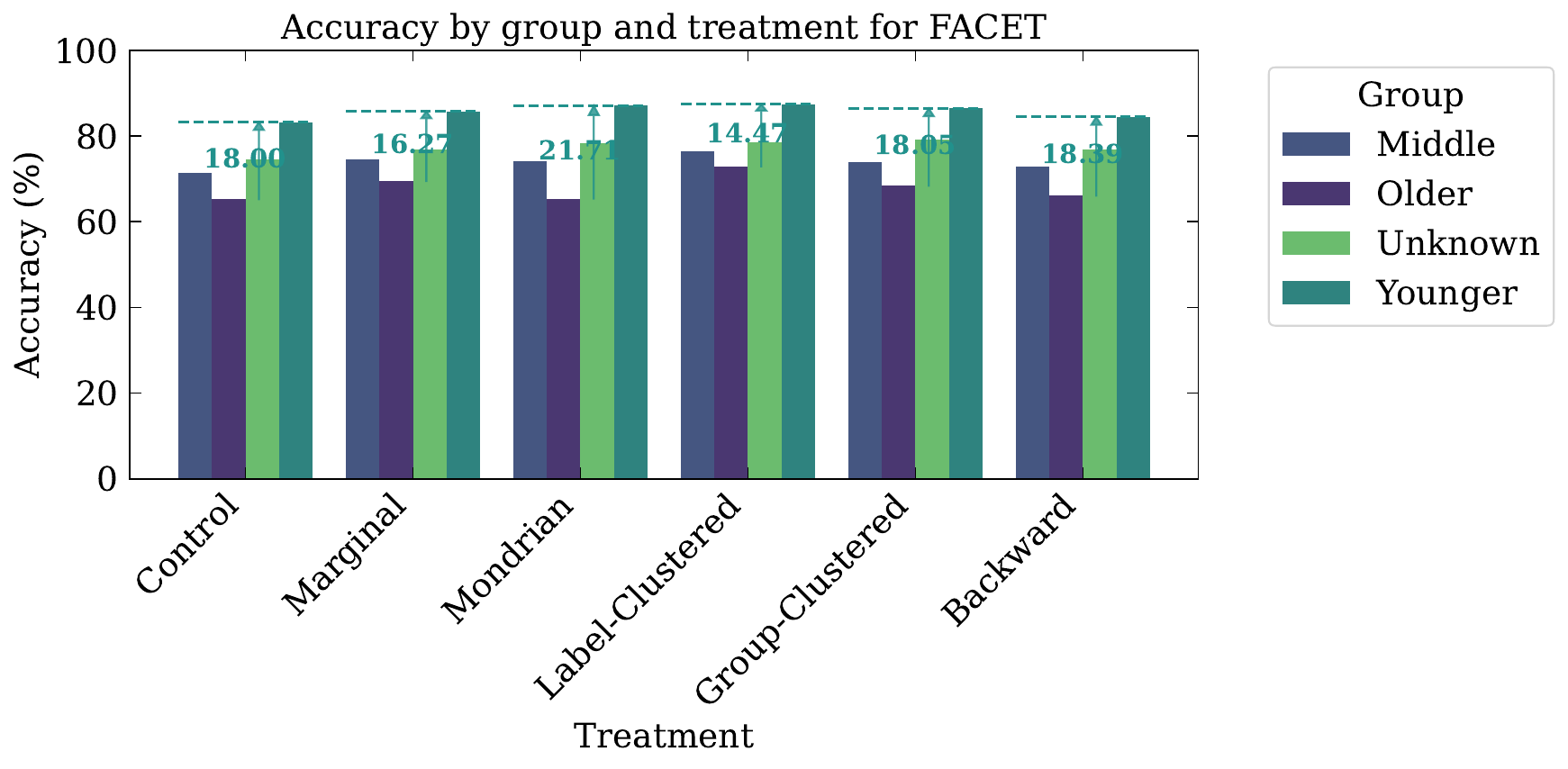}
    \caption{Accuracy rate by group and treatment}
    \label{fig: FACET_acc}
  \end{subfigure}

  \caption{Experiment results of FACET with $K = 2$ in both label-clustered and group-clustered CP. LLM-in-the-loop: Qwen2.5-VL-7B-Instruct.}
  \label{fig: FACET_result}
\end{figure}

%%%%%%%%%%%%%%%%%%%%%%%%ACSIncome%%%%%%%%%%%%%%%%%%%%%
\begin{table}[H]
  \caption{Continuation of Table \ref{table-acs-llm}. Result on ACSIncome experiment for each group.}
  \label{table-acsincome-llm-group}
  \begin{center}
    \begin{small}
      \begin{sc}
        \begin{tabular}{lccccr}
          \toprule
          Treatment     &  Group  & Cvg\%   & Size & Singleton\% & Accuracy\% \\
          \midrule
          \multirow{5}{8em}{Control} & All Other Races & & & &    15.12       \\
          & Asian & & & & 12.90 \\
          & Black or African American & & & & 15.08 \\
          & Two or More Races & & & & 15.02 \\
          & White &  &  &  & 14.72 \\
          \midrule
          \multirow{5}{8em}{Marginal} & All Other Races & 87.83 & 5.44 & .654 &  19.08         \\
          & Asian & 87.33 & 5.19 & 2.71 & 18.74 \\
          & Black or African American & 88.77 & 5.34 & 2.44 & 21.89 \\
          & Two or More Races & 89.35 & 5.33 & 2.78 & 21.44 \\
          & White & 90.51 & 5.37 & 1.87 & 20.37 \\
          \midrule
          \multirow{5}{8em}{Mondrian} & All Other Races & 90.58 & 7.98 & 0.00 &  15.58         \\
          & Asian & 87.18 & 6.90 & 0.00 & 13.90 \\
          & Black or African American & 90.11 & 7.67 & 0.00 & 14.50 \\
          & Two or More Races & 91.05 & 7.64 & 0.00 &  14.19\\
          & White & 89.33 & 6.95 & .015 & 15.10 \\
          \midrule
          \multirow{5}{9em}{Label-Clustered (K = 2)} & All Other Races & 87.70 & 5.43 & 0.00 & 18.36          \\
          & Asian & 87.03 & 5.16 & 0.00 & 16.48 \\
          & Black or African American & 88.52 & 5.32 & 0.00 & 19.35 \\
          & Two or More Races & 88.99 & 5.35 & 0.00 & 18.96 \\
          & White & 90.70  & 5.34 & 0.00 & 18.80 \\
          \midrule
          \multirow{5}{9em}{Group-Clustered (K = 2)} & All Other Races & 87.57 & 5.38 & .262 &  18.55         \\
          & Asian & 89.29 & 5.74 & 0.00 & 14.40 \\
          & Black or African American & 88.40 & 5.35 & .366 & 19.29 \\
          & Two or More Races & 88.81 & 5.35 & .537 & 19.32 \\
          & White & 90.45 & 5.33 & .362 & 19.46 \\
          \midrule
          \multirow{5}{8em}{Backward} & All Other Races & 90.97 & 6.49 & 0.00 &   16.59        \\
          & Asian & 89.89 & 6.49 & 0.00 & 12.25 \\
          & Black or African American & 90.84 & 6.49 & 0.00 & 15.23 \\
          & Two or More Races & 92.93 & 6.50 & 0.00 & 16.63 \\
          & White & 92.75 & 6.50 & 0.00 & 14.82 \\
          \bottomrule
        \end{tabular}
      \end{sc}
    \end{small}
  \end{center}
  \vskip -0.1in
\end{table}

\begin{figure}[H]
  \centering

  \begin{subfigure}[t]{0.49\linewidth}
    \centering
    \includegraphics[width=\linewidth, height=0.22\textheight]{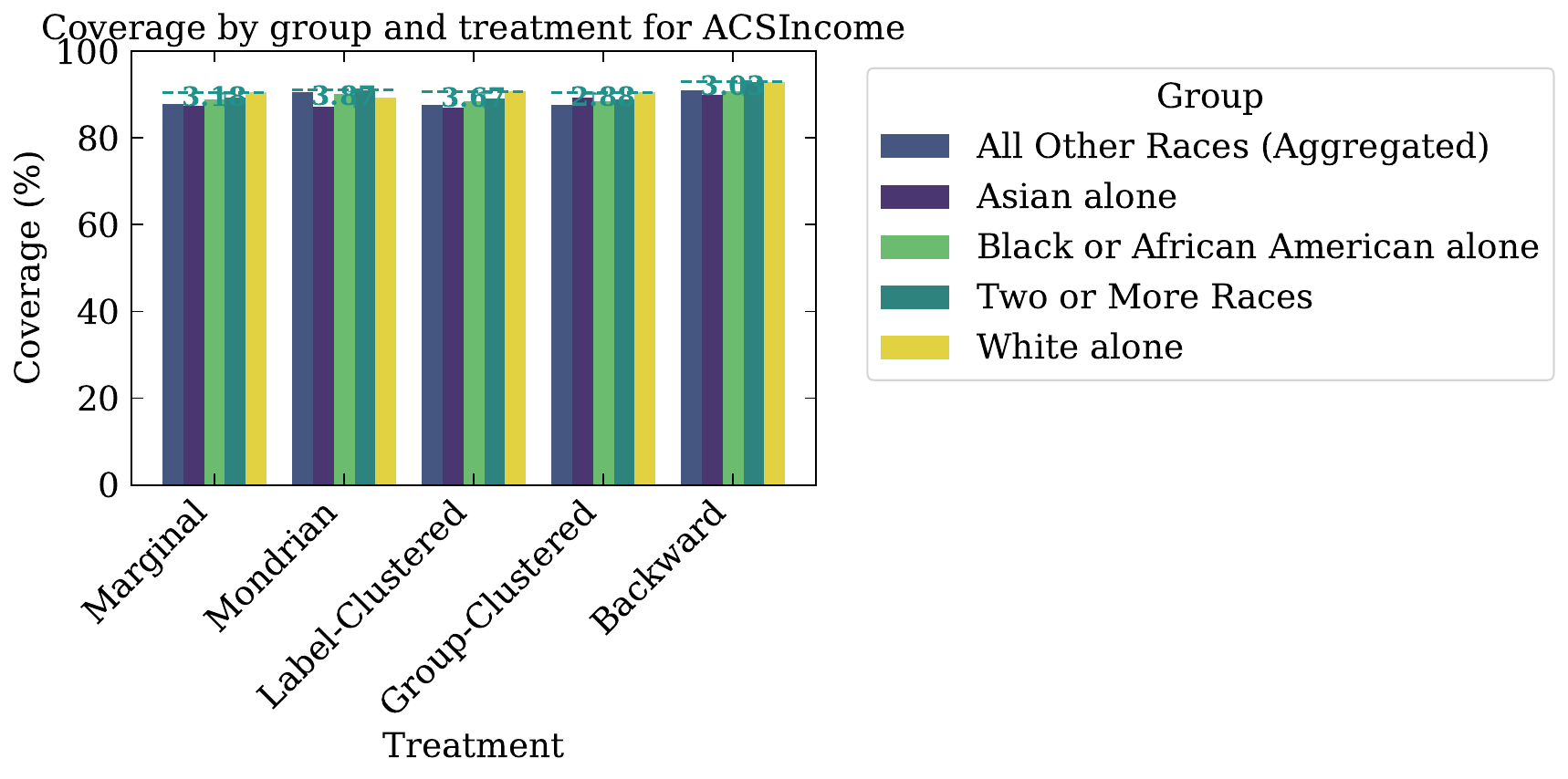}
    \caption{Coverage by group and treatment}
    \label{fig: ACSIncome_coverage}
  \end{subfigure}
  \hfill
  \begin{subfigure}[t]{0.49\linewidth}
    \centering
    \includegraphics[width=\linewidth, height=0.22\textheight]{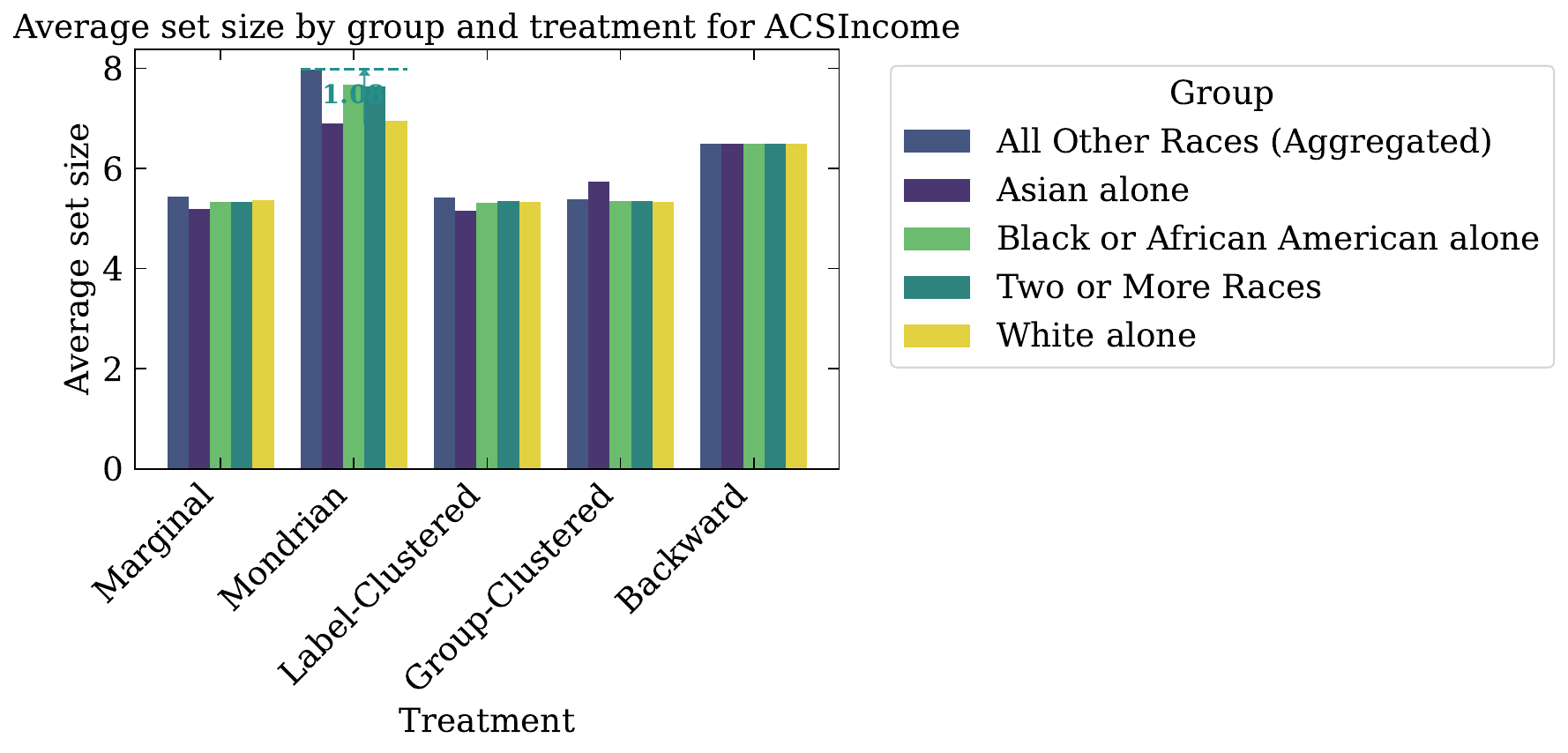}
    \caption{Average set size by group and treatment}
    \label{fig: ACSIncome_size}
  \end{subfigure}
  
  \vspace{0.3em}
  
  \begin{subfigure}[t]{0.49\linewidth}
    \centering
    \includegraphics[width=\linewidth, height=0.22\textheight]{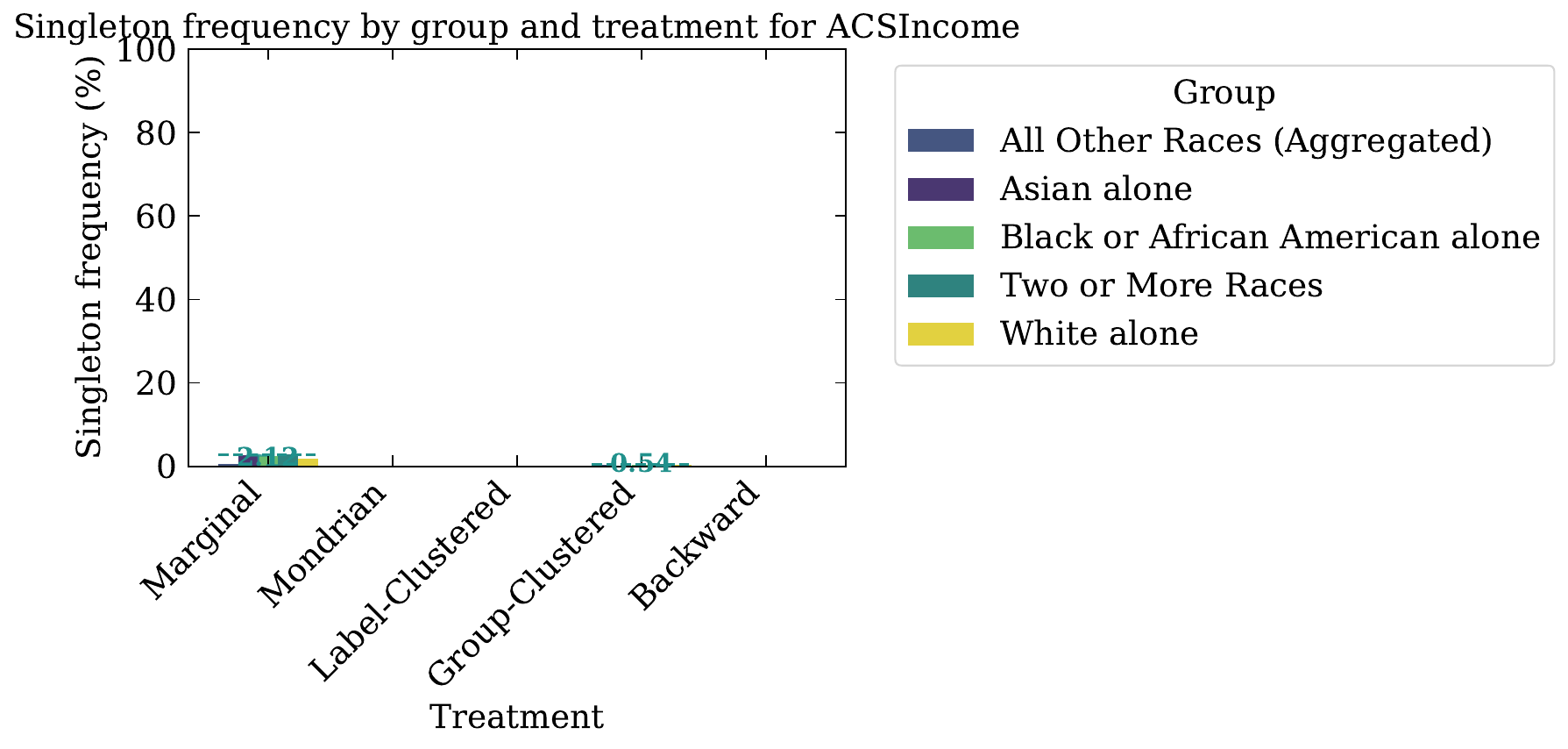}
    \caption{Singleton rate by group and treatment}
    \label{fig: ACSIncome_singleton}
  \end{subfigure}
  \hfill
  \begin{subfigure}[t]{0.49\linewidth}
    \centering
    \includegraphics[width=\linewidth, height=0.22\textheight]{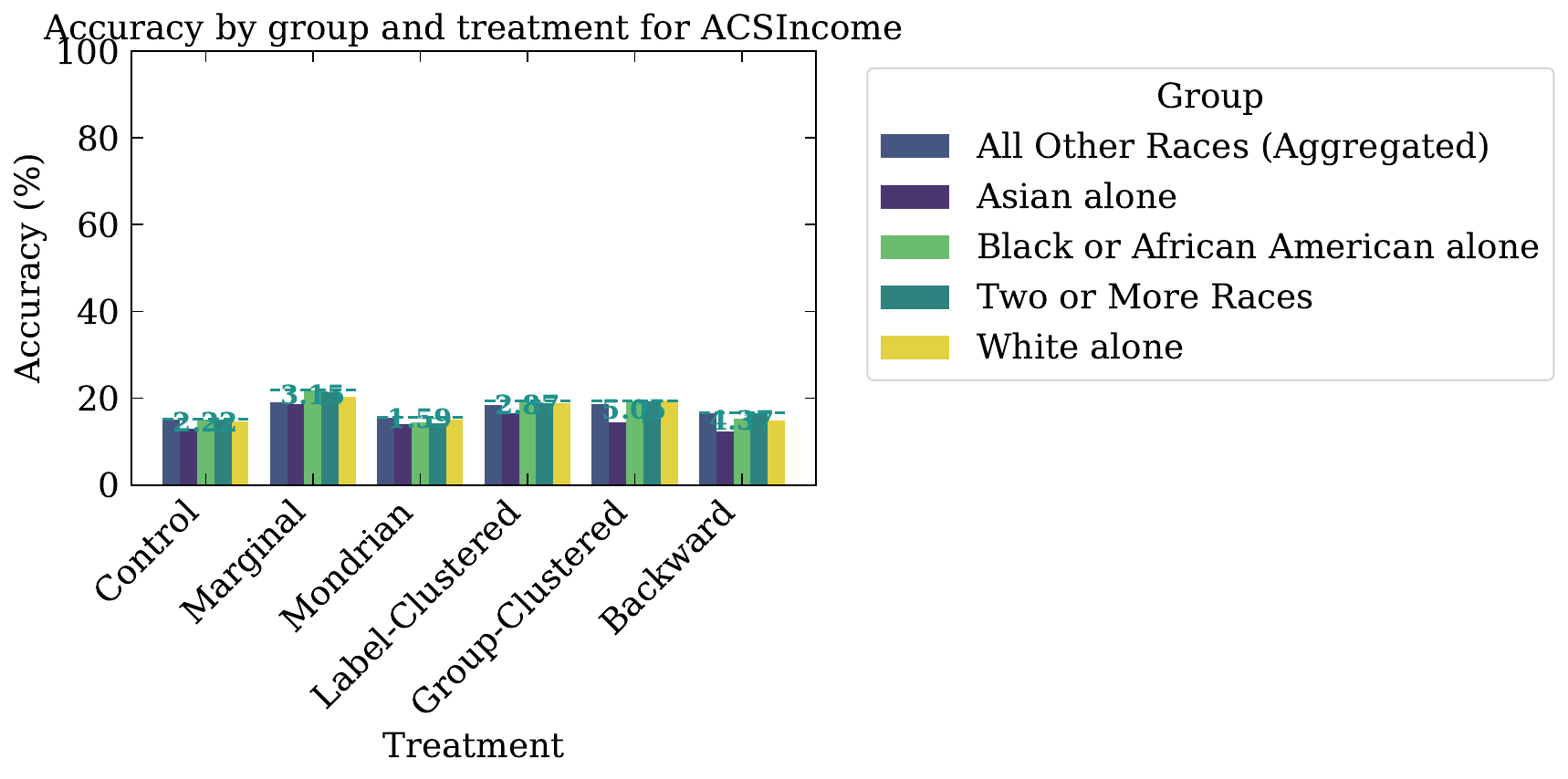}
    \caption{Accuracy rate by group and treatment}
    \label{fig: ACSIncome_acc}
  \end{subfigure}

  \caption{Experiment results of ACSIncome with $K = 2$ in both label-clustered and group-clustered CP. LLM-in-the-loop: Qwen2.5-7B.}
  \label{fig: ACSIncome_result}
\end{figure}

\subsection{Additional Coverage Gap and Set Size Gap Plots}\label{app:cov-gap-plot}

\begin{figure}[h]
  \centering

  \begin{subfigure}[t]{0.23\linewidth}
    \centering
\includegraphics[width=\linewidth, height=0.18\textheight]{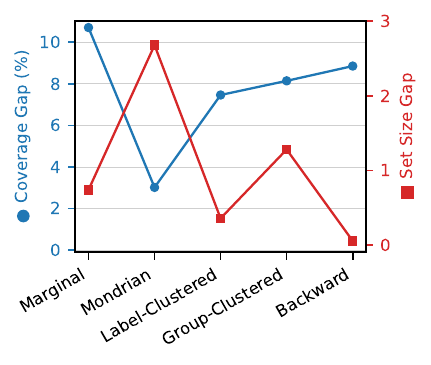}
    \caption{FACET}
    \label{fig: gap-facet-only}
  \end{subfigure}
  \begin{subfigure}[t]{0.23\linewidth}
    \centering
\includegraphics[width=\linewidth, height=0.18\textheight]{imgs/gap_bios_only_cp.pdf}
    \caption{BiosBias}
    \label{fig: gap-bios-only2}
  \end{subfigure}
  \begin{subfigure}[t]{0.23\linewidth}
    \centering
\includegraphics[width=\linewidth, height=0.18\textheight]{imgs/gap_ravdess_only_cp.pdf}
    \caption{RAVDESS}
    \label{fig: gap-ravdess-only2}
  \end{subfigure}
  \begin{subfigure}[t]{0.23\linewidth}
    \centering
\includegraphics[width=\linewidth, height=0.18\textheight]{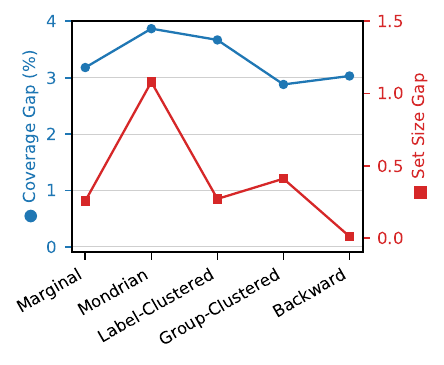}
    \caption{ACSIncome}
    \label{fig: gap-acsincome-only}
  \end{subfigure}

  \caption{Coverage gap (blue dots, left axis) and set size gap (red squares, right axis) across CP methods on FACET, BiosBias, RAVDESS, and ACSIncome.}
  \label{fig:gap-four-separate-extra}
\end{figure}

In \Cref{sec:proc-sub-corr} we examined the coverage gap and set size gap across CP methods and two datasets. \Cref{fig:gap-four-separate-extra} shows all four datasets together.

\subsection{Sensitivity of \textbf{maxROR} to the Number of Label Clusters}
\label[appsec]{app: maxROR-K-sensitivity}

To examine whether the $K$-sensitivity of the set-size gap discussed in \Cref{sec:exp-clustered-fairness} also carries over to downstream fairness, \Cref{tab:biosbias-k-sensitivity} reports the set-size gap and \textbf{maxROR} for Label-Clustered CP on a split of BiosBias across different values of $K$. This analysis uses GPT-4o-mini as the LLM-in-the-loop with $M = 5$ repeated predictions per task-treatment pair. The results show a qualitatively similar pattern: $K = 1$ (Marginal CP) performs poorly, with both a large set-size gap and high \textbf{maxROR}; several intermediate values of $K$ substantially reduce \textbf{maxROR}; and larger values of $K$ can cause the metric  to increase again. However, for this data split, the value of $K$ minimizing \textbf{maxROR} does not exactly coincide with the $K$ minimizing the set-size gap. This is expected because \textbf{maxROR} reflects downstream decision behavior and can depend on factors beyond set-size gap alone. Practically, these results suggest using set-size gap as a low-cost diagnostic to narrow $K$ to a small set of promising candidates, and then selecting among them using a downstream fairness metric when such data are available.

\begin{table}[H]
\centering
\small
\setlength{\tabcolsep}{9pt}
\caption{
Sensitivity of set-size gap and downstream fairness (\textbf{maxROR}) to the number of label clusters $K$ for Label-Clustered CP on BiosBias. 
The case $K = 1$ corresponds to Marginal CP. 
Set-size gap denotes the absolute difference in average prediction set size between female and male groups.
}
\label{tab:biosbias-k-sensitivity}
\begin{tabular}{lcccccccccc}
\toprule
\(K\) & \(1\) & \(2\) & \(3\) & \(4\) & \(5\) & \(6\) & \(7\) & \(8\) & \(9\) & \(10\) \\
\midrule
Set-size gap & .050 & .024 & .033 & .045 & .027 & .034 & .053 & .054 & .055 & .057 \\ \\
\textbf{maxROR} (\%) & 6.9 & 4.4 & 1.6 & 2.7 & 2.0 & 0.5 & 1.1 & 1.8 & 2.7 & 3.1 \\
\bottomrule
\end{tabular}
\end{table}

\end{document}